\pdfoutput=1
\documentclass[10pt,twocolumn,letterpaper]{article}
\ifdefined\pdfsuppressptexinfo\pdfsuppressptexinfo=-1\fi %
\usepackage[T1]{fontenc}
\usepackage[pagenumbers]{cvpr}
\usepackage{graphicx,amsmath,amssymb,booktabs,array,xcolor}
\input{table_style}
\usepackage{microtype,needspace}
\newcommand{\DSSuppRef}[1]{Supplementary Sec.~\ref{#1}}
\newcommand{\DSSuppTableRef}[1]{Supplementary Table~\ref{#1}}
\newcommand{\DSSuppFigureRef}[1]{Supplementary Figure~\ref{#1}}
\definecolor{cvprblue}{rgb}{0.21,0.49,0.74}
\usepackage[pagebackref,breaklinks,colorlinks,allcolors=cvprblue]{hyperref}
\def\confName{CVPR}
\def\confYear{2027}
\newcommand{\score}{\mathrm{score}}
\newcommand{\gen}{\mathrm{gen}}
\newcommand{\main}{\mathrm{main}}
\newcommand{\later}{\mathrm{later}}
\graphicspath{{figures/}}
\title{Interpreting DeepStack-Style Vision-Language Models:\\ Do Input or Deeper-Layer Visual Tokens Drive Answers?}
\author{%
Yucheng Du$^{1,*}$\quad Yi Nian$^{1,*}$\quad Ruixue Lian$^{2,*}$\quad Yue Zhao$^{1,\dagger}$\quad Xiyang Hu$^{3,\dagger}$\\[3pt]
{\normalsize $^{1}$University of Southern California\quad $^{2}$University of Wisconsin--Madison\quad $^{3}$Arizona State University}
}
\begin{document}
\twocolumn[{%
\renewcommand\twocolumn[1][]{#1}%
\maketitle
\begin{center}
\centering
\captionsetup{type=figure}
\includegraphics[width=\textwidth]{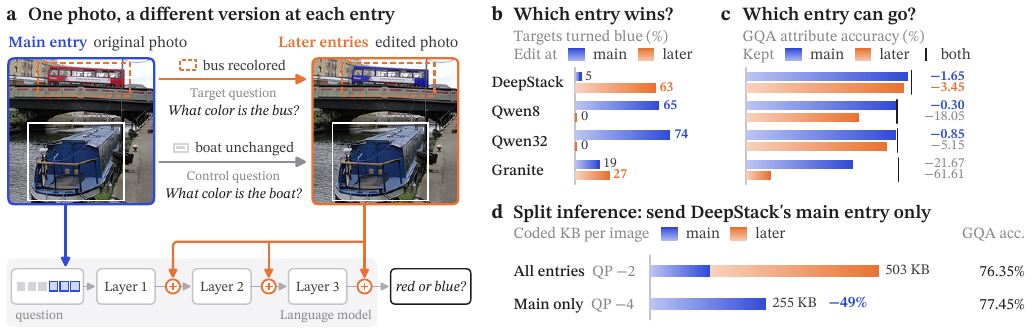}
\captionof{figure}{\textbf{Each entry can receive a different version of one image, and the entry that wins a conflict is not always the one a model needs.} (a)~Only the later entries see the recolored bus. (b)~Target answers turned blue when only main or only later is edited (124--128 edits per model; Table~\ref{tab:natural}). (c)~GQA attribute accuracy with one entry kept and the other zeroed; numbers: change in points from keeping both (Table~\ref{tab:coding}). (d)~DeepStack's coded bytes per image and accuracy when sending all entries or only main. Qwen8/32: Qwen3-VL-8B/32B.}
\label{fig:overview}
\end{center}%
}]
{\let\thefootnote\relax\footnotetext{$^{*}$These authors contributed equally. $^{\dagger}$Corresponding authors.}}
\begin{abstract}
DeepStack-style vision-language models inject visual tokens at the language-model input (\emph{main} entry) and deeper decoder layers (\emph{later} entries), yet same-image visual question-answering accuracy cannot reveal how trained models divide work between them. We therefore feed main and later mismatched versions of one image, differing in an object's attribute (e.g., color), visual quality (undegraded versus desaturated, blurred, or downsampled), or both. Under a single-object color conflict on photographs, original DeepStack follows the later entries, whereas Qwen3-VL-8B and 32B follow main. When main receives a desaturated synthetic image while later keeps an undegraded recolored object, later shifts the log-probability margin between the two colors more strongly than with undegraded main across all four tested models (including Granite Vision). Whether later can also recover lost accuracy, however, depends sharply on the model: when blur at both entries costs both Qwen models 50 percentage points on a synthetic visual task, keeping later undegraded recovers 48 points in Qwen3-VL-32B but only 8 in Qwen3-VL-8B. Unlike conflict dominance, these recovery patterns identify which entry each model can zero out with little GQA attribute loss. Although DeepStack follows later under conflict, either entry suffices on GQA attributes (at most 3.45 points lost when zeroed), letting split inference (transmitting compressed visual features) send only main at higher precision and cut bytes by 49\% at comparable accuracy versus compressing all entries. Thus, an entry's role cannot be inferred from the architecture or same-image accuracy, and must be diagnosed per checkpoint before pruning or compressing features.
\end{abstract}

\section{Introduction}
\label{sec:intro}
Standard evaluations of multi-entry vision-language models present the same image to every visual injection point, leaving it unclear how a trained model divides work across them. Following DeepStack \citep{meng2024deepstack}, architectures such as Qwen3-VL \citep{qwen2025qwen3vl} and Granite Vision \citep{ibm2026granitevision} inject visual tokens at the language-model input (\emph{main entry}, or \emph{main}) and again at deeper decoder layers (\emph{later entries}, or \emph{later}). These interfaces carry different representations: in original DeepStack (DeepStack-L-Vicuna-7B, DeepStack below), the later entries supply higher-resolution spatial crops, whereas in Qwen3-VL they carry intermediate vision-encoder features. Benchmark gains show that adding multi-depth injection improves overall performance, but not what role each entry plays at inference.

Once the language model receives an image through main, later entries could play four distinct roles: deciding the answer when entries disagree, gaining influence as main's evidence weakens, repairing errors caused by degraded main inputs, or duplicating main so that one entry becomes expendable. When both entries see the same image, correct answers cannot distinguish among these roles. Yet knowing which entry carries the answer matters for interpretability, architecture redesign, and compression, such as split inference where coded features from every entry are transmitted between machines \citep{heidari2026compressing}.

To isolate each role, we feed main and the later entries mismatched versions of one image (Figure~\ref{fig:overview}). In our conflict test, one object is recolored for main alone or for later alone, and a question about its color identifies which entry governs the answer. Prior analyses instead ablate DeepStack-style features completely \citep{qwen2025qwen3vl,rahman2026temporal} or patch hidden states that already blend both entries \citep{cui2026dual}. We evaluate across photographs, synthetic scenes of shapes and short text, GQA, and DocVQA. The conflict test spans nine checkpoints: the Qwen lineage (five Qwen3-VL sizes and two Cosmos3 models derived from them \citep{nvidia2026cosmos3}), Granite Vision, and DeepStack; the remaining tests focus on Qwen3-VL-8B/32B (Qwen8/32), Granite, and DeepStack.

\emph{Later entries can dominate photograph answers.} When only DeepStack's later entries receive the recolored object, 63\% of target answers change, against 5\% when only main does. Qwen8/32 show the opposite pattern: main changes 65--74\% of target answers, whereas later changes none. All seven Qwen-lineage checkpoints favor main, and Granite leans weakly toward later. Checkpoints sharing one multi-entry blueprint thus assign the deciding role to opposite entries.

\emph{Later influence depends on the evidence already supplied through main.} Although Qwen's later entries change no photograph answers, they are not inactive. When we degrade only main's input on synthetic color scenes while holding conflicting later features fixed, their shift on the log-probability margin between the two colors grows in all four models, even where degradation causes no errors. On stripe orientation in Qwen32, unchanged conflicting later features switch 20\% of target answers with undegraded main but 81\% with mildly blurred main, even though non-conflict accuracy stays at 100\%.

\emph{Later entries can repair degraded main inputs, subject to model and task limits.} Gaining influence under conflict does not guarantee error recovery. When both entries receive a degraded image, restoring undegraded later features recovers part of the lost GQA accuracy in all four models. Yet at the strongest stripe blur, where Qwen8 and Qwen32 suffer equal accuracy drops, undegraded later features restore nearly all of the loss in Qwen32 but very little in Qwen8, and neither model recovers substantially on short text or DocVQA.

\emph{Expendable entries vary by model.} Where one entry can substitute for another, a system need not retain both. On held-out GQA attribute questions, five of six pre-specified single-entry predictions derived from our recovery profiles held: Qwen8 maintains accuracy with later zeroed, and DeepStack retains at least 76.1\% with either entry zeroed (versus 79.5\% with both intact), showing that its later entries govern conflicts yet add little accuracy here. Consequently, transmitting only DeepStack's main entry in split inference reduces coded bitstream size by 49\% at comparable accuracy (Section~\ref{sec:coding}).

In summary, we make three contributions. First, we introduce a training-free, entry-wise intervention protocol that feeds mismatched image versions to the main and later entries of multi-entry vision-language models, separating conflict dominance, quality-conditioned influence, error recovery, and entry redundancy without modifying model weights. Second, across nine checkpoints spanning three DeepStack-style implementations, we show that an entry's role cannot be inferred from the architecture or same-image accuracy, and that conflict dominance does not determine which entry a model can drop. Third, we demonstrate that checkpoint-specific recovery profiles predict expendable entries on held-out GQA questions and directly inform practical feature pruning and split-inference compression.

\section{Related work}
\label{sec:related}
\begin{table*}[t]
\centering
\caption{\textbf{Related analyses differ in the information sources and routes they separate.} We vary each entry's evidence separately to distinguish influence (under conflict or degraded main), recovery, and which entry a model can lose.}
\label{tab:related_work}
\begingroup
\fontsize{9}{10.5}\selectfont
\setlength{\tabcolsep}{5pt}
\renewcommand{\arraystretch}{1.08}
\begin{tabularx}{\linewidth}{@{}>{\raggedright\arraybackslash}p{.16\linewidth}>{\raggedright\arraybackslash}p{.30\linewidth}>{\raggedright\arraybackslash}X@{}}
\toprule
\rowcolor{DSTableHeader}
Study & What is separated? & Question or finding \\
\midrule
Zhang et al.\ \citep{zhang2025crossmodal} & Image, question, and final-token positions & Two stages transfer global and question-relevant visual information into question representations. \\
\addlinespace[3pt]
Salazar et al.\ \citep{salazar2026pathways} & Direct image-token and text-mediated pathways & Pathway use depends on task and prompt; text mediation can support fallback. \\
\addlinespace[3pt]
Ghosh et al.\ \citep{ghosh2026whichsource} & Conflicting image and text sources & Generation and likelihood margins reveal task-dependent reliance shifts under degradation, including after accuracy-loss calibration. \\
\addlinespace[3pt]
Rahman\ \citep{rahman2026temporal} & Visual sequence and positional order & Similar temporal scores conceal different channel reliance and failure modes. \\
\addlinespace[3pt]
Cui et al.\ \citep{cui2026dual} & Visual ordering and language-model backup representations & Visual ordering supports spatial binding; amplifying it improves natural-image spatial judgments. \\
\addlinespace[3pt]
\rowcolor{DSTableAccent}
Our work & Main and later visual entries & Later dominates in DeepStack, main in Qwen3-VL; main quality changes later influence; recovery and entry removal vary by model. \\
\bottomrule
\end{tabularx}
\endgroup
\end{table*}
\textbf{Visual interfaces.}
DeepStack adds spatial visual features at several decoder depths \citep{meng2024deepstack}, Qwen3-VL adds intermediate encoder features \citep{qwen2025qwen3vl}, Cosmos3 keeps the Qwen3-VL interface after continued training \citep{nvidia2026cosmos3}, and Granite uses a DeepStack variant \citep{ibm2026granitevision}. Other interfaces, object benchmarks, and intervention caveats appear in \DSSuppRef{app:related}. These works specify what each design feeds to its entries; we keep each model's own entry layout (Section~\ref{sec:setup}) and measure how the trained model uses each entry.

\textbf{Internal visual pathways.}
Interventions inside vision-language models trace how visual information is carried and used. Causal tracing restores clean visual states after corruption \citep{palit2023causal}, and patch and attention interventions study object-local information \citep{neo2025interpreting}. The location of an OCR routing bottleneck depends on architecture \citep{steinberg2026vision}, and global views and local crops can play different roles in localization and classification \citep{schaumloffel2026localization}. In video models, activation patching reveals an order-recovery gap that persists after DeepStack-style features are removed \citep{rahman2026temporal}. None of these studies gives a model's visual entries different versions of one image; we vary each entry's input separately under conflicting sources and degraded images (Table~\ref{tab:related_work}).

\textbf{Spatial binding.}
Position-ID and spatial-ID interventions redirect feature retrieval and spatial judgments \citep{assouel2026visualsymbolic,kang2026linear}, and Cui et al.\ patch Qwen3-VL's early-layer states and link the effect to its DeepStack-style injection \citep{cui2026dual}. Those states mix upstream computation with the injected features, whereas we change each entry's input separately and exchange object layouts to check that entry effects follow the object, not the position (Section~\ref{sec:spatial}). Our finding that Qwen3-VL's later entries barely change color answers is therefore compatible with their finding that visual ordering supports spatial binding.

\section{Following an attribute through visual entries}
\label{sec:method}
\subsection{One visual change, two object queries}
Sources $a$ and $b$ contain the same objects and layout and differ only in one object's attribute. The target query asks about this edited object and the control query about the unchanged object; both receive the same intervention (Figure~\ref{fig:overview}). At entry $e$, with features $v_e^a$ and $v_e^b$ from the two sources, selected positions receive
\begin{equation}
 v_e^a+k(v_e^b-v_e^a),
 \label{eq:patch}
\end{equation}
where $k=0$ keeps source $a$ and raw strength $k=1$ installs source $b$. We intervene on main alone or on all later entries together. On photographs, the other entry keeps the original image (\DSSuppTableRef{tab:intervention_spec}). A \emph{local transfer} selects the tokens that overlap the edited object's bounding box; being contextualized, these tokens can also carry information from outside the box. Other positions keep the surrounding scene (the \emph{background}), whose strength $\alpha$ is defined in Section~\ref{sec:setup}; $\alpha=1$ leaves the image intact.

Both queries are scored toward the edited value: for a red-to-blue edit, toward $c_b=$ ``blue'' rather than $c_a=$ ``red''. In every synthetic set, the edit gives the target the control's value, so the control's correct answer is already $c_b$. The same holds for 74 of 128 photograph controls (\DSSuppRef{app:quality}). For query $q$ in intervention state $z$, the candidate score is $y_q^{\score}=\log p(c_b\mid q,z)-\log p(c_a\mid q,z)$, summed over answer tokens without length normalization. Generation uses $y_q^{\gen}=\mathbf{1}[\widehat c_q=c_b]$ for a valid answer. Candidate scores measure graded support, whereas generation records the final answer. Because the parser normalizes case and whitespace and accepts more forms than the scored strings, the two readouts can disagree even without an intervention (\DSSuppRef{app:statistics}). Scores are not pooled across models or answer formats.

Let $g\in\{\main,\later\}$ be main or all later entries together. In state $z_a$, $g$ receives source $a$ ($k=0$), and in $z_b$ it receives the intervention ($k=1$ at raw strength). The directed effect on query $q$ is $E_{q,g}=y_q(z_b)-y_q(z_a)$. Relative object selectivity and its entry contrast are
\begin{equation}
 R_g=E_{\mathrm{target},g}-E_{\mathrm{control},g},\quad
 \Delta R=R_{\main}-R_{\later}.
 \label{eq:selectivity}
\end{equation}
Positive $R_g$ means that the target moves toward $c_b$ more than the control, not that only the target moves, so we report both components. Generation effects are net changes toward $c_b$, not accuracy or flip rates. Because a correctly answered control already answers $c_b$, generation cannot detect a shift toward $c_b$ that both queries share, whereas scores still register the control's response. Subtraction cancels a shared shift only if both queries move equally, so scores limit a shared bias but do not exclude it.

\begin{figure*}[t]
\centering
\includegraphics[width=\textwidth]{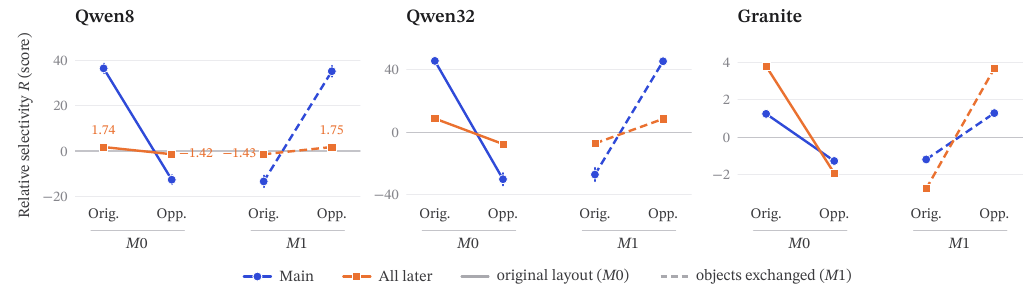}
\caption{\textbf{Entry effects follow the object at each position.} Each point uses the 64 color templates of the controlled panel, direct answers, and background strength $\alpha=0.08$; each model has its own score scale. Features and their region mask are placed at the original (Orig.) or opposite (Opp.) coordinate, in the original layout ($M0$) or with the objects exchanged ($M1$). Numbers label Qwen8's small later effects. Error bars are descriptive normal 95\% intervals; most are smaller than the markers.}
\label{fig:spatial}
\end{figure*}

\subsection{Position and intervention magnitude}
Local transfers test whether an effect follows the object or the image position. The transferred features and their mask go to the edited object's position (original destination) or to the same position in the other image half (opposite destination), and the \emph{destination effect} is $R_g$ at the opposite minus $R_g$ at the original destination. We measure it in the original ($M0$) and exchanged ($M1$) layouts. Exchange changes every entry's unselected positions but not the transferred features, so if relative selectivity follows the object currently at a position, the destination effect reverses. The \emph{spatial interaction} is its $M1$ value minus its $M0$ value, paired within scenes (\DSSuppRef{app:spec}). Because main's input also changes under exchange, the interaction tests whether effects follow the current object, not which entry carries that object.

Magnitude controls test whether feature size alone explains an entry contrast. Each entry's magnitude (its \emph{dose} in the supplement) is the RMS of $v_e^b-v_e^a$ divided by that of the decoder hidden state receiving the entry, and several later entries combine by root sum of squares (RSS) or by sum (L1). To match main's magnitude, we keep main at raw strength and rescale the later difference; matching to later's magnitude reverses the roles (\DSSuppRef{app:dose}).

\subsection{Conditional influence and recovery}
\label{sec:quality_method}
To test whether later influence depends on main's image, main receives an undegraded (\emph{clean}, $C$) or degraded ($D$) version of $a$, while later receives one of two \emph{fixed endpoints}, computed once and reused under $C$ and $D$. Clean source-$a$ features give the \emph{non-conflict input}, and the conflict endpoint installs source $b$ ($k=1$) on the edited object's region. Only the decoder states built from main's image then change with main quality, and the change in later influence is $R_{\later}(D)-R_{\later}(C)$, not the entry contrast $\Delta R$.

Recovery involves no conflict: main and later each receive the clean or degraded version of the same image at all positions, giving four cells. All cells are scored against the clean image's answer and use the same degraded image. Let $A_{uv}$ be accuracy with main quality $u$ and later quality $v$ (for benchmarks, the metric of Section~\ref{sec:setup}). With main degraded, clean later features give a gain $G_D$, and $I_{\mathrm{rec}}$ is $G_D$ minus their gain with main clean:
\begin{equation}
 G_D=A_{DC}-A_{DD},\quad I_{\mathrm{rec}}=G_D-(A_{CC}-A_{CD}).
 \label{eq:recovery}
\end{equation}
Accuracy is \emph{main-limited} when it tracks main's quality alone ($A_{CD}\approx A_{CC}$, $A_{DC}\approx A_{DD}$), \emph{later-limited} when it tracks later's quality alone ($A_{DC}\approx A_{CC}$, $A_{CD}\approx A_{DD}$), and \emph{either-sufficient} when one clean entry suffices ($A_{CD}\approx A_{DC}\approx A_{CC}$). In the last case, $G_D$ and $I_{\mathrm{rec}}$ approach the joint loss $A_{CC}-A_{DD}$. Each label applies to one metric, answer format, and level, and only where the joint loss is clear. For the same task and degradation, conflict and recovery share the $CC$ and $DC$ inputs, so $A_{CC}$ and $A_{DC}$ are the accuracies on the non-conflict input.

\section{Experimental setup}
\label{sec:setup}
\begin{table*}[t]
\centering
\caption{\textbf{On photographs, answers follow the later entries in DeepStack and the main entry in Qwen3-VL.} Generation selectivity is $R=E_{\mathrm{target}}-E_{\mathrm{control}}$; positive $\Delta R$ means larger relative selectivity at main. Conditions are $M0$, $\alpha=1$, direct answers, and raw $k=1$. Cells are means [descriptive normal 95\% intervals]. Each estimate uses $n$ complete photographs out of 64. Main is one entry; later contains three entries for the Qwen lineage, seven for Granite, and four for DeepStack on these images. Shaded: models used in Sections~\ref{sec:conditional}--\ref{sec:coding}.}
\label{tab:natural}
\begingroup
\fontsize{9}{10.5}\selectfont
\setlength{\tabcolsep}{2.8pt}
\renewcommand{\arraystretch}{1.08}
\DSalign{table1-1}
\begin{tabularx}{\linewidth}{@{}l*{3}{>{\centering\arraybackslash}Xr}@{}}
\toprule
& \multicolumn{2}{c}{Main} & \multicolumn{2}{c}{All later entries} & \multicolumn{2}{c}{Paired difference} \\
\cmidrule(lr){2-3}\cmidrule(lr){4-5}\cmidrule(l){6-7}
\rowcolor{DSTableHeader}
Checkpoint & $R_{\mathrm{main}}$ [95\% interval] & $n$ & $R_{\mathrm{later}}$ [95\% interval] & $n$ & $\Delta R$ [95\% interval] & $n$ \\
\midrule
Qwen3-VL-2B & \DSv{0.414}{[0.313,0.515]} & \DSn{64} & \DSv{0.008}{[-0.027,0.042]} & \DSn{64} & \DSv{0.406}{[0.307,0.506]} & \DSn{64} \\[1pt]
Qwen3-VL-4B & \DSv{0.694}{[0.621,0.766]} & \DSn{62} & \DSv{-0.008}{[-0.023,0.008]} & \DSn{63} & \DSv{0.702}{[0.632,0.771]} & \DSn{62} \\[1pt]
\rowcolor{DSTableAccent}
Qwen3-VL-8B & \DSv{0.645}{[0.563,0.728]} & \DSn{62} & \DSv{0.000}{[0.000,0.000]} & \DSn{64} & \DSv{0.645}{[0.563,0.728]} & \DSn{62} \\[1pt]
Qwen3-VL-30B-A3B & \DSv{0.516}{[0.427,0.604]} & \DSn{63} & \DSv{0.016}{[-0.022,0.053]} & \DSn{64} & \DSv{0.500}{[0.409,0.591]} & \DSn{63} \\[1pt]
\rowcolor{DSTableAccent}
Qwen3-VL-32B & \DSv{0.730}{[0.651,0.810]} & \DSn{63} & \DSv{0.000}{[0.000,0.000]} & \DSn{64} & \DSv{0.730}{[0.651,0.810]} & \DSn{63} \\[1pt]
Cosmos3-Nano & \DSv{0.605}{[0.523,0.687]} & \DSn{62} & \DSv{0.000}{[0.000,0.000]} & \DSn{64} & \DSv{0.605}{[0.523,0.687]} & \DSn{62} \\[1pt]
Cosmos3-Super & \DSv{0.754}{[0.681,0.827]} & \DSn{63} & \DSv{0.008}{[-0.008,0.023]} & \DSn{63} & \DSv{0.746}{[0.670,0.822]} & \DSn{63} \\[1pt]
\addlinespace[2pt]
\rowcolor{DSTableAccent}
Granite Vision 4.1-4B & \DSv{0.180}{[0.113,0.246]} & \DSn{64} & \DSv{0.266}{[0.184,0.347]} & \DSn{64} & \DSv{-0.086}{[-0.160,-0.012]} & \DSn{64} \\[1pt]
\rowcolor{DSTableAccent}
DeepStack-L-Vicuna-7B & \DSv{0.047}{[0.011,0.083]} & \DSn{64} & \DSv{0.594}{[0.496,0.691]} & \DSn{64} & \DSv{-0.547}{[-0.649,-0.445]} & \DSn{64} \\[1pt]
\bottomrule
\end{tabularx}
\endgroup
\end{table*}
\textbf{Materials.}
The photograph panel contains 64 photographs from GQA and Visual Genome, selected without inspecting model outputs \citep{hudson2019gqa,krishna2017visualgenome}. In each, two checked, nonoverlapping objects are recolored in turn on a $384\times384$ canvas, with a component mask restricting each edit to one object, which yields 128 red-to-blue image pairs and 256 queries. The controlled panel has 64 synthetic templates of two shapes each, crossing shape pairs, layouts, and attribute assignments; color and stripe versions share geometry. Local transfers vary layout, destination, entry, and background strength $\alpha\in\{0.08,1\}$. Background strength blends the object textures of a template, or a whole photograph, toward gray, and layout exchange swaps the two halves of a template or mirrors a photograph. \DSSuppRef{app:methods} gives construction and filtering details.

The fixed-contrast experiment (Section~\ref{sec:quality_method}) scales image contrast by 0.12 without resizing. It uses the same 64 photographs and 32 further templates for color, stripe orientation, the position of a mark within one shape (relative position), and reading short text (512 queries). It also blurs GQA and DocVQA images, which causes little loss (\DSSuppRef{app:v14}).

The calibrated experiment desaturates color, blurs stripes and relative position (scaled to the stimulus), and lowers the resolution of text, GQA, and DocVQA. On a separate calibration set, we chose three levels per task using only clean accuracy and accuracy with both entries degraded, so that with both entries degraded, models keep a median of 0.75, 0.5, and 0.25 of their clean above-chance accuracy. These \emph{shared levels} and the degraded images are the same for all models. Evaluation uses other items: 32 templates per synthetic task (balanced for object side and answer), 192 GQA validation-balanced images per question group, and 192 Single-Document DocVQA validation documents \citep{hudson2019gqa,mathew2021docvqa}. GQA answers are matched after normalizing case and punctuation, and DocVQA uses the organizers' ANLS (average normalized Levenshtein similarity); other metrics are sensitivity analyses (\DSSuppRef{app:v15}).

\textbf{Models and interfaces.}
We evaluate five Qwen3-VL sizes (2B, 4B, 8B, 30B-A3B, 32B), the Cosmos3-Nano/Super understanding towers, Granite Vision 4.1-4B, and DeepStack-L-Vicuna-7B, all in BF16 on NVIDIA GPUs \citep{qwen2025qwen3vl,nvidia2026cosmosnano,nvidia2026cosmossuper,ibm2026granitevision,meng2024deepstack}. The Cosmos3 towers descend from Qwen8/32 \citep{nvidia2026cosmos3}, so the seven Qwen-lineage checkpoints share one interface and give lineage coverage, not independent architectures.

Qwen's main entry uses the final encoder features, and its three later entries use intermediate encoder features. Separate learned projectors (mergers) give all four the same visual grid and token count, and the later features are added to the outputs of the first three decoder layers. Granite's main entry also uses the final encoder features, whereas its seven later entries are three intermediate-layer entries and four spatial Q-Former entries; its main and spatial branches reduce their visual grids to the same $12\times12$ output. DeepStack instead uses one CLIP layer and one shared projector for a global main view and four spatial later entries (groups), which are added to the outputs of its first four decoder layers. Rectangular synthetic inputs activate only two of the four groups, and the model's preprocessing zero-pads the rest. On square photographs, DeepStack's main entry sees one 336-pixel view as 576 tokens, whereas its four later groups take 576 tokens each from a 672-pixel canvas, together holding four times main's tokens at twice its linear sampling density. \DSSuppRef{app:reproducibility} gives exact checkpoints and geometry.

\textbf{Readouts and inference.}
Outside the benchmarks, each query has three answer formats. \emph{Direct answers} name the attribute (``blue''), whereas \emph{identity} and \emph{cyclic labels} give the number of a listed option in fixed (identity) or shifted (cyclic) order; with two options, the shift reverses them. Greedy generation allows 32 tokens, or 64 for benchmarks, and items are never filtered by model accuracy. Self-patching, which reinstalls a source's own features, checks that interventions are delivered as specified.

Comparisons are paired within photographs, templates, or source documents, which define $n$, so repeated queries do not enlarge it. Conflict contrasts use complete cases within each edit: every answer the contrast needs must be valid, so the $n$ for $\Delta R$ can differ from that for $R_{\main}$ or $R_{\later}$. Benchmark questions are averaged within source images or documents first, and recovery accuracy counts invalid answers as incorrect. Brackets give 95\% intervals: descriptive normal for both panels and for benchmarks, Student-$t$ for templates and photographs in the fixed-contrast and calibrated experiments, and paired bootstrap in Section~\ref{sec:coding}. All intervals are unadjusted. Apart from the predictions fixed in advance in Section~\ref{sec:coding}, analyses are exploratory.

\section{Results}
\label{sec:results}
\subsection{Entry effects follow the object at each position}
\label{sec:spatial}
\begin{figure*}[t]
\centering
\includegraphics[width=\textwidth]{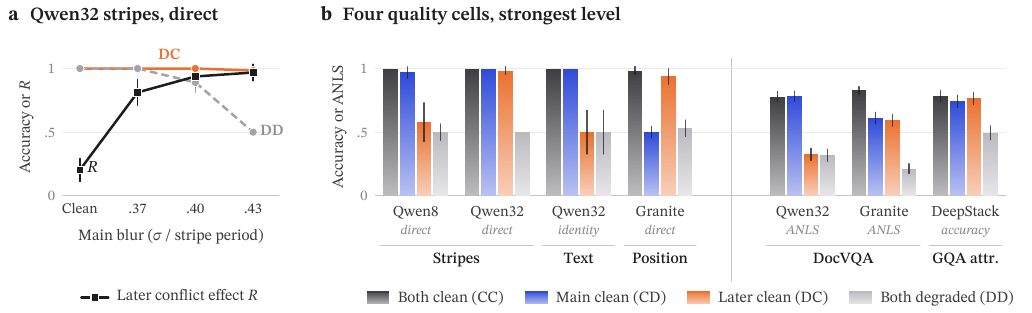}
\caption{\textbf{Fixed later features gain influence under mild main blur; recovery depends on model and task.} (a) Qwen32 stripes, direct answers, as main blur grows ($\sigma$ as a fraction of the stripe period): DC and DD accuracy and the later conflict effect $R$ (Eq.~\ref{eq:selectivity}, later inputs fixed). (b) Strongest calibrated level per task. Cells name main first (C clean, D degraded); bar colors show which entries are clean. Generation accuracy on 32 synthetic templates in the answer format named under each model, GQA accuracy on 192 images, and ANLS on 192 DocVQA documents. Qwen32's stripe cells are identical under identity labels. Error bars are descriptive, unadjusted 95\% intervals; cells without variation have none. Complete values: \DSSuppRef{app:v15}.}
\label{fig:conditional_quality}
\end{figure*}
A local transfer acts on the object at the transferred position, not on the image coordinate.

Before comparing entries, we check this premise. Moving the transferred features to the other object's position changes relative selectivity, and exchanging the two objects reverses this change (Figure~\ref{fig:spatial}). Since both queries receive the same intervention, the object now at the transferred position decides which query responds more; for Qwen32's later color scores, for example, the paired spatial interaction is $32.0$ [$30.0,34.0$].

DeepStack's later entries show the same reversal in color generation. On stripes, paired comparisons in all three answer formats favor a larger spatial interaction for main in Qwen8/32 and for later in Granite. At full background strength ($\alpha=1$), the Cosmos3 comparisons follow the Qwen direction (Supplementary Secs.~\ref{app:quality}, \ref{app:figures}, and~\ref{app:spatial_more}).

\subsection{DeepStack follows later, Qwen follows main}
\label{sec:natural}
On photographs, answers follow the later entries in DeepStack and the main entry in all seven Qwen-lineage checkpoints, whereas Granite leans weakly toward later (Table~\ref{tab:natural}).

\emph{Target and control components explain the contrast.} On the photographs with complete results for both entries, main changes 81/124 target answers toward $b$ in Qwen8 and 93/126 in Qwen32 but only one control answer in each, and neither model changes any answer through its later entries. Later interventions still raise target scores more than control scores, by $0.69$ [$0.52,0.85$] in Qwen8 and $1.30$ [$1.06,1.55$] in Qwen32 (\DSSuppRef{app:scores_answers}). DeepStack's later entries change 80/128 targets and four controls, whereas its main entry changes six targets and no controls (\DSSuppFigureRef{fig:entry_components}). The contrast holds when we keep only queries that the model first answers red (\DSSuppTableRef{tab:control_room}; \DSSuppRef{app:quality}).

\emph{Answer formats and magnitude matching keep the direction in Qwen8, Qwen32, Cosmos3-Super, and DeepStack.} In every format, Qwen8, Qwen32, and Cosmos3-Super have $\Delta R$ intervals above zero for scores and generation, at raw strength and under all four magnitude matchings (to main's or later's magnitude, by RSS or L1; \DSSuppRef{app:census}). DeepStack's intervals are below zero in all 15 comparisons per readout (five settings in three formats). At raw strength, Granite's contrasts favor later or have intervals containing zero in every format, most clearly for generation with identity labels ($-0.22$ [$-0.30,-0.14$]). Under L1 matching, however, its score contrast for direct answers reverses sign (\DSSuppRef{app:boundaries}).

\subsection{Later influence depends on main quality}
\label{sec:conditional}
Degrading main strengthens the effect of unchanged later features on synthetic color in all four models, even where accuracy does not change.

In the fixed-contrast experiment, lowering main contrast raises the score selectivity of fixed later features, $R_{\later}(D)-R_{\later}(C)$, with intervals above zero for all four models in all three formats (32 templates; \DSSuppTableRef{tab:v14_conflict_full}). Because the later features are identical under both main qualities, the change must come from the decoder states built from main's image. In the calibrated experiment, all 36 color score changes (four models, three formats, three levels) again have intervals above zero, and at the strongest desaturation, generation selectivity with direct answers also rises in all four models (Qwen32: $0.42$ [$0.28,0.57$]).

Accuracy does not show this increase. At the mildest shared desaturation level, all four models answer every non-conflict query correctly whether main is clean or degraded ($A_{CC}=A_{DC}=1$; direct answers, 32 templates). Yet the fixed later conflict has a larger score effect under degraded main in every model, with all four paired intervals above zero, and generation selectivity also increases, by $0.19$ [$0.06,0.31$] in Granite and $0.38$ [$0.24,0.51$] in DeepStack (\DSSuppRef{app:v15}).

\emph{A large response change at perfect accuracy.} At the mildest stripe blur, Qwen32 answers every query correctly in all four recovery cells, and degrading main alone lowers the correct answer's log odds only from $19.6$ to $18.5$. Yet the later features switch $0.20$ of target answers toward $b$ with clean main and $0.81$ with degraded main, a rise of $0.61$ [$0.50,0.72$] (Figure~\ref{fig:conditional_quality}a). The rise is again $0.61$ with identity labels and $0.47$ [$0.34,0.60$] with cyclic labels, and it also holds for relative position (\DSSuppRef{app:conditional_more}). Candidate scores already show the change: the later intervention's mean effect on the target score rises from $8.8$ to $27.6$, whereas its effect on the control moves only from $-1.1$ to $-1.5$ (paired selectivity change $19.16$ [$16.31,22.02$]). If main and later contributions simply added, the fixed later change would have the same score effect under both main states, so the dependence is not only a by-product of turning scores into answers.

The direction varies outside synthetic color. Degrading main lowers later score selectivity for Qwen8's stripes at the strongest blur ($-1.68$ [$-2.40,-0.97$]) and for Granite's relative position. On photographs with direct answers, lowering main contrast increases later score selectivity in Qwen8/32, with an uncertain change in DeepStack and a negative change in Granite. The Qwen increase barely reaches answers: Qwen32's generation selectivity rises by only $0.01$ [$-0.03,0.04$] (\DSSuppRef{app:v14}). With clean main, later interventions on photographs change Qwen8/32's candidate scores little: a median 1.04\% of the pre-intervention score in Qwen8 and 2.35\% in Qwen32, against 77.85\% in Granite (\DSSuppRef{app:scores_answers}).

\subsection{Later recovery depends on model and task}
\label{sec:recovery}
Clean later features recover part of the accuracy lost to degradation in every model, but how much depends on the model and the task.

On GQA, we restore later under degraded main at the strongest shared levels: a 5-pixel long side for attributes and 10 for spatial questions. In all four models, $G_D$ ($0.07$--$0.27$ for attributes, $0.06$--$0.28$ for spatial questions) and $I_{\mathrm{rec}}$ have intervals above zero for both tasks, so each gain exceeds the gain under clean main (\DSSuppRef{app:v15}).

On GQA attributes, the four cells show which entry limits accuracy in each model. Qwen32's $G_D$ grows over the three shared levels to $0.22$, with $I_{\mathrm{rec}}=0.19$ [$0.12,0.26$] at the strongest level, whereas Qwen8's is only $0.07$ there. DeepStack is either-sufficient: degrading either entry alone costs at most $0.04$ of its $0.29$ joint loss ($I_{\mathrm{rec}}=0.23$ [$0.17,0.29$]). Granite is closer to later-limited: degrading later alone costs $0.20$ of its $0.27$ joint loss, leaving $I_{\mathrm{rec}}$ at $0.06$--$0.07$ across levels.

\emph{Equal stripe losses, different recovery.} With the same degraded images and direct answers, Qwen8 and Qwen32 both lose $0.50$ stripe accuracy, yet clean later features restore $0.48$ [$0.45,0.52$] of it in Qwen32 and only $0.08$ [$-0.06,0.22$] in Qwen8, a paired difference of $0.41$ [$0.27,0.55$] (Figure~\ref{fig:conditional_quality}b). Qwen8 is thus closer to main-limited, and Qwen32 is close to either-sufficient. With identity labels, Qwen32 again recovers stripes but not text: both tasks fall from $1.00$ to $0.50$ when both entries are degraded, but clean later features restore none of the text accuracy ($0.00$ [$-0.05,0.05$]), a recovery difference of $0.48$ [$0.43,0.54$] from stripes. Qwen8's text gain has no interval above zero at any level or in any format. Granite's relative-position accuracy is later-limited: degrading later alone lowers $A_{CD}$ to $0.50$, close to $A_{DD}$, so clean later features help about as much under either main quality, and their $I_{\mathrm{rec}}$ interval contains zero despite a large recovery of $0.41$ [$0.30,0.51$] (\DSSuppRef{app:v15}).

\emph{Little DocVQA recovery in Qwen8/32.} At a 180-pixel long side, Qwen8 and Qwen32 lose $0.52$ and $0.46$ DocVQA ANLS when both entries are degraded, and their accuracy is main-limited: $G_D$ never exceeds $0.03$, and every $I_{\mathrm{rec}}$ interval contains zero. Qwen3-VL reports DocVQA and InfoVQA gains from adding DeepStack-style injection in pretraining \citep{qwen2025qwen3vl}, whereas we change the inputs of fixed, post-trained models. Granite's $I_{\mathrm{rec}}$ is positive under ANLS but negative under lowercase exact matching (\DSSuppRef{app:v15}). DeepStack's accuracy looks later-limited, but the lower resolution removes more of its high-resolution later input than of its global main view.

\subsection{Which entry a model can lose differs by model}
\label{sec:coding}
\begin{table}[t]
\centering
\caption{\textbf{The entry that keeps accuracy alone differs by model and replacement.} GQA attribute accuracy (\%) on 1,000 new images (Granite: 999 images). Each entry receives the image's features, zeros (all its features set to zero), or its mean feature over 48 calibration images. Intervals: \DSSuppTableRef{tab:coding_corners}.}
\label{tab:coding}
\begingroup
\fontsize{9}{10.5}\selectfont
\setlength{\tabcolsep}{3.3pt}
\renewcommand{\arraystretch}{1.08}
\DSalign{table_coding-1}
\begin{tabular}{@{}llcccc@{}}
\toprule
\rowcolor{DSTableHeader}
Main input & Later input & Qwen8 & Qwen32 & Granite & DeepStack \\
\midrule
image & image & \DSn{72.50} & \DSn{72.75} & \DSn{73.27} & \DSn{79.50} \\
\addlinespace[3pt]
image & zeros & \DSn{72.20} & \DSn{71.90} & \DSn{51.60} & \DSn{77.85} \\[1pt]
image & mean & \DSn{71.95} & \DSn{72.65} & \DSn{58.81} & \DSn{75.70} \\
\addlinespace[3pt]
zeros & image & \DSn{54.45} & \DSn{67.60} & \DSn{11.66} & \DSn{76.05} \\[1pt]
mean & image & \DSn{50.50} & \DSn{65.40} & \DSn{69.77} & \DSn{29.35} \\
\addlinespace[3pt]
zeros & zeros & \DSn{31.55} & \DSn{43.20} & \DSn{1.00} & \DSn{11.45} \\[1pt]
mean & mean & \DSn{43.10} & \DSn{43.30} & \DSn{42.29} & \DSn{10.35} \\
\bottomrule
\end{tabular}
\endgroup
\end{table}
The recovery patterns of Section~\ref{sec:recovery} predicted five of six single-entry outcomes, and DeepStack can send only its main entry in split inference with about half the bytes.

Heidari et al.\ \citep{heidari2026compressing} code Qwen3-VL's four entry tensors at one shared quantization parameter (QP; larger is coarser) for split inference (Section~\ref{sec:intro}) and leave the split of bytes across entries to future work. We remove an entry's image information by setting its features to zeros or to their mean over 48 calibration images, and we code each entry tensor with NNCodec \citep{becking2023nncodec}, an implementation of the ISO/IEC neural network coding standard, counting bitstream bytes per image. A zeroed entry still runs through the model, so bytes measure features, not latency or memory. We use 1,000 new GQA images with 1,838 attribute questions (Granite: 999 images); accuracies are percentages, and differences are percentage points (\DSSuppRef{app:coding}).

Which entry suffices depends on the model, as in Section~\ref{sec:recovery}, and on the replacement value (Table~\ref{tab:coding}). Relative to the full input, Qwen8 changes by $-0.30$ [$-1.55,0.95$] with later zeroed, Granite by $-3.50$ [$-5.16,-1.85$] with main set to its mean, and DeepStack by at worst $-3.45$ [$-4.80,-2.10$] with either entry zeroed. The replacement matters: zeroing main instead lowers Granite from $69.77\%$ to $11.66\%$ but raises DeepStack from $29.35\%$ to $76.05\%$. Before these images were scored, we predicted which single entry each model can keep (main for Qwen8, later for Granite, and either entry for Qwen32 and DeepStack), allowing each kept entry to cost at most 25\% (Qwen8, Granite) or 50\% (Qwen32, DeepStack) of the gap between the full input and both entries zeroed. With zeros, five of six tested predictions held, and Granite's failed (\DSSuppRef{app:coding_single}).

Sending only DeepStack's main entry roughly halves the bitstream size at comparable accuracy. Main alone at QP~$-4$ (later zeroed) requires 255~KB and scores $77.45\%$, whereas all entries at shared QP~$-2$ require 503~KB and score $76.35\%$: $49.3\%$ fewer bytes at an accuracy difference of $1.10$ [$-0.40,2.55$] points. We fixed this comparison after an exploratory run had shown the same pattern and after the first results on these images, which included uncoded main alone ($77.85\%$; \DSSuppRef{app:coding_deepstack}). Post hoc, coding all entries at a similar size (shared QP~0, 243~KB) drops accuracy to $42.65\%$ (\DSSuppTableRef{tab:coding_deepstack}). A second comparison starts from the scale-normalized setting, which scales each later entry's quantization step with its feature scale (\DSSuppRef{app:coding_design}). A configuration selected on calibration images before testing keeps main at QP~$-2$ and codes later two QP steps coarser than this setting; relative to the scale-normalized setting at the same main QP, it uses $39.8\%$ fewer bytes (278 against 462~KB) and scores $75.80\%$ ($-0.45$ [$-1.40,0.50$] points). At equal main QP, dropping later removes roughly 74\% of the bytes at a cost of $1.45$ to $3.60$ points (\DSSuppTableRef{tab:coding_deepstack_pairs}).

Granite and Qwen8/32 show no coding gain attributable to entry roles. Qwen8/32 do not consistently beat a shared QP at the tested budgets (\DSSuppRef{app:coding_qwen}), and our prediction that Granite, being closer to later-limited, should code its later entries more finely failed (\DSSuppRef{app:coding_role}). A post hoc configuration shifting Granite's bytes from main to later scores $2.64$ [$0.47,4.80$] points above a shared QP at nearly equal bytes on 531 held-out images ($69.30\%$ against $66.67\%$; \DSSuppTableRef{tab:coding_granite}); however, the scale-normalized setting codes later identically and already outperformed the shared QP by $2.46$ [$0.11,4.81$] points on the 468 discovery images.

\section{Discussion and limitations}
\label{sec:discussion}
\textbf{What later entries contribute.}
An entry's place in the architecture does not fix its role in the trained model. Later entries change 63\% of photograph answers about the edited object in DeepStack and none in Qwen8/32. Their influence also depends on main: in Qwen32, the same later stripe features switch 20\% of target answers with clean main and 81\% with mildly blurred main.

\textbf{Using the findings.}
The entry a model needs cannot be read off its architecture: on GQA attributes, Qwen8 needs main, DeepStack needs either entry, and Granite needs later when main is set to its mean. A system that drops or codes entry features, as in split inference, should therefore test its own checkpoint and task; in our tests, the recovery patterns predicted five of six single-entry outcomes on new images.

\textbf{Differences between the designs.}
Feature construction, projectors, group sizes, sampling, and injection depth differ across the designs and covary with training, so we cannot say which of them makes DeepStack follow later and Qwen follow main. Our interventions show how answers depend on the evidence at each entry, not the mechanism behind it, and a stronger later effect under degraded main does not show that a model detects the degradation.

\textbf{Limitations.}
The equal-loss comparisons match only in sample mean, and the stripe-versus-text comparison also alters the stimulus and failure mode. Qwen's direct color accuracy saturates at contrast 0.12, Granite's direct stripe recognition is weak, and several recovery patterns hold only for direct answers. The photograph panel covers red-to-blue edits on 64 images, and GQA and DocVQA evaluations use selected validation subsets; GQA accuracy surviving a 5-pixel long side may reflect language priors or coarse visual cues that we did not separate. Finally, removal and coding results focus primarily on GQA attributes: dropping DeepStack's later entries costs $1.65$ points on these attribute images, compared with $3.91$ points on GQA spatial questions and $9.48$ ANLS points on DocVQA in exploratory Apple-silicon GPU runs (\DSSuppRef{app:coding_deepstack}).

\section{Conclusion}
Supplying mismatched image versions to each visual entry of DeepStack-style models reveals sharp functional divergences among architectures that share one multi-entry design. Later entries decide photograph answers in DeepStack, whereas main decides them in Qwen3-VL; in both families, however, unchanged later features gain influence as main's input degrades. On GQA attribute questions, Qwen8 can drop its later entries and DeepStack can drop either entry, enabling DeepStack to transmit only its main entry in split inference at roughly half the bitstream size. Matching-input accuracy conceals all of these differences, whereas our entry-wise protocol requires no training and applies to any multi-entry vision-language model.

\section*{LLM Usage Statement}
OpenAI Codex and Anthropic Claude Code assisted research planning, code and experiment execution, analysis, figures, literature review, and substantive manuscript drafting and editing. The authors are responsible for all content and results.
{\small
\bibliographystyle{ieeenat_fullname}
\bibliography{references}
}

\clearpage
\makeatletter
\let\Hy@backout\@gobble
\makeatother
\renewcommand{\topfraction}{.9}
\renewcommand{\dbltopfraction}{.9}
\renewcommand{\textfraction}{.07}
\renewcommand{\floatpagefraction}{.8}
\renewcommand{\dblfloatpagefraction}{.8}
\setcounter{dbltopnumber}{2}
\setlength{\emergencystretch}{2em}
\maketitlesupplementary
\noindent Evidence includes two quality experiments: a fixed-contrast experiment (Appendix~\ref{app:v14}) and calibrated degradation curves on held-out units (Appendix~\ref{app:v15}). A coding experiment removes or codes each entry's features on new GQA images and DocVQA documents (Appendix~\ref{app:coding}). This supplement gives intervention and setup definitions, supporting figures, paired analyses of the conflict panels, the quality, recovery and coding results that the main text cites, and further experiments (Appendix~\ref{app:additional}). Complete estimates for every model, task, readout, level and metric of the two quality experiments, and for every configuration and comparison of the coding experiment, are in the CSV files of the accompanying \texttt{supplementary\_data} folder (Appendix~\ref{app:data}). Reference numbers follow the shared bibliography in the main paper.
\appendix
\section{Methods and reproducibility details}
\label{app:methods}
\label{app:reproducibility}

\subsection{Checkpoints and native visual entries}
Throughout, the \emph{main entry} (``main'') is the set of visual tokens at the language-model input, and the \emph{later entries} (``later'') are the visual features added at deeper decoder layers. Table~\ref{tab:checkpoint} identifies the nine fixed checkpoints. All complete the photograph, synthetic local, whole-source (each entry receives all of source $a$ or $b$), and dose (entry-magnitude) panels. Qwen8, Qwen32, and Granite also complete the native-entry decomposition, which intervenes on each checkpoint's own (native) later entries singly and in groups, giving 39 model/panel analyses. We chose this three-model subset before any decomposition result existed, and the quality, removal, and coding experiments use the same three models plus original DeepStack. The five Qwen3-VL models and two Cosmos3 understanding towers form the Qwen lineage: seven checkpoints within one multidepth interface family, not seven independent architectures \citep{qwen2025qwen3vl,nvidia2026cosmosnano,nvidia2026cosmossuper}. Cosmos Nano and Super initialize from Qwen3-VL-8B and 32B and continue training the vision encoder, projectors, and language model \citep{nvidia2026cosmos3}; their system names do not specify the loaded understanding tower's parameter count.

\begin{table}[!htb]
\centering\footnotesize
\setlength{\tabcolsep}{3pt}
\caption{Fixed checkpoint identities. Complete revision hashes accompany the experiment configurations.}
\begin{tabular}{@{}ll@{}}
\toprule
\rowcolor{DSTableHeader}
Hub identifier & Revision prefix \\
\midrule
\texttt{Qwen/Qwen3-VL-2B-Instruct} & \texttt{89644892e4d8} \\
\texttt{Qwen/Qwen3-VL-4B-Instruct} & \texttt{ebb281ec70b0} \\
\texttt{Qwen/Qwen3-VL-8B-Instruct} & \texttt{0c351dd01ed8} \\
\texttt{Qwen/Qwen3-VL-32B-Instruct} & \texttt{0cfaf48183f5} \\
\texttt{Qwen/Qwen3-VL-30B-A3B-Instruct} & \texttt{9c4b90e1e4ba} \\
\texttt{nvidia/Cosmos3-Nano} & \texttt{e59a53c25979} \\
\texttt{nvidia/Cosmos3-Super} & \texttt{f543c56225b2} \\
\texttt{ibm-granite/granite-vision-4.1-4b} & \texttt{37d591f06319} \\
\texttt{menglc/deepstack-l-vicuna-7b} & \texttt{76f7d20cd835} \\
\bottomrule
\end{tabular}
\label{tab:checkpoint}
\end{table}

Qwen3-VL-2B/4B take intermediate vision features from encoder indices 5, 11, and 17, and the remaining Qwen/Cosmos towers use 8, 16, and 24. Separate learned mergers produce later representations added after decoder blocks 0, 1, and 2, while main uses the final vision representation and the output grids align. Granite maps vision indices $-1,-7,-13,-19$ to decoder entries 0, 3, 6, and 9 \citep{ibm2026granitevision}; entries 12, 15, 18, and 21 use final-layer features with four phase-seeded Q-Former branches. Each branch compresses a $24\times24$ grid to $12\times12$; within an $8\times8$ window, a phase changes query initialization but cross-attention can read the whole window.

Original DeepStack uses CLIP hidden-state index $-2$ and a shared projector for its global $336\times336$ main view and its dense spatial groups \citep{meng2024deepstack}. Four later groups enter at decoder layers 1--4: square photograph inputs activate all four, whereas rectangular synthetic inputs activate two and retain two officially zero-padded groups, and each active group has 576 tokens. ``All later'' always includes every native later entry of the checkpoint. These comparisons preserve native geometry and jointly vary feature depth, spatial sampling, projector, token count, entry count, decoder depth, and training, so they do not isolate architecture as a causal factor. Inference uses BF16 CUDA with architecture-specific adapters, and Appendix~\ref{app:v14} reports the batching checks of the quality experiments. The photograph and controlled panel experiments ran on single NVIDIA A100 GPUs.

\textbf{Three implementations of DeepStack-style injection.}
The nine checkpoints implement multi-layer injection in three independent ways. Qwen3-VL names its later entries DeepStack \citep{qwen2025qwen3vl}, and Granite's model card describes its design as ``a variant of Deepstack'' in which visual features are added to the language-model hidden states at multiple layers \citep{ibm2026granitevision}. The three differ in what the later entries add to main. In original DeepStack, main is one 336-pixel global view, whereas the later groups take 336-pixel crops from a 672-pixel canvas (up to 1008 pixels on DocVQA), each view having 576 tokens. On square photographs the later entries therefore carry four times main's tokens at twice its linear sampling density. By contrast, Qwen's later entries share main's grid and token count and differ in encoder depth and merger, and Granite's intermediate entries likewise read main's grid at other encoder depths. Granite's spatial entries read main's final-layer features and grid through the Q-Former branches described above, so they differ from main in query phase, branch weights, and injection depth. Only DeepStack's later entries thus add tokens or input resolution.

\subsection{Photographs, synthetic scenes, and intervention support}
The photograph panel contains 64 images from fixed 2019 GQA scene graphs and Visual Genome sources \citep{hudson2019gqa,krishna2017visualgenome}, excluding 320 photographs used during method development. The development experiments used other materials, and Appendix~\ref{app:additional} reports the informative ones without pooling any with the main panels. Selection and visual review never inspect model responses. Each image has two checked referents (the objects a query asks about) with nonoverlapping boxes and is fitted, without object cropping, to a $384\times384$ gray canvas. Within each box, a fixed HSV rule selects the largest connected color component, and recoloring changes its hue to red or blue while preserving saturation, value, and all other pixels. This is controlled recoloring rather than instance segmentation. Editing each object gives 128 edits and 256 target/control queries, always directed red-to-blue, and the queries identify the referent without naming its tested color.

The pixel-edit mask differs from the intervention's support: the selected visual tokens (the region of interest, ROI) follow the object box and can include surrounding context. Backgrounds are original or horizontally mirrored photographs ($M0/M1$), blended toward gray at strength $\alpha=1$ (image intact) or $.08$. Mirroring need not move an object onto the other referent's coordinates. Raw interventions (full feature replacement, $k=1$) cover both backgrounds and both strengths, whereas dose matching to a native target (Appendix~\ref{app:dose}) uses only $M0$ and $\alpha=1$.

The controlled panel (paired synthetic scenes) uses a fixed seed and 64 templates: two shape pairs, two layouts, two assignments, and eight replicates. Color (red/blue) and stripe (vertical/horizontal) versions share geometry on a $384\times256$ canvas, and stripes have period 14 and bar width 6. Each object is edited in turn, with target and control queries. Local conditions cross entry group, original/opposite destination, $M0/M1$, and $\alpha\in\{.08,1\}$, where attenuation blends object texture toward gray while preserving shape and background. $M1$ exchanges equal image halves pixel for pixel, with no extra shape transformation. Across layouts, the local donor features (copied in from the other source) and their mask at a given destination stay fixed, while the scene supplying main and the unselected later tokens change. The layout contrast therefore tests current object--position correspondence without isolating main's contribution to it. Moving an intervention transports its contextualized features and token mask together.

Whole-source conditions fix both source strengths at $\alpha=1$ and use four layout pairs and all main/later source assignments. Dose and native-entry decomposition use a 32-template subset. A deterministic hash ranking, balanced across shape, layout, and assignment, selects it without using outcomes. Dose retains full-source support, $M0/M0$, and both fixed-source assignments (the unchanged entry holds $a$ or $b$), but decomposition crosses layouts and strengths. All repeated edits, queries, formats, and conditions are clustered by their shared photograph or template.

\subsection{Readouts, pairing, and intervals}
\label{app:statistics}
For the edited object's directed $a\!\rightarrow\!b$ attribute change, the score is $y_q^{\score}=\log p(c_b\mid q,z)-\log p(c_a\mid q,z)$ and valid generation is $y_q^{\gen}=\mathbf{1}[\widehat c_q=c_b]$. We retain this direction for control queries and for reversed numeric mappings (cyclic labels). Paired source effects are $E_q=y_q(z_b)-y_q(z_a)$, relative selectivity is $R_g=E_{\mathrm{target},g}-E_{\mathrm{control},g}$, and the entry contrast is $\Delta R=R_{\main}-R_{\later}$. Generation effects are net changes toward $b$, generally neither accuracy nor total transition rates, and a positive $R$ can reflect target increases, control decreases, or both.

Direct-answer parsing normalizes case and surrounding whitespace and permits an optional final period, but requires a complete candidate phrase. Numeric parsing requires one unambiguous legal label, with options (1)/(2) shown in identity or reversed (cyclic) order. In the conflict panels, generation is greedy with a 32-token limit, and invalid, truncated, missing, and unmatched outcomes remain unavailable. Native correctness (accuracy on the unedited source images) is a separate diagnostic, not an inclusion filter. Appendix~\ref{app:v14} specifies the accuracy policies of the quality experiments and their 64-token benchmark setting.

Candidate scoring sums the declared continuation's token log probabilities, without length normalization or an appended end-of-sequence (EOS) token. It therefore measures a narrower event than the generation parser accepts: a lower-case phrase excludes its capitalized spelling, and a phrase and that phrase plus a period can overlap as continuation events, so their probabilities cannot simply be summed into mutually exclusive semantic answers. Alternative complete-answer scoring has not been run, and raw score scales are not pooled across models, attributes, or formats.

Contrasts are formed within edits and averaged over both edited objects only for complete cases, those whose required observations are all valid. Each estimate reports its complete-case $n$; single-cell denominators cannot replace a paired contrast's intersection. Conflict-panel intervals use $\bar x\pm1.96s/\sqrt n$, quality-experiment synthetic and photograph estimates use Student-$t$ quantiles, and benchmark estimates retain the scorer's normal approximation. All are descriptive 95\% intervals without multiplicity adjustment, estimated from paired unit differences rather than aggregate interval endpoints. Zero-containing intervals do not establish equivalence, and zero standard errors describe the observed sample.

\subsection{Fixed source tables and role exchange}
\label{app:source_tables}
For fixed physical main/later layouts and strengths, let $y_{uv}$ denote the response when main receives source $u$ and later receives source $v$. We distinguish
\begin{equation}
 D_{\mathrm{fixed}}=y_{ba}-y_{ab},\qquad
 J=y_{bb}-y_{ba}-y_{ab}+y_{aa}.
 \label{eq:app_source_table}
\end{equation}
The first compares conflicting assignments, and the second measures output nonadditivity. Subtracting entry effects with the other source fixed to $a$ or to $b$ gives the same $D_{\mathrm{fixed}}$ algebraically, so the two versions are not replications. In a role exchange, the changing source's layout and strength follow its identity as it moves between entries. Mixed-layout exchanges span different physical tables, so their $D_{\mathrm{swap}}$ cannot replace $D_{\mathrm{fixed}}$, $J$, or the query-relative $\Delta R$.

\subsection{Operational dose and native-entry decomposition}
\label{app:dose}
A selected entry installs $v_e^a+k(v_e^b-v_e^a)$ on its specified tokens; $k=1$ replaces source features and $k>1$ extrapolates. Equal replacement need not give equal magnitudes across entries, so we compare each entry's magnitude, called its \emph{dose} here (the entry magnitude of the main text). For each edited object and role/context, a canonical (fixed reference) forward pass on a direct target query measures the post-entry residual $h_e^a$ on the $a$ side. On selected tokens, define
\begin{equation}
 \begin{aligned}
 r_e&=\frac{\operatorname{RMS}(v_e^b-v_e^a)}{\operatorname{RMS}(h_e^a)},\\
 \rho_{\mathrm{RSS}}(G)&=\sqrt{\sum_{e\in G}r_e^2},\qquad
 \rho_{\mathrm{L1}}(G)=\sum_{e\in G}r_e.
 \end{aligned}
 \label{eq:app_dose}
\end{equation}
L1 is a sum of entry RMS ratios, not an elementwise tensor norm, and RSS assumes no cross-layer orthogonality. A main-native target keeps main raw and scales later to main's native dose (its own dose at $k=1$), and a later-native target reverses these roles. Target/control queries and answer mappings reuse the selected tokens, canonical denominators, and matching multiplier. The delivered dose is checked at inference precision, and unmatched cases remain unavailable. Matching equates the declared measure, not semantic information or every query's residual normalization. Since L1 exceeds RSS for multi-entry groups, it lowers later's multiplier at the main-native target and raises main's at the later-native target. The two measures are therefore different physical interventions, and their raw anchors must remain visible.

Decomposition starts from a shared baseline $B$, which installs donor-$a$ ROI features at every entry over a fixed background; only the selected group then receives its increment. One complete forward pass on $B$ supplies all denominators, and the shared-low target is the minimum native singleton ratio across main and all later entries. A group's multiplier is that target divided by its RSS or L1 aggregate. Qwen8/32 test three singletons, their three pairs, and all later entries. Granite tests seven singletons, the triples $(3,6,9)$, $(15,18,21)$, and $(3,12,21)$, and all seven later entries, where the spatial triple $(15,18,21)$ excludes layer 12.

At raw strength only, the composition residual $C=R_{\mathrm{all\ later}}-\sum_eR_{\{e\}}$ compares identical joint and singleton increments from $B$. For Granite, $G=R_{\{15,18,21\}}-R_{\{3,6,9\}}$ compares equal-count spatial and intermediate subsets, and $H=G_{\mathrm{stripe}}-G_{\mathrm{color}}$ is their paired attribute interaction. A separate experiment measures $M_{21\mid15,18}=R_{\{15,18,21\}}-R_{\{15,18\}}$ on the fixed-contrast templates, keeping entries 15 and 18 fixed. Neither $C$ nor this conditional marginal assigns additive neural contributions, and matched groups, whose multipliers differ, are not used for this additive contrast. Qwen32's direction-aligned layout comparison (the $M1$ residual multiplied by $-1$ to align spatial direction) is $-C(M1)-C(M0)$, whether or not its components separately reverse sign with layout. These output contrasts do not establish neural synergy or separate feature type from decoder depth and dose.

\subsection{Intervention specification}
\label{app:spec}
Local transfers cross original and exchanged layouts ($M0/M1$) with original and opposite destinations ($o/p$). The spatial interaction
\begin{equation}
 I_g=[R_g(M1,p)-R_g(M1,o)]-[R_g(M0,p)-R_g(M0,o)]
 \label{eq:app_spatial}
\end{equation}
tests whether destination effects change with the object currently occupying that position, and the entry difference $I_{\main}-I_{\later}$ is paired directly. Table~\ref{tab:intervention_spec} lists, for each experiment family, the source identity, intervened entries, spatial scope, dose, readouts and inference unit. Designed contrasts vary specified factors within a family. These factors are the entry set for $\Delta R$, the multiplier for matched $\Delta R$, scene layout and destination for $I$, main quality for $I_{\mathrm{quality}}=R_{\later}(D)-R_{\later}(C)$, main and later quality for $G_D$ and $I_{\mathrm{rec}}$, the readout for cross-format differences, and the checkpoint for Qwen8 versus Qwen32 at one level. By contrast, comparisons across families, such as photograph conflict against synthetic recovery or stripes against text, change several elements at once and are descriptive.

\begin{table*}[tp]
\centering\scriptsize
\setlength{\tabcolsep}{2.5pt}
\renewcommand{\arraystretch}{1.12}
\caption{Intervention specification of each experiment family. $a/b$: sources differing only in the edited object's attribute; C/D: clean/degraded versions of $a$. ROI: visual tokens overlapping the edited object's box, one index set at every changed entry (DeepStack: one set per crop group); ``all'': every visual position. Readouts D/I/C: direct, identity-label, cyclic-label; kinds S/G: candidate score and greedy generation. In both quality experiments the conflict $a$ endpoint is the same visual input as recovery cell CC (clean main) or DC (degraded main), and conflict cells never degrade later input.}
\label{tab:intervention_spec}
\begin{tabularx}{\textwidth}{@{}>{\raggedright\arraybackslash}p{2.0cm}>{\raggedright\arraybackslash}X>{\raggedright\arraybackslash}p{2.0cm}>{\raggedright\arraybackslash}p{2.3cm}>{\raggedright\arraybackslash}p{2.1cm}>{\raggedright\arraybackslash}p{1.6cm}>{\raggedright\arraybackslash}p{1.8cm}@{}}
\toprule
\rowcolor{DSTableHeader}
Family (estimands) & Source identity & Entries intervened & Spatial scope & Dose / multiplier & Readouts, kinds & Unit \\
\midrule
Photo conflict ($R$, $\Delta R$) & $a/b$: referent recolored red/blue, clean; unchanged entry and non-ROI tokens show the original or mirrored photo at $\alpha\in\{1,.08\}$, edited object in its native color & main alone, or all native later (Qwen 3, Granite 7, DeepStack 4) & ROI, fixed donor box (Qwen: median 6\% of tokens) & raw $k\in\{0,1\}$; at $M0$, $\alpha{=}1$ also RSS/L1 opposite-native $k$ & D/I/C; S+G & photograph (64) \\
Local transfer ($I$, $I_{\main}-I_{\later}$) & donors $a/b$ ($M0$, $\alpha{=}1$), clean, over the source-$a$ scene at $M0/M1$, $\alpha\in\{1,.08\}$ & main alone, or all native later & ROI at the original or opposite half; features and mask move together & raw replacement & D/I/C; S+G & scene or template (64) \\
Whole source ($D_{\mathrm{fixed}}$, $J$) & main and later each $a$ or $b$ (four assignments), clean; four layout pairs; $\alpha{=}1$ & main and all later, jointly & all & raw & D/I/C; S+G; target query only & template (64) \\
Synthetic dose (matched $\Delta R$) & changed entry $a\rightarrow b$; other entry fixed to $a$ or $b$; $M0/M0$, $\alpha{=}1$ & main alone, or all native later & all & raw $k\in\{0,1\}$; RSS/L1 opposite-native $k$ & D/I/C; S+G & template (32) \\
Groups ($R_G$, $C$, $G$, $H$) & shared $B$: donor-$a$ ROI at every entry over the source-$a$ scene ($M0/M1$, $\alpha\in\{1,.08\}$); the selected group moves toward clean $b$ & singletons, pairs or triples, all later & ROI & raw $k{=}1$; shared-low RSS/L1 & D/I/C; S+G & template (32) \\
Granite marginal ($M_{21\mid15,18}$) & main and later base clean $a$; donor clean $b$ (fixed-contrast templates) & entry 21 added to fixed $\{15,18\}$ & ROI & $k{=}1$ & D/I/C; S+G & template (32) \\
Fixed-contrast conflict ($R_{\later}(C)$, $R_{\later}(D)$, $I_{\mathrm{quality}}$) & main: C or D (contrast .12) of $a$; later: clean $a$ with ROI replaced by clean $b$ at $k{=}1$ & all native later & ROI; non-ROI later tokens stay clean $a$ & $k_e{=}1$ (exact replacement) & D/I/C; S+G & template (32 per task); photograph (64) \\
Fixed-contrast recovery ($A_{uv}$, $G_D$, $I_{\mathrm{rec}}$) & one image $a$, C or D separately at main and later; no $b$; D: contrast .12 (synthetic, photos) or blur radius 2 (GQA, DocVQA) & main and all later, as whole inputs & all & none & D/I/C accuracy (S, G); free-form GQA exact, DocVQA ANLS & template (32 per task); photograph (64); GQA image (256 per group); document (256) \\
Calibrated conflict (as fixed-contrast) & as fixed-contrast; D is a calibrated level (desaturation, relative blur, lower resolution) & all native later (DeepStack: 2 active groups) & ROI (median 19--26\% of later tokens); non-ROI later tokens stay clean $a$ & $k{=}1$ & D/I/C; S+G & template (32 per task); synthetic only \\
Calibrated recovery (as fixed-contrast) & as fixed-contrast at each calibrated level; CC shared across levels & main and all later, as whole inputs & all & none & D/I/C accuracy (S, G); free-form GQA exact, DocVQA ANLS & template (32 per task); GQA image (192 per group); document (192) \\
Removal and coding (accuracy, bytes; Appendix~\ref{app:coding}) & one image $a$, no $b$; each entry unchanged, zeros, calibration mean, or coded and decoded by NNCodec & main, all native later, or both; each entry tensor separately & all & none; one QP per entry & accuracy (G); free-form GQA exact, DocVQA ANLS & GQA image (1,000; Granite 468 and 531); document (474) \\
\bottomrule
\end{tabularx}
\end{table*}

\section{Quality checks and sensitivity analyses}
\label{app:quality}
\subsection{Unavailable responses and native capability}
In the photograph panel, Qwen8 and Qwen32 have 37 and 11 invalid generation records, so their reference paired contrasts (the $M0$, $\alpha=1$, direct-generation, raw-dose cell) keep 62 and 63 photographs, versus Granite's 64. Some weak-background local color interactions keep only 34 and 53 complete templates, respectively, but the full-strength stripe interaction has all 64 in each of the three models. These counts differ because each contrast requires different observations, not because the planned sample changed.

For Qwen8 and Qwen32, the entry decomposition contains all 78,336 planned observations in each answer measure (score and generation). Qwen8 has 217 unparseable, EOS-terminated generations and no recorded truncation, so its cell and paired-contrast counts range over 20--32 and 19--32 templates. Qwen32 has 319 truncated responses plus one unparseable EOS-terminated response, giving ranges of 24--32 and 23--32. Score estimates and every raw/shared-low reference in Table~\ref{tab:p4} have all 32 templates, but this completeness does not extend to the whole grid.

Nano full-source and DeepStack local outcomes have no missing, invalid, truncated, or nonfinite records ($n=64$), and Super full-source and both Cosmos dose panels likewise keep all 64 and 32 templates, respectively. Super's photograph panel has 37 invalid generations, no missing, truncated, or unmatched records, and $n=62$--64 (reference $n=63$). Nano and Super local color have 852 and 1,032 invalid records among 29,184 each, without truncation, so ordinary effects use $n=6$--64 and 8--64 and spatial contrasts use 0--64. At $\alpha=.08$ with direct answers, neither model has a complete template for the paired entry interaction (Nano's separate main and later interactions have $n=1/6$, and Super has neither), while at full strength the color comparisons are complete and local stripe generation and all local scores have 64 throughout.

For both Cosmos checkpoints, full-strength native color is correct and valid on all 128 queries in all formats, and native stripe correctness is 98/81/87 for Nano and 114/115/116 for Super under direct/identity/cyclic readouts, each out of 128 repeated queries from 64 templates. DeepStack's dose panel has 34 invalid color and 32 invalid stripe generations, including 13 truncated stripe outputs under amplification, so affected estimates use $n=28$--31. The Qwen8, Qwen32 and Granite dose panels have no invalid, truncated, or unmatched observations, and their delivered doses deviate from the canonical dose by at most 0.60\% for Granite, 0.19\% for Qwen8, and 0.08\% for Qwen32. This checks the declared anchor, not every query-specific normalization.

\subsection{Photograph exclusion and answer transitions}
In photograph \texttt{vg-2389325}, the referring expression and the bounding box are inconsistent, so a post hoc analysis excludes the whole photograph for Qwen2/4/30, Nano, Super, and DeepStack. The exclusion preserves the principal entry-ordering conclusions and leaves the original results unchanged, although one Qwen2 direct-generation $R$ interval newly contains zero and two Qwen4 control intervals newly exclude zero. For Super, reference $\Delta R$ changes from $0.75$ [$0.67,0.82$], $n=63$, to $0.75$ [$0.67,0.83$], $n=62$. All 24 entry comparisons remain positive in each measure, but two control-related intervals newly contain zero and one near-zero control-score mean changes sign, so invariance of the individual components is not implied.

In the reference condition, Table~\ref{tab:reference_transitions} counts answer changes on each model's shared complete-photograph set. Dividing its counts by their denominators gives the main and later target and control means, which reconstruct the paired contrasts before rounding; separately complete entry columns can use different sets.

\begin{table*}[t]
\centering\footnotesize
\caption{\textbf{Main changes Qwen answers and later changes DeepStack answers.} Granite's answers change with both entries, and controls almost never change. Counts use each model's shared complete-photograph set. Query pairs and photographs are different denominators: each photograph supplies two target-query pairs and two control-query pairs. All observed changes are toward $b$.}
\DSalign{appendix-1}
\begin{tabular}{@{}llrrr@{}}
\toprule
\rowcolor{DSTableHeader}
Model & Entry & Target changes & Control changes & Photos with target change \\
\midrule
Qwen8 & Main & \DSat{81}{/124} & \DSat{1}{/124} & \DSat{55}{/62} \\
Qwen8 & Later & \DSat{0}{/124} & \DSat{0}{/124} & \DSat{0}{/62} \\
Qwen32 & Main & \DSat{93}{/126} & \DSat{1}{/126} & \DSat{59}{/63} \\
Qwen32 & Later & \DSat{0}{/126} & \DSat{0}{/126} & \DSat{0}{/63} \\
Granite & Main & \DSat{24}{/128} & \DSat{1}{/128} & \DSat{22}{/64} \\
Granite & Later & \DSat{34}{/128} & \DSat{0}{/128} & \DSat{28}{/64} \\
DeepStack & Main & \DSat{6}{/128} & \DSat{0}{/128} & \DSat{6}{/64} \\
DeepStack & Later & \DSat{80}{/128} & \DSat{4}{/128} & \DSat{54}{/64} \\
\bottomrule
\end{tabular}
\label{tab:reference_transitions}
\end{table*}

Qwen8/32 also show no later changes among all 128 target and 128 control pairs that are complete for the later-entry intervention alone, but under the reported matched RSS condition their later target changes number one and seven. For unavailable reference answers, we assign each binary endpoint 0 or 1, consistently across its repeated uses. Over the full 64 photographs, this gives $\Delta R$ extrema $[0.64,0.66]$ and $[0.73,0.73]$ ($[0.641,0.664]$ and $[0.727,0.734]$ to three decimals), which are deterministic point-estimate bounds, not sampling intervals.

\subsection{Control composition and room to move}
\label{app:control_room}
Both queries are scored toward $b$, so a control whose answer is already $b$ cannot move toward $b$ in generation. All 128 photograph edits run red to blue; 54 unchanged referents are red and 74 blue (34 red--blue, 20 blue--blue and 10 red--red photographs), so Table~\ref{tab:control_room} restricts changes to queries that answer red with source $a$ and therefore have room to move. Even these controls rarely change: at most 4 of 63, for DeepStack's later entry, and each such change is a blue control misread as red at baseline. Across models, only one natively red control (one answered red without intervention) changes (Qwen8 main, 1 of 52). This rules out the explanation that controls were all at the generation ceiling, but equal starting answers need not mean equal margins, so it does not exclude a shared bias.

Candidate scores can move on both queries, and in the reference cell they show the same contrast. For Qwen8 and Qwen32 main, control effects are $-0.05$ [$-0.28,0.18$] and $0.23$ [$-0.02,0.47$] against target effects of $18.20$ and $21.63$. For DeepStack main and later, they are $0.12$ [$0.06,0.18$] and $0.40$ [$0.24,0.57$] against $1.09$ and $3.37$. However, subtraction cancels a shared shift only if it moves both queries equally.

Restricted to photographs whose control is red, $\Delta R$ is $0.51$ [$0.36,0.66$], $0.56$ [$0.41,0.71$], $-0.52$ [$-0.67,-0.38$] and $-0.16$ [$-0.29,-0.03$] for Qwen8, Qwen32, DeepStack and Granite (edit-level complete cases). With blue controls it is 0.71, 0.81, $-0.54$ and $-0.01$, so the lower Qwen values with red controls come from the target component. When red targets and red controls start from nearly equal log odds, targets change in 40 of 52, 42 of 52 and 34 of 53 cases (Qwen8 main, Qwen32 main, DeepStack later) against 1 of 52, 0 of 52 and 0 of 53 controls, and photograph-level score effects are 19.37, 24.52 and 4.02 for targets against $-0.19$, 0.34 and 0.29 for controls.

DeepStack's direct generations are capitalized, whereas its candidate scores use lower-case words, so we rescored 8 paired queries from 2 photographs using capitalized candidates. This raised the candidate probability mass from $<0.001$ to $0.998$ and kept all 88 $b-a$ signs, all 16 selectivity signs and all 8 main-minus-later signs.

The synthetic sets differ from the photographs in this respect. In every template of the controlled panel and of both quality experiments, the two objects take different values, and the edit gives the target the control's value. All 256 control rows per set therefore have correct answer $b$, and source $b$ always shows two objects sharing the queried value. A correctly answered control can then move only away from $b$ in generation, so generation control components cannot reveal a shared shift toward $b$, while candidate scores can. In the calibrated color task, for example, DeepStack's direct control effect is $0.42$ against a target effect of $1.32$ with clean main and $1.80$ against $4.55$ at the strongest desaturation. The local-transfer reversals are unaffected, because there the moved features land on the control and its answers do change (Appendix~\ref{app:figures}). Controls whose value equals $a$ would require other materials.

\begin{table*}[t]
\centering\footnotesize
\setlength{\tabcolsep}{5pt}
\renewcommand{\arraystretch}{1.04}
\caption{\textbf{Controls that can still move toward blue rarely do.} Answer changes toward blue in the photograph reference cell (raw strength, $M0$, full background, direct generation, shared complete photographs). ``At risk'' counts queries that answer red at the $a$ endpoint of that entry's intervention, where the other entry shows the original photograph, and so can still move toward blue (the counts therefore differ between entries); ``all'' repeats the full denominators of Table~\ref{tab:reference_transitions}. No answer moves from blue to red. Post hoc counts from the paired endpoints.}
\label{tab:control_room}
\DSalign{appendix-2}
\begin{tabular}{@{}llrrrrr@{}}
\toprule
\rowcolor{DSTableHeader}
Model & Entry & Photos & Target, at risk & Target, all & Control, at risk & Control, all \\
\midrule
Qwen8 & Main & \DSn{62} & \DSat{81}{/95} & \DSat{81}{/124} & \DSat{1}{/54} & \DSat{1}{/124} \\
 & Later &  & \DSat{0}{/54} & \DSat{0}{/124} & \DSat{0}{/54} & \DSat{0}{/124} \\
Qwen32 & Main & \DSn{63} & \DSat{93}{/107} & \DSat{93}{/126} & \DSat{1}{/56} & \DSat{1}{/126} \\
 & Later &  & \DSat{0}{/56} & \DSat{0}{/126} & \DSat{0}{/56} & \DSat{0}{/126} \\
Granite & Main & \DSn{64} & \DSat{24}{/77} & \DSat{24}{/128} & \DSat{1}{/59} & \DSat{1}{/128} \\
 & Later &  & \DSat{34}{/92} & \DSat{34}{/128} & \DSat{0}{/58} & \DSat{0}{/128} \\
DeepStack & Main & \DSn{64} & \DSat{6}{/65} & \DSat{6}{/128} & \DSat{0}{/62} & \DSat{0}{/128} \\
 & Later &  & \DSat{80}{/107} & \DSat{80}{/128} & \DSat{4}{/63} & \DSat{4}{/128} \\
\bottomrule
\end{tabular}
\end{table*}

\section{Supporting figures and source-conflict results}
\label{app:figures}
The following figures cover the three models of the native-entry group analyses (Qwen8, Qwen32, and Granite) and show component, readout, and dose differences.

\begin{figure*}[t]
\centering\includegraphics[width=\textwidth]{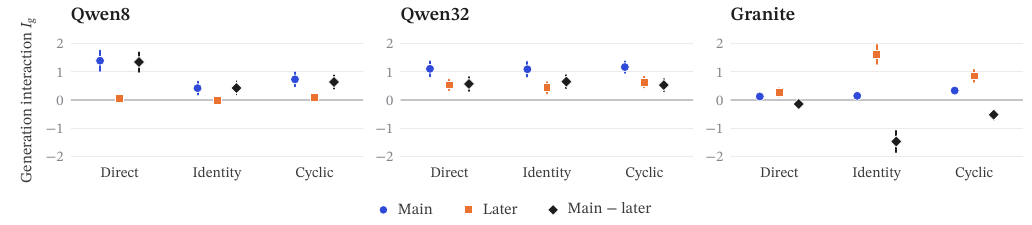}
\caption{\textbf{Stripe spatial interactions favor main in Qwen8 and Qwen32, and later in Granite under label readouts.} Generation interaction $I_g$ of Eq.~\eqref{eq:app_spatial} at full background strength under direct answers and two label formats, with 64 complete templates. The main and later interactions and their directly paired entry difference are separate estimates.}
\label{fig:stripes}
\end{figure*}

\begin{figure*}[t]
\centering\includegraphics[width=\textwidth]{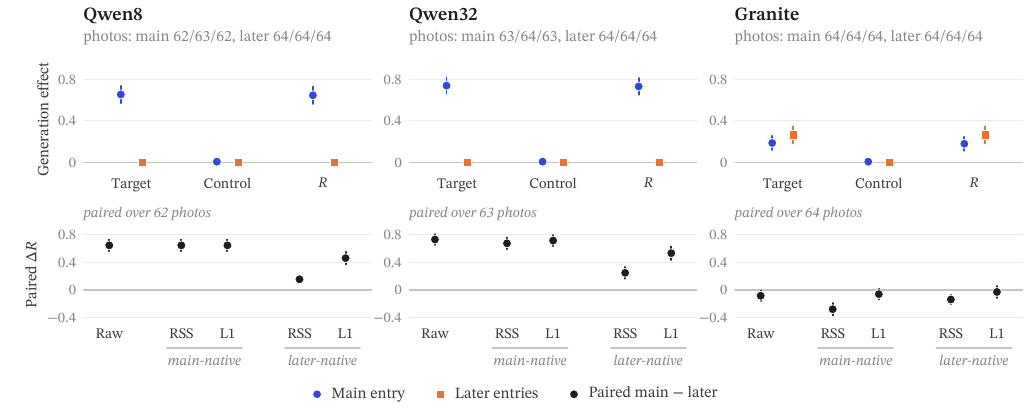}
\caption{\textbf{On photographs, Qwen8 and Qwen32 favor main at every dose, while Granite's contrast is near zero or negative.} Generation components and dose controls for Qwen8, Qwen32 and Granite: target and control effects, relative selectivity $R$ and the directly paired entry difference $\Delta R$, with direct answers, $M0$ and $\alpha=1$. Here $R$ is the target minus the control effect and $\Delta R=R_{\main}-R_{\later}$. Counts are complete photographs for the target, control and $R$ estimates, not repeated intervention records. Table~\ref{tab:natural} covers all nine checkpoints.}
\label{fig:naturalgencomponents}
\end{figure*}

\begin{figure*}[t]
\centering\includegraphics[width=\textwidth]{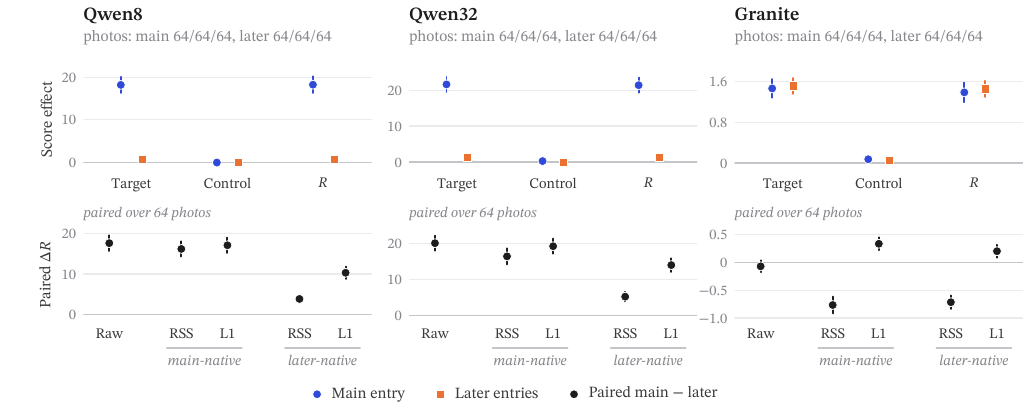}
\caption{\textbf{Candidate scores give the same Qwen ordering, whereas Granite's score contrast is negative under RSS matching and positive under L1.} Score counterpart of Figure~\ref{fig:naturalgencomponents}, with model-specific score scales: target, control and relative effects, then the paired entry contrast. Granite's L1 generation contrast remains uncertain. The same photographs underlie the repeated dose settings.}
\label{fig:naturalscore}
\end{figure*}

\begin{figure*}[t]
\centering\includegraphics[width=\textwidth]{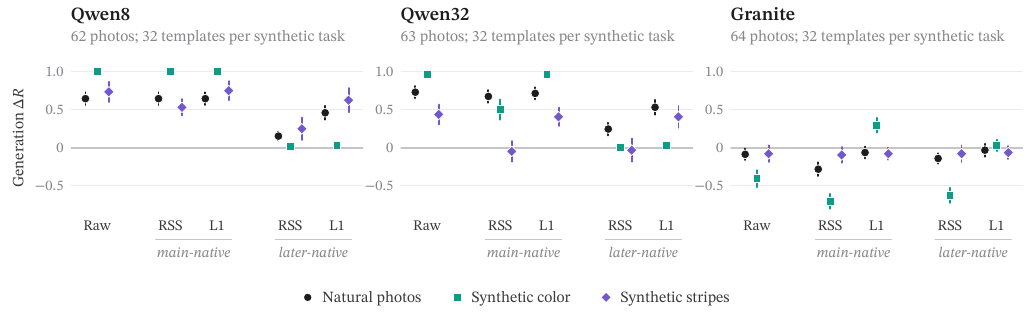}
\caption{\textbf{Later-native matching removes the Qwen main advantage on synthetic color but not on photographs.} All 45 three-model generation reference points: raw dose and both native targets under RSS/L1 for photographs, synthetic color, and synthetic stripes. Synthetic comparisons use 32 templates, $M0/M0$, and fixed source $a$, while photograph comparisons use their actual complete-photograph counts. Material, intervention support, and sample set differ across materials, so their amplitude differences do not isolate a causal effect of material type.}
\label{fig:doseall}
\end{figure*}

At full strength, original layouts, and direct color generation, target-query $D_{\mathrm{fixed}}$ is $0.91$ [$0.86,0.97$] for Qwen30 and $-0.79$ [$-0.85,-0.73$] for DeepStack ($n=64$). Nano's fixed-$a$ generation $\Delta R$ is $1.00$ [$1.00,1.00$] for color and $0.59$ [$0.47,0.70$] for stripes. That stripe value combines $R_{\main}=0.53$ [$0.41,0.65$] and $R_{\later}=-0.05$ [$-0.16,0.05$], whereas the corresponding $D_{\mathrm{fixed}}$ is $0.80$ [$0.74,0.86$]. Super's color $\Delta R$ is also $1.00$ [$1.00,1.00$], and its stripe $\Delta R=0.35$ [$0.26,0.44$] differs from $D_{\mathrm{fixed}}=0.41$ [$0.32,0.51$] because control effects contribute to selectivity. These are 64-template estimates, and saturated color intervals do not imply certainty outside the sample. With mixed layouts, 21 of DeepStack's 24 role-selectivity intervals include zero despite later effects on both queries. Fixed-source influence, role exchange, and relative selectivity thus remain distinct (Appendix~\ref{app:source_tables}).

\section{Additional native-entry comparisons}
\label{app:groups}
\subsection{Composition and direction alignment}
\begin{figure*}[t]
\centering
\includegraphics[width=\textwidth]{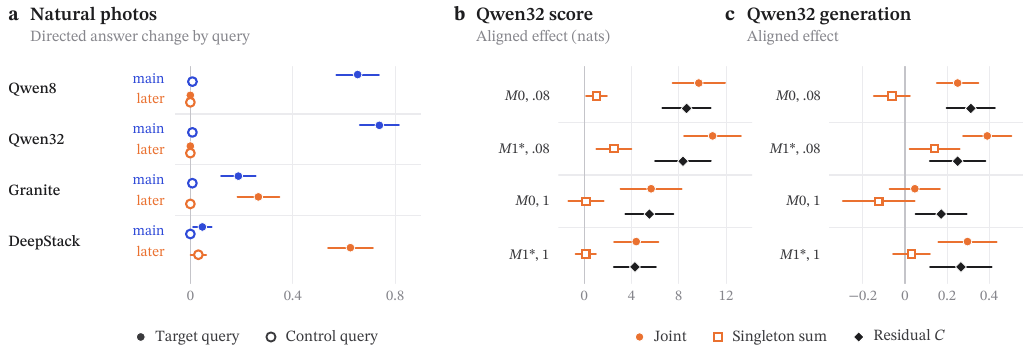}
\caption{\textbf{Strong object-directed changes and nonadditive joint effects.} (a) Photograph panel, $M0$, $\alpha=1$, direct generation, $k=1$: target and control components use the same complete photographs within each model (62/63/64/64 for Qwen8/Qwen32/Granite/DeepStack). Each later point intervenes on the checkpoint's whole native later-entry collection. The strong Qwen-main and DeepStack-later effects mostly come from the target query. (b,c) Qwen32 stripes, direct readout, raw dose, 32 templates: joint later effect, sum of the three singleton effects, and their directly paired composition residual $C$. Rows give layout and background strength. For $M1$ (marked *), all plotted effects and interval endpoints are multiplied by $-1$ to align spatial direction; this does not assert layout equivalence. Bars are descriptive normal 95\% intervals. Generation is a directed answer change, not task accuracy.}
\label{fig:entry_components}
\end{figure*}
Figure~\ref{fig:entry_components} shows the shared-photo target/control components and the joint, singleton-sum, and aligned residual estimates. Table~\ref{tab:composition} reports Qwen32's raw composition residual for stripes, and Table~\ref{tab:composition_aligned} reports its paired layout difference after direction alignment. The residual reverses sign with layout, and all aligned-difference intervals include zero, without establishing equivalence. For $\alpha=.08/1$, the aligned joint-generation differences are $0.14$ [$0.02,0.26$] and $0.25$ [$0.10,0.40$], and the singleton-sum differences are $0.20$ [$0.04,0.37$] and $0.16$ [$0.01,0.31$]. The residual is the difference between these two changes. Later entry 2, by contrast, has differences $-0.02/0$, so neither invariant components nor a changed neural combination rule follows from the residual alone. Qwen8's corresponding full-strength stripe generation intervals include zero.

\begin{table*}[t]
\centering\footnotesize
\setlength{\tabcolsep}{7pt}
\caption{\textbf{Qwen32's stripe composition residual reverses sign with layout.} Raw composition residual $C$ with direct answers. Each value is a paired template-level estimate over 32 complete templates, with a descriptive normal 95\% interval.}
\DSalign{appendix-3}
\begin{tabular}{@{}lcllr@{}}
\toprule
\rowcolor{DSTableHeader}
Layout & $\alpha$ & Score $C$ [95\% interval] & Generation $C$ [95\% interval] & $n$ \\
\midrule
Original & \DSn{.08} & \DSv{8.655}{[6.608,10.701]} & \DSv{0.313}{[0.198,0.427]} & \DSn{32} \\
Original & \DSn{1} & \DSv{5.508}{[3.476,7.541]} & \DSv{0.172}{[0.050,0.293]} & \DSn{32} \\
Exchanged & \DSn{.08} & \DSv{-8.351}{[-10.702,-6.000]} & \DSv{-0.250}{[-0.382,-0.118]} & \DSn{32} \\
Exchanged & \DSn{1} & \DSv{-4.282}{[-6.060,-2.504]} & \DSv{-0.266}{[-0.411,-0.120]} & \DSn{32} \\
\bottomrule
\end{tabular}
\label{tab:composition}
\end{table*}

\begin{table}[!htb]
\centering\footnotesize
\caption{\textbf{After direction alignment, no layout difference in Qwen32's residual is resolved.} Stripe composition residual with the direction of spatial correspondence aligned. Each contrast pairs the two layouts within the same 32 templates.}
\DSalign{appendix-4}
\begin{tabular}{@{}lcll@{}}
\toprule
\rowcolor{DSTableHeader}
Readout & $\alpha$ & $-C(M1)-C(M0)$ & 95\% interval \\
\midrule
Generation & \DSn{.08} & \DSv{-0.063}{} & \DSci{[-0.220,0.095]} \\
Generation & \DSn{1} & \DSv{0.094}{} & \DSci{[-0.095,0.283]} \\
Score & \DSn{.08} & \DSv{-0.304}{} & \DSci{[-2.496,1.888]} \\
Score & \DSn{1} & \DSv{-1.226}{} & \DSci{[-3.311,0.858]} \\
\bottomrule
\end{tabular}
\label{tab:composition_aligned}
\end{table}

\subsection{Granite subsets and individual entries}
For the spatial $(15,18,21)$ versus intermediate $(3,6,9)$ subsets at $M0$, $\alpha=1$, and raw dose, the score interaction $H$ is $0.28$ [$0.22,0.34$], $0.12$ [$0.03,0.21$], and $0.19$ [$0.10,0.29$] under direct, identity, and cyclic readouts, while the generation interactions are $-0.02$ [$-0.09,0.05$], $0.13$ [$-0.01,0.26$], and $0.02$ [$-0.07,0.10$], all $n=32$. At $\alpha=.08$, score intervals remain positive and generation intervals contain zero. Identity-label stripe $G$ is $0.16$ [$0.05,0.27$] at full strength, versus color $G=0.03$ [$-0.03,0.09$], and at $.08$ stripe $G=0.16$ [$0.04,0.28$] and cyclic stripe $G=0.06$ [$0.004,0.12$]. Group differences within stripes therefore need not yield resolved attribute interactions.

Native direct-score stripe correctness (accuracy on the unedited sources) is 64/128 and 65/128 for the two sources, versus generation's 72/128 and 68/128, while label formats have higher native correctness and retain the score interaction. We divide each attribute's group difference by its mean all-later score effect. This gives direct full-strength $H=0.32$, with paired bootstrap interval $[0.27,0.37]$ (10,000 template resamples with a fixed seed, denominators recomputed). This sensitivity preserves the direction under one output scaling but does not remove capability, dose, or depth confounds. The separate conditional marginal $M_{21\mid15,18}$ (defined in Appendix~\ref{app:dose}; results in Appendix~\ref{app:v14}) avoids combining endpoints measured in separate experiments.

At $M0$ and $\alpha=.08$, Qwen32's direct-score differences, later entry 0 minus later entry 1, are $-0.31$ [$-0.45,-0.17$] for color and $-0.69$ [$-1.34,-0.03$] for stripes at raw dose, and shared-low matching gives $-0.01$ [$-0.11,0.08$] and $-0.07$ [$-0.23,0.10$], all $n=32$. Qwen8's corresponding matched differences are $-0.23$ [$-0.32,-0.13$] and $-0.42$ [$-0.68,-0.16$], and its generation intervals contain zero ($n=25/32$). Because RSS and L1 singleton interventions are physically identical, their agreement is not independent confirmation, and other layouts can change the ordering.

Granite also varies by condition. At $M0$ and $\alpha=.08$, raw direct-score stripe selectivity is $0.08$ [$0.04,0.11$] for main and $0.60$ [$0.55,0.64$] for entry 21, and under shared-low RSS the color difference, main minus entry 15, is $-0.08$ [$-0.09,-0.06$]. With exchanged layout and full-strength color, main score selectivity is $0.87$ [$0.67,1.06$] but generation selectivity is $-0.25$ [$-0.35,-0.15$]. These conditional effects do not assign fixed semantic roles to layers.

\section{Dose and low-strength boundaries}
\label{app:boundaries}
\begin{figure*}[t]
\centering
\includegraphics[width=\textwidth]{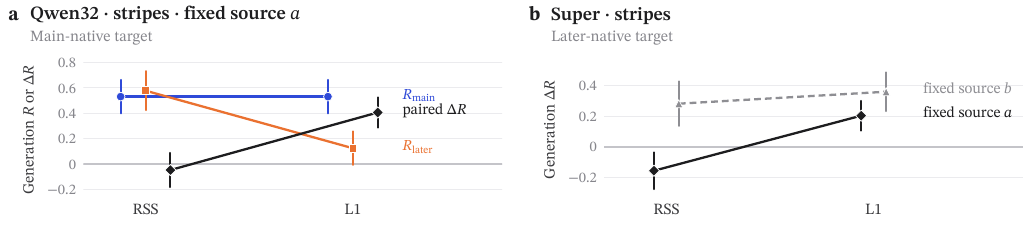}
\caption{\textbf{Dose matching can tie two selective entries or reverse their order in one cell.} These full-source conditions use direct generation, $M0/M0$, both source strengths $\alpha=1$, and 32 complete shared templates per point. (a) With Qwen32 stripes and fixed source $a$, main-native RSS leaves both $R_{\mathrm{main}}$ and $R_{\mathrm{later}}$ positive while their directly paired $\Delta R$ interval contains zero; main-native L1 changes the ordering. (b) With Cosmos3-Super stripes at the later-native target, fixed source $a$ yields a negative $\Delta R$ under RSS and a positive one under L1; fixed source $b$ yields positive differences under both measures. The negative Super cell is one condition in a larger grid and is not treated as an independent replicated reversal. Generation is the net change in the frequency of the edited object's $b$ answer. Bars are descriptive normal 95\% intervals; $\Delta R$ intervals are directly paired.}
\label{fig:boundaries}
\end{figure*}
\textbf{Photograph directions across formats and doses.}
\label{app:census}
For the photographs at $M0$ and full background, each checkpoint has 15 entry contrasts per measure: three answer formats crossed with raw strength and the RSS and L1 targets at main- and later-native doses. Qwen8, Qwen32 and Cosmos3-Super have positive $\Delta R$ intervals in all 15 cells for both scores and generation, as do the nine further raw cells with layout exchange or weak background. DeepStack's intervals are negative in all 15 cells and all nine further cells for both measures, with generation contrasts from $-0.58$ to $-0.28$ (the weakest is cyclic labels at the main-native L1 target, $-0.28$ [$-0.36,-0.20$]). By contrast, Granite has 3 positive, 9 negative and 3 zero-containing score cells and 1, 9 and 5 generation cells, and every positive cell uses L1 matching; its identity-label raw generation contrast is $-0.22$ [$-0.30,-0.14$].

A small contrast can arise when both entries are selective. In Qwen32 stripes, fixed $a$, and direct generation, main-native RSS gives $R_{\main}=0.53$ [$0.40,0.66$] and $R_{\later}=0.58$ [$0.43,0.73$], but paired $\Delta R=-0.05$ [$-0.18,0.09$] ($n=32$; Figure~\ref{fig:boundaries}a). Conversely, for Nano color with fixed $a$ and the later-native RSS target, generation selectivity is zero in both groups but score $\Delta R=0.68$ [$0.42,0.94$].

Super supplies an isolated negative cell among its matched stripe conditions (Figure~\ref{fig:boundaries}b). With fixed $a$ and the later-native target, RSS score/generation differences are $-1.60$ [$-2.36,-0.85$] and $-0.16$ [$-0.28,-0.04$] and L1 gives $2.37$ [$1.77,2.97$] and $0.20$ [$0.11,0.30$], while fixing $b$ under RSS also gives positive generation $0.28$ [$0.14,0.43$], all $n=32$. Across 24 matched conditions, score intervals are positive, zero-containing, and negative in 18/5/1 cells, and generation intervals in 17/6/1; the negative generation cell has four positive, 14 tied, and 14 negative template contrasts. The interval does not account for selecting this cell from an exploratory grid.

Qwen32 stripes favor main under L1 at both targets. Granite fixed-$a$ color generation moves from $-0.41$ [$-0.52,-0.30$] at raw dose to $0.30$ [$0.20,0.39$] under main-native L1, while fixed-$b$ L1 remains later-favoring. Qwen30 identity-label color at the later-native target has score $\Delta R=-0.17$ [$-0.75,0.41$] under RSS and $6.02$ [$4.81,7.24$] under L1, and DeepStack color remains later-favoring across all 24 matched settings in both measures, whereas its matched direct stripe intervals contain zero. Scalar matching therefore constrains magnitude explanations without establishing a context-free ordering.

\begin{table*}[t]
\centering
\caption{\textbf{At the shared-low dose, neither Qwen model keeps generation selectivity.} Entry decomposition for color, direct answers, $M0$, $\alpha=1$. Each cell gives a mean above its descriptive normal 95\% interval, and every estimate uses 32 complete templates. All-later contains the three later entries and excludes main, and the rightmost column within each endpoint is the directly paired $R_{\mathrm{main}}-R_{\mathrm{all}}$. Rows labeled Low use shared-low matching: the minimum native singleton dose (the smallest dose of any one entry at raw replacement) at the source-$a$ baseline shared by all groups. Under RSS (shaded rows), the small matched score contrast is positive for Qwen8 and negative for Qwen32.}
\label{tab:p4}
\begingroup
\fontsize{9}{10.5}\selectfont
\setlength{\tabcolsep}{3pt}
\renewcommand{\arraystretch}{1.0}
\DSalign{table_common_low_supp-1}
\begin{tabularx}{\linewidth}{@{}l*{6}{>{\centering\arraybackslash}X}@{}}
\toprule
& \multicolumn{3}{c}{Score ($n=32$ per estimate)} & \multicolumn{3}{c}{Generation ($n=32$ per estimate)} \\
\cmidrule(lr){2-4}\cmidrule(l){5-7}
\rowcolor{DSTableHeader}
Dose & $R_{\mathrm{main}}$ & $R_{\mathrm{all}}$ & Main $-$ all & $R_{\mathrm{main}}$ & $R_{\mathrm{all}}$ & Main $-$ all \\
\midrule
\multicolumn{7}{@{}l}{\textbf{Qwen8}} \\[2pt]
Raw ($k=1$) & \DSstack{27.286}{[23.282,31.290]} & \DSstack{0.269}{[0.148,0.389]} & \DSstack{27.017}{[23.022,31.013]} & \DSstack{0.563}{[0.418,0.707]} & \DSstack{0.000}{[0.000,0.000]} & \DSstack{0.563}{[0.418,0.707]} \\[3pt]
\rowcolor{DSTableAccent}
Low RSS & \DSstack{0.217}{[0.124,0.310]} & \DSstack{0.102}{[0.044,0.159]} & \DSstack{0.115}{[0.033,0.197]} & \DSstack{0.000}{[0.000,0.000]} & \DSstack{0.000}{[0.000,0.000]} & \DSstack{0.000}{[0.000,0.000]} \\[3pt]
Low L1 & \DSstack{0.217}{[0.124,0.310]} & \DSstack{0.070}{[-0.005,0.146]} & \DSstack{0.147}{[0.050,0.244]} & \DSstack{0.000}{[0.000,0.000]} & \DSstack{0.000}{[0.000,0.000]} & \DSstack{0.000}{[0.000,0.000]} \\[3pt]
\addlinespace[2pt]\midrule\addlinespace[2pt]
\multicolumn{7}{@{}l}{\textbf{Qwen32}} \\[2pt]
Raw ($k=1$) & \DSstack{38.314}{[35.941,40.688]} & \DSstack{0.667}{[0.532,0.802]} & \DSstack{37.647}{[35.250,40.044]} & \DSstack{0.813}{[0.707,0.918]} & \DSstack{0.000}{[0.000,0.000]} & \DSstack{0.813}{[0.707,0.918]} \\[3pt]
\rowcolor{DSTableAccent}
Low RSS & \DSstack{0.067}{[0.020,0.113]} & \DSstack{0.115}{[0.054,0.177]} & \DSstack{-0.048}{[-0.093,-0.004]} & \DSstack{0.000}{[0.000,0.000]} & \DSstack{0.000}{[0.000,0.000]} & \DSstack{0.000}{[0.000,0.000]} \\[3pt]
Low L1 & \DSstack{0.067}{[0.020,0.113]} & \DSstack{0.097}{[0.043,0.151]} & \DSstack{-0.030}{[-0.071,0.010]} & \DSstack{0.000}{[0.000,0.000]} & \DSstack{0.000}{[0.000,0.000]} & \DSstack{0.000}{[0.000,0.000]} \\[3pt]
\bottomrule
\end{tabularx}
\endgroup
\end{table*}
Table~\ref{tab:p4} reports the raw and shared-low color comparisons. At shared-low strength both Qwen checkpoints have zero net generation selectivity in both groups, yet their score differences remain distinct: Qwen8 stays positive, but Qwen32 gives $-0.05$ [$-0.09,-0.004$] under RSS and an L1 interval containing zero. This small score-only reversal is an exploratory boundary. These comparisons use a different baseline and target from the photograph dose controls.

\section{Quality-conditioned influence and recovery}
\label{app:v14}
\subsection{Design, materials, and interpretation}
The conflict comparison changes main's quality while later's clean-$a$ baseline, region mask, source difference, and multiplier $k$ stay fixed (Table~\ref{tab:intervention_spec}; Appendix~\ref{app:dose}). Target and control queries receive the same intervention, and $I_{\mathrm{quality}}=R_{\later}(D)-R_{\later}(C)$ is the change in later's relative selectivity $R$ (Appendix~\ref{app:statistics}) when main is degraded rather than clean. Recovery instead supplies clean and degraded versions of one image without conflicting attribute sources. In $CC/CD/DC/DD$, the first letter denotes main quality. Native accuracy is $A_{CC}$, later recovery is $G_D=A_{DC}-A_{DD}$, and the interaction is $I_{\mathrm{rec}}=G_D-(A_{CC}-A_{CD})$. A positive gain need not yield a positive interaction, because $I_{\mathrm{rec}}$ subtracts the gain that clean later features already give when main is clean.

Synthetic materials use a fixed seed and 32 templates shared across four tasks: color, stripe orientation, relative position within one shape, and short text. Two single-object edits with two queries (target and control) per edit give 128 queries per task and 512 overall. Because position marks lie within one shape, that task differs from exchanging two objects. The 64 reused photographs give 256 queries under new quality conditions, not an independent replication of the materials. Both sets use contrast 0.12 and retain image dimensions; the four cells reuse identical degraded pixels, and processed grids and visual hashes are recorded.

In all 512 synthetic queries, the clean-$a$ correct option follows object side (left: first; right: second), and reversing the numeric labels preserves this predictability. Edited-$b$ target answers flip while controls stay unchanged, so the same rule does not hold across both sources. Native success therefore need not demonstrate attribute recognition independent of position. Within-query pairing identifies effects on these stimuli but does not exclude interactions between degraded evidence and a position--answer shortcut, and recovery cannot isolate which visual information returns. An always-one-option baseline scores 0.5 with zero standard error, yet matching that endpoint does not identify a model's strategy. The calibrated experiment (Appendix~\ref{app:v15}) balances all four side/option combinations on separate templates.

\begin{table}[!htb]
\centering\footnotesize
\setlength{\tabcolsep}{5pt}
\renewcommand{\arraystretch}{1.02}
\caption{\textbf{Qwen8/32 color accuracy stays at ceiling except under cyclic labels.} Synthetic-color generation accuracy (\%) at contrast $0.12$. Cell labels give main's source first and later's second, each clean (C) or degraded (D). All 32 templates contribute in every format, and invalid answers count as incorrect. Paired contrasts and intervals for every task are in \texttt{fixed\_\allowbreak{}contrast\_\allowbreak{}recovery.csv}.}
\label{tab:v14_color_cells}
\DSalign{table_color_cells-1}
\begin{tabular}{@{}llrrrr@{}}
\toprule
\rowcolor{DSTableHeader}
Model & Format & CC & CD & DC & DD \\
\midrule
Qwen8 & Direct & \DSn{100.0} & \DSn{100.0} & \DSn{100.0} & \DSn{100.0} \\[0pt]
 & Identity & \DSn{100.0} & \DSn{100.0} & \DSn{100.0} & \DSn{100.0} \\[0pt]
 & Cyclic & \DSn{100.0} & \DSn{100.0} & \DSn{59.4} & \DSn{48.4} \\[0pt]
\addlinespace[2pt]
Qwen32 & Direct & \DSn{100.0} & \DSn{100.0} & \DSn{100.0} & \DSn{100.0} \\[0pt]
 & Identity & \DSn{100.0} & \DSn{100.0} & \DSn{100.0} & \DSn{100.0} \\[0pt]
 & Cyclic & \DSn{100.0} & \DSn{100.0} & \DSn{100.0} & \DSn{70.3} \\[0pt]
\addlinespace[2pt]
Granite & Direct & \DSn{100.0} & \DSn{98.4} & \DSn{100.0} & \DSn{93.8} \\[0pt]
 & Identity & \DSn{100.0} & \DSn{100.0} & \DSn{100.0} & \DSn{100.0} \\[0pt]
 & Cyclic & \DSn{92.2} & \DSn{71.9} & \DSn{92.2} & \DSn{50.0} \\[0pt]
\addlinespace[2pt]
DeepStack & Direct & \DSn{100.0} & \DSn{100.0} & \DSn{100.0} & \DSn{84.4} \\[0pt]
 & Identity & \DSn{100.0} & \DSn{100.0} & \DSn{100.0} & \DSn{92.2} \\[0pt]
 & Cyclic & \DSn{82.8} & \DSn{96.9} & \DSn{95.3} & \DSn{84.4} \\[0pt]
\bottomrule
\end{tabular}
\end{table}
Table~\ref{tab:v14_color_cells} gives all four cells for color. Qwen8/32 stay saturated with direct answers and identity labels, so only cyclic labels expose recovery. With cyclic labels, Granite's later features also help when main is clean ($A_{CC}>A_{CD}$), whereas DeepStack's positive cyclic interaction partly reflects a negative clean-main benefit ($A_{CC}<A_{CD}$). The cells thus separate these cases from recovery that comes solely from better degraded-main performance.

Granite's color gain with cyclic labels is $G_D=0.42$ [$0.36,0.49$], with $I_{\mathrm{rec}}=0.22$ [$0.11,0.33$]; DeepStack's direct-color gain is $0.16$ [$0.07,0.24$]. The cross-format comparison keeps all planned synthetic-color queries from the same 32 templates, counts invalid outputs as incorrect, and does not restrict to complete cases within each readout. For Granite, the cyclic-minus-direct difference, estimated within templates, is $0.36$ [$0.28,0.44$] for $G_D$ but $0.17$ [$0.06,0.28$] for $I_{\mathrm{rec}}$ (paired estimates in \texttt{fixed\_\allowbreak{}contrast\_\allowbreak{}format\_\allowbreak{}pairs.csv}). The $G_D$ difference measures the change in degraded-main benefit, whereas the $I_{\mathrm{rec}}$ difference also subtracts the change in clean-main benefit. Readout dependence therefore concerns the four-cell response pattern, not only whether one recovery interval excludes zero.

\subsection{Execution, readouts, and pairing}
All four models complete the synthetic and photograph quality and recovery runs; Granite also completes the conditional marginal. Experiments and benchmarks use separately validated batched implementations, with conditions batched when text and geometry match. Inference uses A100 80GB GPUs and BF16 (PyTorch 2.11.0/2.10.0/2.1.2 and Transformers 5.8.0/5.8.0/4.39.3 for Qwen/Granite/DeepStack, respectively). Two jobs shared one GPU only after a check: on each of the two machines used, a completed unit re-executed alongside a running job reproduced all 768 of its observations bitwise. The final runs comprise 176,512 observations. Scalar outputs are reused only when the complete visual tensors are identical; residual states, attention states, and key/value caches are not shared across main conditions.

Batched and unbatched checks on selected units agree on all 384 DeepStack synthetic decisions and token strings. Granite and Qwen32 agree on all 384 decisions, with exact tokens on 379/380; Granite's five differences append words after the same labels. Qwen8 on photographs agrees on 95/96 decisions and token strings, the exception being a near-tie margin that changes from $-0.02$ to $0.06$. A separate synthetic check agrees on decisions and 370/384 token strings. Maximum candidate log-probability differences are 0, 0.31, 0.53, and 0.71 for DeepStack, Granite, Qwen32, and Qwen8 on photographs. These document BF16 batch-shape sensitivity, not equality across all runs. No paired DocVQA batching check was performed; unbatched runs and answer-format checks remain separate.

Direct, identity-label, and cyclic-label answers are all retained (Appendix~\ref{app:statistics}). Native and recovery accuracy counts invalid answers as incorrect over all planned queries, with valid-only sensitivity estimates. Conflict and marginal effects instead require valid paired endpoints and report actual counts: Granite's direct-text marginal has 25 complete templates, versus 32 for its other displayed generation marginals. In the marginal, entries 15 and 18 keep identical increments, and only entry 21 changes from the source-$a$ baseline to its $k=1$ increment. Native and recovery cells were not planned for the Granite conditional-marginal runs, so the native accuracy in Table~\ref{tab:v14_marginal_full} comes from the separate quality runs, not from a matched marginal endpoint.

Readout differences subtract formats within shared templates before intervals are estimated. Synthetic-color generation comparisons keep all planned queries and are the principal readout evidence; elsewhere, shared templates can contain different valid query sets across formats, so those comparisons are flagged and secondary. All cell outputs, including invalid and truncated responses, remain in the observations.

\subsection{Benchmark subsets and scoring}
GQA 1.2 validation-balanced sampling uses a fixed seed and at most two questions per image, after removing 2,663 candidate questions on images used in our other experiments. It gives 465 attribute questions on 256 images and 472 spatial questions on 256 images, with eight images shared between the groups. DocVQA 1.0 official validation sampling uses the same seed and gives 495 questions from 256 original documents and 296 pages, clustered by document. These are validation subsets, not full test-set evaluations.

Pillow Gaussian blur (radius 2.0) is applied before native processing without resizing the source. Qwen uses 4,096--262,144 pixels; Granite uses native $384\times384$ tiling; DeepStack uses CLIP $336\times336$ processing and spatial crops. Generation is greedy with 64 new tokens and sequence limit 4,096. All four cells share questions and references, and all 12 panels retain 256 planned clusters without missing observations (Table~\ref{tab:v14_benchmark_full}).

The GQA instruction (a single lowercase word or short phrase) was selected once for all models, after checking answer formats on clean inputs and before any recovery comparison. Strict equality is the reported metric. Named sensitivities lowercase and strip whitespace (first rule), then additionally remove terminal periods, exclamation and question marks, and surrounding quotes (second rule). Granite's attribute recovery is $0.09$ [$0.05,0.12$] under the first rule but zero [$-0.03,0.03$] under the second; its spatial recovery changes from $0.10$ [$0.07,0.14$] to $0.04$ [$0.01,0.07$]. The interactions under the second rule contain zero. Zero clean accuracy under strict matching therefore cannot establish absent visual competence in Granite and DeepStack; DeepStack's lowercase-and-strip native accuracies are 0.78/0.71 for attributes/spatial questions.

DocVQA uses organizer-framework ANLS (average normalized Levenshtein similarity): the best normalized similarity across references after lowercasing and stripping. Values below 0.5 are set to zero, but exactly 0.5 is retained.\footnote{Reference implementations: \href{https://github.com/rubenpt91/MP-DocVQA-Framework/blob/656d4e5a316720c83a1b1679fb39f424265655d7/metrics.py}{DocVQA organizer framework, revision \texttt{656d4e5}}; \href{https://github.com/dorarad/gqa/blob/9f88c1118598ba459266b9d3a6b74cd3db60bc44/evaluation/eval.py}{official GQA evaluator, revision \texttt{9f88c11}}.} The strict-paper-cutoff sensitivity excludes that boundary; lowercase-and-strip exact matching is separate. ANLS is not binary accuracy. For DeepStack/Granite/Qwen8/Qwen32, joint losses $A_{CC}-A_{DD}$ are $0.03/0.01/0.02/-0.01$, and later recoveries are $0.02/0.01/0.00/-0.01$, all with intervals containing zero (Table~\ref{tab:v14_benchmark_full}). Because these induced losses are small, they limit conclusions about recovery from substantial damage; Appendix~\ref{app:v15} calibrates larger losses on separate units.

\subsection{Complete estimates}
The files \texttt{fixed\_\allowbreak{}contrast\_\allowbreak{}recovery.csv}, \texttt{fixed\_\allowbreak{}contrast\_\allowbreak{}conflict.csv}, \texttt{fixed\_\allowbreak{}contrast\_\allowbreak{}granite\_\allowbreak{}marginal.csv}, \texttt{fixed\_\allowbreak{}contrast\_\allowbreak{}format\_\allowbreak{}pairs.csv} and \texttt{fixed\_\allowbreak{}contrast\_\allowbreak{}benchmarks.csv} (Appendix~\ref{app:data}) retain all task/readout conditions, target/control components, four cells, recovery interactions, conditional marginals, and named benchmark sensitivities, with original clustering and estimate-specific denominators. The tables below give the estimates that the main text cites. Accuracy policies distinguish invalid-as-zero, scoring returned strings, and complete valid four-cell sets where available. Intervals follow Appendix~\ref{app:statistics}; none of these results is presented as confirmatory.

\Needspace{6\baselineskip}\subsection{Conflict quality interactions}
Degrading main raises later score selectivity on synthetic color in all four models, and Table~\ref{tab:v14_conflict_full} gives the quality interactions cited in the main text. Other tasks and readouts are in the CSV files listed above.
\begin{table*}[t]
\centering\footnotesize
\setlength{\tabcolsep}{4pt}
\renewcommand{\arraystretch}{1.04}
\caption{\textbf{Degrading main raises later score selectivity on synthetic color in all four models.} Quality interaction $I_{\rm quality}=R_{\later}(D)-R_{\later}(C)$ in the fixed-contrast experiment: the change in later target-minus-control selectivity when main is degraded to contrast 0.12 while the later features stay fixed. Score values are candidate log odds, and generation values are directed answer-frequency changes. Synthetic color uses 32 templates and photographs 64, counting only complete units (every required answer valid). Intervals are descriptive, unadjusted 95\% $t$ intervals.}
\label{tab:v14_conflict_full}
\DSalign{appendix_results-1}
\begin{tabular}{@{}lccccc@{}}
\toprule
 & \multicolumn{4}{c}{Synthetic color} & Photographs \\
\cmidrule(lr){2-5}\cmidrule(l){6-6}
\rowcolor{DSTableHeader}
Model & Score, direct & Score, identity & Score, cyclic & Gen., direct & Score, direct \\
\midrule
Qwen8 & \DSv{1.715}{[1.479, 1.951]} & \DSv{1.047}{[0.801, 1.293]} & \DSv{1.879}{[1.385, 2.373]} & \DSv{0.000}{[0.000, 0.000]} & \DSv{0.182}{[0.026, 0.339]} \\
Qwen32 & \DSv{2.834}{[2.381, 3.288]} & \DSv{2.521}{[1.982, 3.060]} & \DSv{4.024}{[2.869, 5.178]} & \DSv{0.000}{[0.000, 0.000]} & \DSv{0.489}{[0.232, 0.746]} \\
Granite & \DSv{1.103}{[0.897, 1.309]} & \DSv{1.153}{[0.942, 1.364]} & \DSv{0.465}{[0.311, 0.619]} & \DSv{0.234}{[0.132, 0.337]} & \DSv{-0.080}{[-0.148, -0.013]} \\
DeepStack & \DSv{0.680}{[0.558, 0.803]} & \DSv{1.369}{[1.228, 1.510]} & \DSv{1.014}{[0.786, 1.243]} & \DSv{0.188}{[0.060, 0.315]} & \DSv{0.055}{[-0.015, 0.125]} \\
\bottomrule
\end{tabular}
\end{table*}
\Needspace{6\baselineskip}\subsection{Granite conditional marginal}
With Granite's entries 15 and 18 held fixed, adding later entry 21 raises generation selectivity in three of the twelve task and readout cells (Table~\ref{tab:v14_marginal_full}); Appendix~\ref{app:dose} defines this conditional marginal.
\begin{table}[!htb]
\centering\footnotesize
\setlength{\tabcolsep}{3.5pt}
\renewcommand{\arraystretch}{1.02}
\caption{\textbf{Entry 21 adds selectivity in three of the twelve task and readout cells.} Granite entry-21 marginal given fixed entries 15 and 18: paired target-minus-control generation effect, not a share of a joint effect. Native accuracy comes from the separate fixed-contrast condition. Units and $t$ intervals follow that panel; score marginals and components are in \texttt{fixed\_\allowbreak{}contrast\_\allowbreak{}granite\_\allowbreak{}marginal.csv}.}
\label{tab:v14_marginal_full}
\DSalign{appendix_results-2}
\begin{tabular}{@{}llrrc@{}}
\toprule
\rowcolor{DSTableHeader}
Task & Readout & $n$ & Native & Marginal \\
\midrule
Color & Direct & \DSn{32} & \DSn{1.000} & \DSv{0.000}{[0.000, 0.000]} \\
 & Identity & \DSn{32} & \DSn{1.000} & \DSv{0.000}{[0.000, 0.000]} \\
 & Cyclic & \DSn{32} & \DSn{0.922} & \DSv{0.156}{[0.071, 0.241]} \\
\addlinespace[2pt]
Relative position & Direct & \DSn{32} & \DSn{0.922} & \DSv{0.094}{[0.022, 0.165]} \\
 & Identity & \DSn{32} & \DSn{0.828} & \DSv{0.000}{[-0.046, 0.046]} \\
 & Cyclic & \DSn{32} & \DSn{0.984} & \DSv{0.000}{[0.000, 0.000]} \\
\addlinespace[2pt]
Stripes & Direct & \DSn{32} & \DSn{0.516} & \DSv{0.000}{[0.000, 0.000]} \\
 & Identity & \DSn{32} & \DSn{1.000} & \DSv{0.266}{[0.154, 0.378]} \\
 & Cyclic & \DSn{32} & \DSn{0.656} & \DSv{0.031}{[-0.047, 0.110]} \\
\addlinespace[2pt]
Text & Direct & \DSn{25} & \DSn{0.891} & \DSv{0.000}{[0.000, 0.000]} \\
 & Identity & \DSn{32} & \DSn{1.000} & \DSv{0.000}{[0.000, 0.000]} \\
 & Cyclic & \DSn{32} & \DSn{0.609} & \DSv{0.016}{[-0.016, 0.047]} \\
\bottomrule
\end{tabular}
\end{table}
\Needspace{6\baselineskip}\subsection{Standard-task cells}
Blur at radius 2 costs little GQA or DocVQA accuracy and resolves no positive later recovery; Table~\ref{tab:v14_benchmark_full} gives the four cells for GQA and DocVQA, with later recovery $G_D$ and its interaction $I_{\rm rec}$.
\begin{table*}[t]
\centering\footnotesize
\setlength{\tabcolsep}{4pt}
\renewcommand{\arraystretch}{1.02}
\caption{\textbf{Blur at radius 2 costs little GQA or DocVQA accuracy, and no positive later recovery is resolved.} Fixed-contrast standard-task cells under Gaussian blur (radius 2.0), over 256 image or document clusters per task. The reported metric is strict matching for GQA and organizer ANLS for DocVQA, with invalid outputs scored zero. $G_D$ is later recovery with degraded main, and $I_{\rm rec}$ is its difference from the clean-main gain; intervals are descriptive normal 95\% intervals. Named metric sensitivities and invalid-answer policies are in \texttt{fixed\_\allowbreak{}contrast\_\allowbreak{}benchmarks.csv}. Granite and DeepStack capitalize their GQA answers, so their strict GQA rows sit at a floor set by answer format.}
\label{tab:v14_benchmark_full}
\DSalign{appendix_results-3}
\begin{tabular}{@{}llrrrrcc@{}}
\toprule
\rowcolor{DSTableHeader}
Task & Model & CC & CD & DC & DD & $G_D$ & $I_{\rm rec}$ \\
\midrule
DocVQA & Qwen8 & \DSn{0.800} & \DSn{0.794} & \DSn{0.780} & \DSn{0.785} & \DSv{-0.004}{[-0.018, 0.009]} & \DSv{-0.010}{[-0.031, 0.010]} \\
 & Qwen32 & \DSn{0.794} & \DSn{0.796} & \DSn{0.786} & \DSn{0.800} & \DSv{-0.014}{[-0.030, 0.002]} & \DSv{-0.012}{[-0.032, 0.007]} \\
 & Granite & \DSn{0.847} & \DSn{0.840} & \DSn{0.850} & \DSn{0.839} & \DSv{0.011}{[-0.003, 0.024]} & \DSv{0.004}{[-0.011, 0.019]} \\
 & DeepStack & \DSn{0.374} & \DSn{0.357} & \DSn{0.363} & \DSn{0.343} & \DSv{0.021}{[-0.001, 0.042]} & \DSv{0.003}{[-0.014, 0.021]} \\
\addlinespace[2pt]
GQA attr. & Qwen8 & \DSn{0.705} & \DSn{0.709} & \DSn{0.670} & \DSn{0.652} & \DSv{0.018}{[-0.004, 0.040]} & \DSv{0.021}{[-0.010, 0.053]} \\
 & Qwen32 & \DSn{0.719} & \DSn{0.713} & \DSn{0.701} & \DSn{0.689} & \DSv{0.012}{[-0.014, 0.038]} & \DSv{0.006}{[-0.026, 0.038]} \\
 & Granite & \DSn{0.000} & \DSn{0.004} & \DSn{0.000} & \DSn{0.014} & \DSv{-0.014}{[-0.024, -0.004]} & \DSv{-0.010}{[-0.018, -0.001]} \\
 & DeepStack & \DSn{0.000} & \DSn{0.000} & \DSn{0.000} & \DSn{0.000} & \DSv{0.000}{[0.000, 0.000]} & \DSv{0.000}{[0.000, 0.000]} \\
\addlinespace[2pt]
GQA spatial & Qwen8 & \DSn{0.604} & \DSn{0.592} & \DSn{0.572} & \DSn{0.553} & \DSv{0.020}{[-0.001, 0.040]} & \DSv{0.008}{[-0.020, 0.035]} \\
 & Qwen32 & \DSn{0.576} & \DSn{0.572} & \DSn{0.539} & \DSn{0.564} & \DSv{-0.025}{[-0.053, 0.002]} & \DSv{-0.029}{[-0.062, 0.003]} \\
 & Granite & \DSn{0.000} & \DSn{0.002} & \DSn{0.000} & \DSn{0.002} & \DSv{-0.002}{[-0.006, 0.002]} & \DSv{0.000}{[0.000, 0.000]} \\
 & DeepStack & \DSn{0.000} & \DSn{0.000} & \DSn{0.000} & \DSn{0.000} & \DSv{0.000}{[0.000, 0.000]} & \DSv{0.000}{[0.000, 0.000]} \\
\bottomrule
\end{tabular}
\end{table*}

\section{Recovery along calibrated degradation curves}
\label{app:v15}
\subsection{Design and units}
The calibrated experiment asks how later recovery changes with each model's measured loss. It keeps the four models, the four synthetic tasks, and the recovery and conflict definitions of Appendix~\ref{app:v14}. It replaces the single contrast level with three calibrated degradation levels per task, chosen on separate calibration units (described below) and shared by all models (the shared levels). GQA and DocVQA receive the same treatment, and evaluation uses units that played no part in calibration.

Calibration units serve only to choose the levels. Synthetic calibration uses 16 templates (256 queries). GQA attribute and spatial calibration each reuse 48 images from the fixed-contrast experiment (87 and 85 questions), and DocVQA calibration reuses 48 of its documents (56 pages, 94 questions); this is named reuse, not replication. The 32 synthetic templates per task (128 queries per task) balance object side against the answer option, so each of the four combinations occurs in 32 rows per task. This removes the side--answer association of the clean-$a$ image described in Appendix~\ref{app:v14}. Each template has two edits with a target and a control query each. Both edits share one clean image, so recovery has 64 distinct template--object items per task, each counted twice, which leaves template means unchanged.

GQA 1.2 validation-balanced sampling (at most two questions per image) gives 192 images per question group, with 349 attribute and 340 spatial questions. Single-Document DocVQA 1.0 validation sampling gives 192 documents, 216 pages, and 373 questions, at most two per document. All sampling uses fixed seeds, and benchmark selection excludes every benchmark unit of the fixed-contrast experiment and the photograph panel, so calibration and evaluation units are disjoint. These are selected validation subsets, not leaderboard evaluations.

\subsection{Degradation families and level selection}
Each task uses one degradation family scaled to its stimulus. Color keeps a fraction of chroma with luminance unchanged, and stripes and relative position use Gaussian blur with $\sigma$ relative to the stripe period or to the dot separation. Text, GQA, and DocVQA reduce resolution, expressed as equivalent font size for text and as long side for images and pages. Each level is one image file derived from the clean image with Pillow 12.2.0, keeps the clean pixel size, and is given to all four models. Calibration runs only the clean cell (CC) and the jointly degraded cell (DD) of each candidate level, under every readout. Recovery and conflict cells were never run on calibration units, so no recovery or conflict effect could inform the choice.

For each model, retained ability at a level is $r=(A_{DD}-\text{chance})/(A_{CC}-\text{chance})$, averaged over native-capable readouts. A readout is native-capable when $A_{CC}$ exceeds chance by at least $0.2(1-\text{chance})$, and a model without such a readout is excluded from that task's median. The shared levels are the three ordered levels whose cross-model median $r$ values jointly minimize the summed distance from $0.75$, $0.5$, and $0.25$, with ties broken toward milder levels. GQA calibration uses case-and-punctuation normalization (the second GQA sensitivity of Appendix~\ref{app:v14}), and DocVQA calibration uses organizer ANLS. The GQA metric was fixed before any run because answer formatting had made strict GQA accuracy zero for Granite and DeepStack. Table~\ref{tab:v15_grid} lists the selected levels.

The initial candidate levels left text and most stripe levels at ceiling (little or no loss) and barely affected GQA. We therefore derived heavier or intermediate levels from the same clean images: six text levels (font sizes 4 to 1), four intermediate stripe levels, and six GQA levels (long sides 28 to 5 pixels). This decision was based on calibration accuracy alone. The resulting levels have known limits (Table~\ref{tab:v15_grid}). The most severe stripe level is at the floor (median $r=0.02$), and the mildest relative-position level falls below its target (median $r=0.73$). GQA keeps a median $r$ of at least $0.56$ even at a 5-pixel long side, so its levels span moderate loss only, and whether language priors or coarse visual cues sustain this accuracy was not tested.

Besides the shared levels, a model-specific design evaluates, for each model and task, the calibration level whose $r$ is nearest $0.5$ (\texttt{calibrated\_\allowbreak{}capability\_\allowbreak{}levels.csv}). Most of these levels coincide with shared levels, and four were run outside them. These rows estimate a different quantity and are not pooled with the shared levels.

\begin{table}[!htb]
\centering\footnotesize
\setlength{\tabcolsep}{3pt}
\renewcommand{\arraystretch}{1.02}
\caption{Shared degradation levels chosen on calibration units from clean and jointly degraded cells only. Retained ability is $r=(A_{DD}-\text{chance})/(A_{CC}-\text{chance})$, with chance $0.5$ for two-choice tasks and $0$ for free-form tasks; each model averages its native-capable readouts. The three levels jointly minimize the summed distance of the cross-model median $r$ from $0.75$, $0.5$ and $0.25$. Levels give retained chroma (color), blur $\sigma$ relative to the stripe period or dot separation, equivalent font size, or long side. NN marks non-native models excluded from the median. Q8, Q32, Gr.\ and DS abbreviate the four models.}
\label{tab:v15_grid}
\DSalign{appendix_grid-1}
\begin{tabular}{@{}llrrrrr@{}}
\toprule
\rowcolor{DSTableHeader}
Task & Level & Median $r$ & Q8 & Q32 & Gr. & DS \\
\midrule
Color & \DSn{0.2} & \DSv{0.865}{} & \DSv{0.812}{} & \DSv{0.917}{} & \DSv{0.726}{} & \DSv{0.998}{} \\
 & \DSn{0.1} & \DSv{0.583}{} & \DSv{0.562}{} & \DSv{0.750}{} & \DSv{0.482}{} & \DSv{0.603}{} \\
 & \DSn{0.05} & \DSv{0.286}{} & \DSv{0.281}{} & \DSv{0.375}{} & \DSv{0.073}{} & \DSv{0.290}{} \\
\addlinespace[2pt]
Stripes & \DSn{0.37} & \DSv{0.867}{} & \DSv{0.867}{} & \DSv{1.000}{} & \DSv{0.214}{} & NN \\
 & \DSn{0.4} & \DSv{0.475}{} & \DSv{0.475}{} & \DSv{0.771}{} & \DSv{0.143}{} & NN \\
 & \DSn{0.43} & \DSv{0.021}{} & \DSv{0.021}{} & \DSv{0.000}{} & \DSv{0.143}{} & NN \\
\addlinespace[2pt]
Rel.\ position & \DSn{0.1} & \DSv{0.727}{} & \DSv{0.727}{} & \DSv{0.952}{} & \DSv{0.254}{} & NN \\
 & \DSn{0.2} & \DSv{0.455}{} & \DSv{0.455}{} & \DSv{0.753}{} & \DSv{0.031}{} & NN \\
 & \DSn{0.3} & \DSv{0.182}{} & \DSv{0.182}{} & \DSv{0.488}{} & \DSv{0.062}{} & NN \\
\addlinespace[2pt]
Text & \DSn{4} & \DSv{0.803}{} & \DSv{0.803}{} & \DSv{0.938}{} & \DSv{0.290}{} & NN \\
 & \DSn{3} & \DSv{0.427}{} & \DSv{0.427}{} & \DSv{0.562}{} & \DSv{0.061}{} & NN \\
 & \DSn{2.5} & \DSv{0.229}{} & \DSv{0.319}{} & \DSv{0.229}{} & \DSv{0.061}{} & NN \\
\addlinespace[2pt]
GQA attr. & 28 px & \DSv{0.743}{} & \DSv{0.742}{} & \DSv{0.743}{} & \DSv{0.658}{} & \DSv{0.870}{} \\
 & 10 px & \DSv{0.622}{} & \DSv{0.545}{} & \DSv{0.486}{} & \DSv{0.699}{} & \DSv{0.714}{} \\
 & 5 px & \DSv{0.562}{} & \DSv{0.500}{} & \DSv{0.443}{} & \DSv{0.630}{} & \DSv{0.623}{} \\
\addlinespace[2pt]
GQA spatial & 20 px & \DSv{0.716}{} & \DSv{0.743}{} & \DSv{0.894}{} & \DSv{0.689}{} & \DSv{0.648}{} \\
 & 14 px & \DSv{0.681}{} & \DSv{0.686}{} & \DSv{0.742}{} & \DSv{0.622}{} & \DSv{0.676}{} \\
 & 10 px & \DSv{0.652}{} & \DSv{0.629}{} & \DSv{0.758}{} & \DSv{0.622}{} & \DSv{0.676}{} \\
\addlinespace[2pt]
DocVQA & 360 px & \DSv{0.801}{} & \DSv{0.895}{} & \DSv{0.901}{} & \DSv{0.706}{} & \DSv{0.628}{} \\
 & 256 px & \DSv{0.477}{} & \DSv{0.613}{} & \DSv{0.582}{} & \DSv{0.357}{} & \DSv{0.372}{} \\
 & 180 px & \DSv{0.257}{} & \DSv{0.262}{} & \DSv{0.307}{} & \DSv{0.155}{} & \DSv{0.251}{} \\
\bottomrule
\end{tabular}
\end{table}

\begin{figure*}[t]
\centering
\includegraphics[width=\textwidth]{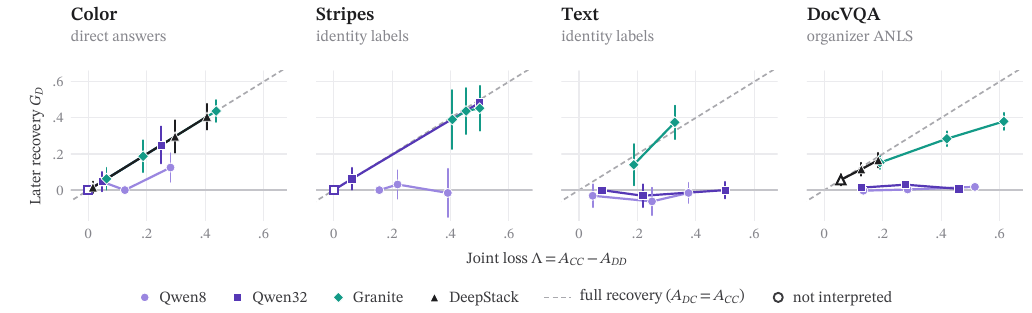}
\caption{\textbf{Equal losses do not give equal recovery.} On text and DocVQA, Granite recovers much of its loss while Qwen8 and Qwen32 recover little; on identity-label stripes, Qwen32 recovers nearly all of its loss and Qwen8 little. Later recovery $G_D=A_{DC}-A_{DD}$ against each model's joint loss $\Lambda=A_{CC}-A_{DD}$ at the three shared levels, joined in severity order (generation; organizer ANLS for DocVQA). Points on the dashed diagonal have $A_{DC}=A_{CC}$. Both axes share $A_{DD}$, so their sampling errors are positively correlated. Because $G_D=\Lambda-U_M$, a point on the diagonal can be either-sufficient or later-limited, and only $U_L$ separates the two. Bars are descriptive 95\% intervals for $G_D$; points with zero standard error have none, and loss intervals are omitted. Hollow points mark either no net loss observed or DeepStack's 360-pixel DocVQA level, which barely changes main. DeepStack is not native-capable on stripes or text.}
\label{fig:v15_curves}
\end{figure*}
\begin{table*}[t]
\centering\footnotesize
\setlength{\tabcolsep}{3pt}
\renewcommand{\arraystretch}{1.04}
\caption{\textbf{Single-entry losses separate main-limited, later-limited and either-sufficient cells.} Post hoc single-entry losses and item changes for selected cells of the calibrated experiment, recomputed from the stored model outputs without new inference. $U_M=A_{CC}-A_{DC}$ and $U_L=A_{CC}-A_{CD}$ carry paired intervals over templates (Student-$t$) or images and documents (normal); $\Lambda=G_D+U_M$. Changes count items whose score rises or falls from DD to DC: the 64 distinct template--object items for synthetic cells (duplicated recovery rows were checked to be identical) and all questions for benchmarks. C+P: case-and-punctuation matching.}
\label{tab:v15_posthoc}
\DSalign{appendix_posthoc-1}
\begin{tabular}{@{}llcccl@{}}
\toprule
\rowcolor{DSTableHeader}
Model & Cell & $\Lambda$ & $U_M$ & $U_L$ & Changes \\
\midrule
Qwen32 & Stripes 0.43, identity & \DSv{0.500}{} & \DSv{0.016}{[-0.016, 0.047]} & \DSv{0.000}{[0.000, 0.000]} & 32 up, 1 down of 64 \\
Qwen32 & Text 2.5, identity & \DSv{0.500}{} & \DSv{0.500}{[0.329, 0.671]} & \DSv{0.000}{[0.000, 0.000]} & 1 up, 1 down of 64 \\
Qwen32 & Stripes 0.43, direct & \DSv{0.500}{} & \DSv{0.016}{[-0.016, 0.047]} & \DSv{0.000}{[0.000, 0.000]} & 32 up, 1 down of 64 \\
Qwen8 & Stripes 0.43, direct & \DSv{0.500}{} & \DSv{0.422}{[0.269, 0.574]} & \DSv{0.031}{[-0.013, 0.076]} & 12 up, 7 down of 64 \\
Qwen32 & Text 2.5, direct & \DSv{0.406}{} & \DSv{0.406}{[0.258, 0.554]} & \DSv{0.000}{[0.000, 0.000]} & 1 up, 1 down of 64 \\
Qwen32 & Color 0.05, direct & \DSv{0.250}{} & \DSv{0.000}{[0.000, 0.000]} & \DSv{0.000}{[0.000, 0.000]} & 16 up, 0 down of 64 \\
DeepStack & Color 0.05, direct & \DSv{0.406}{} & \DSv{0.000}{[0.000, 0.000]} & \DSv{0.016}{[-0.016, 0.047]} & 26 up, 0 down of 64 \\
Granite & Position 0.3, direct & \DSv{0.453}{} & \DSv{0.047}{[-0.023, 0.117]} & \DSv{0.484}{[0.429, 0.540]} & 28 up, 2 down of 64 \\
Granite & Text 2.5, direct & \DSv{0.391}{} & \DSv{0.094}{[-0.003, 0.190]} & \DSv{0.391}{[0.315, 0.466]} & 21 up, 2 down of 64 \\
Granite & GQA attr.\ 5 px, C+P & \DSv{0.266}{} & \DSv{0.005}{[-0.033, 0.043]} & \DSv{0.198}{[0.143, 0.253]} & 116 up, 28 down of 349 \\
DeepStack & GQA attr.\ 5 px, C+P & \DSv{0.289}{} & \DSv{0.018}{[-0.000, 0.036]} & \DSv{0.042}{[0.005, 0.079]} & 112 up, 18 down of 349 \\
Qwen8 & DocVQA 180 px, ANLS & \DSv{0.516}{} & \DSv{0.498}{[0.446, 0.550]} & \DSv{0.025}{[0.002, 0.048]} & 22 up, 7 down of 373 \\
Qwen32 & DocVQA 180 px, ANLS & \DSv{0.461}{} & \DSv{0.454}{[0.404, 0.503]} & \DSv{-0.002}{[-0.029, 0.025]} & 30 up, 23 down of 373 \\
Granite & DocVQA 180 px, ANLS & \DSv{0.615}{} & \DSv{0.234}{[0.189, 0.280]} & \DSv{0.217}{[0.175, 0.260]} & 216 up, 14 down of 373 \\
DeepStack & DocVQA 180 px, ANLS & \DSv{0.185}{} & \DSv{0.019}{[-0.013, 0.051]} & \DSv{0.151}{[0.114, 0.188]} & 97 up, 16 down of 373 \\
\bottomrule
\end{tabular}
\end{table*}

\begin{table*}[t]
\centering\footnotesize
\setlength{\tabcolsep}{4pt}
\renewcommand{\arraystretch}{1.04}
\caption{\textbf{At similar loss, Qwen32 recovers more on stripes than on text, and more than Qwen8 on stripes.} Post hoc paired contrasts and the Granite DocVQA format check. Contrasts subtract within shared templates or documents before the interval is computed. The format check rescores the stored 180-pixel answers with the experiment's scorer: each row removes one kind of final mark before exact matching, C+P applies all three, and the last row gives the share of answers ending in a period.}
\label{tab:v15_posthoc_contrasts}
\DSalign{appendix_posthoc-2}
\begin{tabular}{@{}lcc@{}}
\toprule
\rowcolor{DSTableHeader}
Paired contrast & $G_D$ difference & $\Lambda$ difference \\
\midrule
Qwen32 stripes minus text, identity & \DSv{0.484}{[0.429, 0.540]} & \DSv{0.000}{[-0.171, 0.171]} \\
Qwen32 stripes minus text, direct & \DSv{0.484}{[0.429, 0.540]} & \DSv{0.094}{[-0.054, 0.242]} \\
Qwen32 minus Qwen8 stripes, direct & \DSv{0.406}{[0.266, 0.547]} & \DSv{0.000}{[-0.065, 0.065]} \\
Qwen32 minus Qwen8 DocVQA 180 px & \DSv{-0.011}{[-0.036, 0.014]} & \DSv{-0.055}{[-0.102, -0.009]} \\
\midrule
\rowcolor{DSTableHeader}
Granite DocVQA, 180 px & CC / CD / DC / DD & $I_{\rm rec}$ \\
\midrule
Organizer ANLS & $0.828 / 0.611 / 0.594 / 0.214$ & \DSv{0.163}{[0.104, 0.221]} \\
Lowercase exact & $0.307 / 0.049 / 0.159 / 0.008$ & \DSv{-0.107}{[-0.149, -0.064]} \\
Exact, final period removed & $0.839 / 0.523 / 0.523 / 0.128$ & \DSv{0.081}{[0.008, 0.153]} \\
Exact, final ! or ? removed & $0.310 / 0.049 / 0.159 / 0.008$ & \DSv{-0.109}{[-0.152, -0.067]} \\
Exact, outer quotes removed & $0.310 / 0.049 / 0.159 / 0.008$ & \DSv{-0.109}{[-0.152, -0.067]} \\
C+P exact (all three) & $0.841 / 0.523 / 0.523 / 0.128$ & \DSv{0.078}{[0.005, 0.151]} \\
Ends with a period & $0.682 / 0.914 / 0.776 / 0.888$ & \DSv{0.120}{[0.069, 0.170]} \\
\bottomrule
\end{tabular}
\end{table*}
\begin{table*}[t]
\centering\footnotesize
\setlength{\tabcolsep}{4pt}
\renewcommand{\arraystretch}{1.04}
\caption{\textbf{Degrading main raises later color score selectivity for every model, readout and level: all 36 $I_{\rm quality}$ intervals lie above zero.} Color conflict scores in the calibrated experiment. $R$ is the later target-minus-control effect on the candidate score with clean main; $I_{\rm quality}=R(\text{degraded main})-R(\text{clean main})$ at the three shared desaturation levels of Table~\ref{tab:v15_grid}, mild to severe. Intervals are descriptive, unadjusted 95\% Student-$t$ intervals over 32 templates. Every task, readout, kind, level and the target and control components are in \texttt{calibrated\_\allowbreak{}synthetic\_\allowbreak{}conflict.csv}.}
\label{tab:v15_conflict}
\DSalign{appendix_tables-1}
\begin{tabular}{@{}llcccc@{}}
\toprule
\rowcolor{DSTableHeader}
Model & Readout & $R$, clean main & $I_{\rm quality}$, level 1 & Level 2 & Level 3 \\
\midrule
Qwen8 & Direct & \DSv{0.242}{[0.145, 0.338]} & \DSv{1.921}{[1.615, 2.227]} & \DSv{2.535}{[2.215, 2.855]} & \DSv{2.658}{[2.275, 3.041]} \\
 & Identity & \DSv{-0.042}{[-0.138, 0.054]} & \DSv{1.864}{[1.278, 2.451]} & \DSv{2.654}{[1.833, 3.475]} & \DSv{1.771}{[0.418, 3.124]} \\
 & Cyclic & \DSv{-0.011}{[-0.179, 0.157]} & \DSv{1.731}{[1.358, 2.104]} & \DSv{2.584}{[1.729, 3.439]} & \DSv{1.856}{[0.863, 2.848]} \\
\addlinespace[2pt]
Qwen32 & Direct & \DSv{0.531}{[0.361, 0.702]} & \DSv{3.609}{[2.820, 4.398]} & \DSv{5.788}{[4.945, 6.632]} & \DSv{9.552}{[8.255, 10.850]} \\
 & Identity & \DSv{0.169}{[0.096, 0.242]} & \DSv{3.466}{[2.538, 4.393]} & \DSv{6.792}{[5.669, 7.914]} & \DSv{10.383}{[8.488, 12.278]} \\
 & Cyclic & \DSv{0.133}{[0.015, 0.251]} & \DSv{3.336}{[2.093, 4.578]} & \DSv{5.742}{[4.849, 6.636]} & \DSv{9.409}{[7.891, 10.927]} \\
\addlinespace[2pt]
Granite & Direct & \DSv{4.470}{[4.122, 4.819]} & \DSv{0.537}{[0.298, 0.776]} & \DSv{1.154}{[0.862, 1.445]} & \DSv{2.006}{[1.618, 2.394]} \\
 & Identity & \DSv{3.954}{[3.127, 4.781]} & \DSv{0.258}{[0.005, 0.511]} & \DSv{0.840}{[0.452, 1.229]} & \DSv{1.633}{[1.082, 2.184]} \\
 & Cyclic & \DSv{3.482}{[2.569, 4.394]} & \DSv{0.452}{[0.152, 0.753]} & \DSv{1.100}{[0.650, 1.550]} & \DSv{1.623}{[1.045, 2.202]} \\
\addlinespace[2pt]
DeepStack & Direct & \DSv{0.904}{[0.723, 1.085]} & \DSv{0.549}{[0.425, 0.673]} & \DSv{1.178}{[0.970, 1.386]} & \DSv{1.838}{[1.658, 2.017]} \\
 & Identity & \DSv{0.977}{[0.758, 1.196]} & \DSv{0.813}{[0.642, 0.984]} & \DSv{1.691}{[1.468, 1.913]} & \DSv{2.578}{[2.415, 2.741]} \\
 & Cyclic & \DSv{0.932}{[0.791, 1.073]} & \DSv{0.625}{[0.528, 0.722]} & \DSv{1.289}{[1.069, 1.508]} & \DSv{2.215}{[1.978, 2.452]} \\
\bottomrule
\end{tabular}
\end{table*}
\subsection{Execution, geometry, and scoring}
Inference ran in BF16 on A100 80GB GPUs, with the checkpoints of Table~\ref{tab:checkpoint} and the software versions of Appendix~\ref{app:v14}. Before analysis, every run passed four technical checks. Native and self-patched scores (each entry patched with its own source) differ by $0$, self-patched generations are identical, processed geometry is identical across all sources of a query, and every cell receives its intended visual source. The 609 observations computed on both machines were bitwise identical. Generation allows 32 new tokens for synthetic tasks and 64 for benchmarks.

The same file does not give every model the same input. Qwen8/32 process DocVQA pages at a median long side of 576 pixels and Granite at 1152 pixels, so a given level removes a larger share of Granite's processed resolution. DeepStack's main entry receives one 336-pixel global view, whereas its later groups receive crops from a larger canvas (Appendix~\ref{app:methods}). Its 360-pixel DocVQA level therefore barely changes main ($A_{DC}=A_{CC}=0.33$, $A_{CD}=A_{DD}=0.28$) and is not interpreted, and its lower levels remove more of the later input than of the main input. Apart from this main view, no evaluated level leaves a processed input unchanged, and processed geometry is identical across the clean and degraded sources of each query. Qwen8 and Qwen32 share one processed geometry yet lose different amounts on the synthetic tasks, so geometry alone does not account for differences in loss. Loss is therefore measured per model rather than inferred from the level.

Synthetic recovery cells count invalid answers as incorrect over all 128 planned queries, whereas conflict endpoints exclude invalid generations. Conflict and recovery cells also differ in what they change: conflict cells install source $b$ only on the edited object's later tokens, whereas recovery cells change all later tokens (Table~\ref{tab:intervention_spec}). The synthetic joint loss $\Lambda=A_{CC}-A_{DD}$ is a difference of two means and has no interval. Intervals for the other estimates follow Appendix~\ref{app:statistics}, and benchmark $\Lambda$, $G_D$, and $I_{\mathrm{rec}}$ are paired estimates clustered by image or document.

Benchmarks retain all 192 clusters in every cell. GQA is scored with case-and-punctuation normalization, the metric fixed before the runs, and DocVQA with organizer-framework ANLS; both score invalid outputs as zero. Official strict GQA matching is kept as a format-sensitive variant. The named sensitivities are lowercase-and-strip GQA matching, strict-cutoff ANLS and lowercase exact matching for DocVQA, and valid-four-cell and returned-string policies. Truncated generations are the only invalid benchmark outputs; they are most frequent for DeepStack on DocVQA, with at most six in one cell.

Granite and DeepStack capitalize nearly all GQA answers (``Right'' for the reference ``right''): 349 of 349 and 348 of 349 clean attribute answers, and all 340 clean spatial answers for each model, against none for Qwen8/32. Strict matching therefore scores their clean answers zero, and Granite's negative strict-match loss reflects a change of answer format under degradation. Case-and-punctuation normalization, which equals strict matching for Qwen8/32, is the comparable metric. At Granite's 5-pixel attribute level, lowercase-and-strip matching shows extra loss only in cells with degraded later input, and this extra loss disappears once trailing punctuation is removed, so the format effect is not counted as loss.

\subsection{Reading the curves}
On GQA, all later recovery and interaction means are positive, with 23 recovery and 21 interaction intervals above zero. These are 24 estimates per contrast (four models, three shared levels, and two question groups, attribute and spatial) under the preselected case-and-punctuation metric, with invalid outputs scored as zero. At the strongest shared levels (5-pixel long side for attributes, 10 for spatial questions), all eight intervals for each contrast are above zero. Conditions share images, so they do not count as independent replications.

Write the joint loss as $\Lambda=A_{CC}-A_{DD}$ and the single-entry losses as $U_M=A_{CC}-A_{DC}$ and $U_L=A_{CC}-A_{CD}$, so that $G_D=\Lambda-U_M$ and $I_{\mathrm{rec}}=\Lambda-U_M-U_L$. We do not single out one readout: the main-text figure shows the strongest level of selected cells, and Figure~\ref{fig:v15_curves} plots $G_D$ against $\Lambda$ for four tasks. Model--readout pairs that are not native-capable (defined above) count as non-native; they are listed but not interpreted: DeepStack on stripes, relative position, and text; Granite's direct and cyclic stripes; Qwen8's label readouts for relative position; and Granite's cyclic relative position. Where $A_{DD}=A_{CC}$, no net loss is observed to recover, although $G_D$ is still computed. At chance, two-choice accuracy can fall further only through systematically wrong answers, which occur here only in non-native readouts. Because $G_D=\Lambda-U_M$ and both axes share $A_{DD}$, a rising curve is partly definitional and its sampling errors are correlated, so Table~\ref{tab:v15_posthoc} gives the single-entry losses $U_M$ and $U_L$.

Several patterns depend on readout. Granite reads stripes near chance with direct answers but accurately with identity labels. There, degrading only its later input lowers accuracy about as far as degrading both ($A_{CD}$ from $0.50$ to $0.59$, $A_{DD}$ from $0.48$ to $0.58$), so $G_D$ is large while $I_{\mathrm{rec}}$ stays near zero. With the two label readouts, Qwen8's color loss reaches $0.47$ while $G_D$ stays at or below $0.08$, and Qwen32's $A_{DC}$ falls to $0.89$ and $0.88$ at the strongest desaturation. On text, Qwen8/32 show little recovery across readouts, with no positive gain resolved by the descriptive intervals and one label-readout interval below zero. Granite recovers part of its direct text loss but nearly all of its label-readout text losses, again with $A_{CD}$ near $A_{DD}$.

The single-entry losses $U_M$ and $U_L$ separate three patterns: main-limited (accuracy tracks main's quality alone), later-limited (it tracks later's alone), and either-sufficient (one clean entry suffices). Granite's relative-position, text, and identity-label stripe accuracy is later-limited, and its GQA attribute accuracy at 5 pixels is closer to later-limited. DeepStack on GQA is either-sufficient: $A_{CD}$ stays near $A_{CC}$, and $I_{\mathrm{rec}}$ intervals exclude zero at every level of both question groups.

On DocVQA, these labels depend on the metric. Granite's single-entry losses sum to less than the joint loss under both ANLS variants (organizer and strict cutoff), but under lowercase exact matching its interaction reverses at 180 pixels. Removing only a final period restores a positive interaction, whereas removing final exclamation or question marks, or outer quotes, leaves the reversal; Granite also ends more answers with a period when only later input is degraded than when both entries are clean (Table~\ref{tab:v15_posthoc_contrasts}). Normalized exact interactions are 0.11, 0.13, and 0.08 from mild to severe, so growth with loss is an ANLS description only, and pattern labels are therefore specific to the metric. Qwen32's main-limited text accuracy coexists with later effects on candidate scores: its score $I_{\mathrm{quality}}$ is $0.54$ [$0.30,0.77$] at the strongest level.

Generation conflict effects use a bounded answer indicator, so plateaus or zeros can reflect that bound. Recovery and $I_{\mathrm{quality}}$ measure different uses of later input and need not agree in direction. Likewise, each model's later input is a different intervention, with its own number, depth, and source of injected entries, so equal $G_D$ across models does not indicate use of the same information.

\subsection{Complete estimates}
Degrading main raises later color score selectivity for every model, readout, and shared level: all 36 $I_{\mathrm{quality}}$ intervals lie above zero (Table~\ref{tab:v15_conflict}, the color conflict scores that the main text cites). Tables~\ref{tab:v15_posthoc} and~\ref{tab:v15_posthoc_contrasts} add post hoc diagnostics (analyses not planned in advance), recomputed from the stored model outputs, and they are exploratory. The files \texttt{calibrated\_\allowbreak{}synthetic\_\allowbreak{}recovery.csv}, \texttt{calibrated\_\allowbreak{}synthetic\_\allowbreak{}conflict.csv}, \texttt{calibrated\_\allowbreak{}benchmarks.csv} and \texttt{calibrated\_\allowbreak{}capability\_\allowbreak{}levels.csv} (Appendix~\ref{app:data}) retain every model, task, readout, kind (score or generation), level, metric and invalid-answer policy of the calibrated experiment, including target and control conflict components and the four levels run outside the shared levels. Their values are the unrounded estimates behind the paper's tables, with nothing re-estimated. As elsewhere, if one interval excludes zero and another includes it, the pair is not read as a difference.

\section{Removing and coding entry features}
\label{app:coding}
This appendix gives the design and the complete results of the coding experiment in Section~\ref{sec:coding} of the main paper. We remove one entry's image information by replacing its features with zeros or a calibration mean, or we code each entry tensor with NNCodec (a neural-network compression codec) and count bitstream bytes. Main is the entry of visual tokens at the language-model input, later the entries added at deeper decoder layers, and a shared quantization parameter (QP) codes all entries at one QP (larger is coarser). Accuracy is in percent and differences are in percentage points (``points''). With later zeroed, Qwen8, Qwen32 and DeepStack lose at most 1.65 points of GQA attribute accuracy, whereas Granite keeps most of its accuracy from later alone only when main is replaced by its mean; five of six single-entry predictions fixed in advance hold (Appendix~\ref{app:coding_single}). Coding only DeepStack's main entry at QP~$-4$ takes 49.31\% fewer bytes than coding all entries at shared QP~$-2$, with a difference of $1.10$ [$-0.40,2.55$] points, in a comparison fixed after the first results on the new (confirmation) images (Appendix~\ref{app:coding_deepstack}). On held-out images, a fixed Granite configuration scores 2.64 points above a shared QP at nearly equal bytes; we credit this to the configuration and not to the role diagnosis (which entry limits accuracy), whose prediction fails (Appendix~\ref{app:coding_granite}). Qwen8/32 show no consistent advantage over a shared QP (Appendix~\ref{app:coding_qwen}).

\begin{table*}[t]
\centering\footnotesize
\setlength{\tabcolsep}{4pt}
\renewcommand{\arraystretch}{1.04}
\caption{\textbf{Which entry suffices depends on the model and the replacement.} GQA attribute accuracy on new images when an entry's features are replaced; Qwen32 DocVQA reports ANLS$\times$100 on all 474 remaining DocVQA documents. ``image'' keeps the entry's features unchanged; ``zeros'' sets all of them to zero; ``mean'' sets every position to the entry's mean feature vector over 48 calibration images (for DocVQA, the 56 page images of 48 calibration documents), so neither replacement depends on the test image. Qwen8, Qwen32 and DeepStack are evaluated on the 1,000 confirmation images. Granite's first 468 images come from the confirmation runs; its seven configurations were then run on the 531 remaining images, which Granite had not seen, and 999 pools the two. Values in percent. Intervals are paired bootstrap 95\% intervals over images (documents for DocVQA), 4,000 resamples.}
\label{tab:coding_corners}
\DSalign{table_coding_corners-1}
\begin{tabular}{@{}llcccc@{}}
\toprule
\rowcolor{DSTableHeader}
Main input & Later input & Qwen8 & Qwen32 & Granite, 999 & DeepStack \\
\midrule
\multicolumn{2}{@{}l}{\textit{Images or documents / questions}} & 1{,}000 / 1{,}838 & 1{,}000 / 1{,}838 & 999 / 1{,}836 & 1{,}000 / 1{,}838 \\
image & image & \DSv{72.50}{[70.40, 74.50]} & \DSv{72.75}{[70.60, 74.85]} & \DSv{73.27}{[71.17, 75.33]} & \DSv{79.50}{[77.60, 81.40]} \\
image & zeros & \DSv{72.20}{[70.10, 74.15]} & \DSv{71.90}{[69.70, 74.00]} & \DSv{51.60}{[49.15, 53.90]} & \DSv{77.85}{[75.80, 79.75]} \\
image & mean & \DSv{71.95}{[69.80, 73.90]} & \DSv{72.65}{[70.50, 74.65]} & \DSv{58.81}{[56.46, 61.11]} & \DSv{75.70}{[73.55, 77.70]} \\
zeros & image & \DSv{54.45}{[51.95, 56.85]} & \DSv{67.60}{[65.30, 69.80]} & \DSv{11.66}{[10.11, 13.26]} & \DSv{76.05}{[74.00, 78.00]} \\
mean & image & \DSv{50.50}{[48.00, 52.90]} & \DSv{65.40}{[63.10, 67.65]} & \DSv{69.77}{[67.57, 71.97]} & \DSv{29.35}{[27.30, 31.50]} \\
zeros & zeros & \DSv{31.55}{[29.30, 33.80]} & \DSv{43.20}{[40.85, 45.55]} & \DSv{1.00}{[0.60, 1.45]} & \DSv{11.45}{[10.00, 13.05]} \\
mean & mean & \DSv{43.10}{[40.80, 45.50]} & \DSv{43.30}{[40.90, 45.60]} & \DSv{42.29}{[39.94, 44.54]} & \DSv{10.35}{[8.90, 11.85]} \\
\midrule
\rowcolor{DSTableHeader}
Main input & Later input & Granite, 531 & Granite, 468 & Qwen32 DocVQA &  \\
\multicolumn{2}{@{}l}{\textit{Images or documents / questions}} & 531 / 970 & 468 / 866 & 474 / 907 &  \\
image & image & \DSv{74.86}{[72.03, 77.68]} & \DSv{71.47}{[68.38, 74.57]} & \DSv{78.64}{[76.09, 81.10]} &  \\
image & zeros & \DSv{53.11}{[49.81, 56.50]} & \DSv{49.89}{[46.58, 53.31]} & \DSv{76.05}{[73.30, 78.64]} &  \\
image & mean & \DSv{59.89}{[56.69, 63.28]} & \DSv{57.59}{[54.06, 61.11]} & \DSv{76.99}{[74.38, 79.47]} &  \\
zeros & image & \DSv{12.24}{[10.17, 14.41]} & \DSv{11.00}{[8.87, 13.25]} & \DSv{7.09}{[5.57, 8.58]} &  \\
mean & image & \DSv{71.19}{[68.27, 74.11]} & \DSv{68.16}{[64.85, 71.47]} & \DSv{4.04}{[2.85, 5.31]} &  \\
zeros & zeros & \DSv{0.75}{[0.28, 1.41]} & \DSv{1.28}{[0.64, 2.03]} & \DSv{6.60}{[5.22, 7.98]} &  \\
mean & mean & \DSv{42.56}{[39.55, 45.76]} & \DSv{41.99}{[38.57, 45.51]} & \DSv{5.34}{[4.07, 6.71]} &  \\
\bottomrule
\end{tabular}
\end{table*}
\begin{table*}[t]
\centering\footnotesize
\setlength{\tabcolsep}{3.5pt}
\renewcommand{\arraystretch}{1.04}
\caption{\textbf{Paired differences between single-entry configurations.} Percentage points (ANLS$\times$100 points for DocVQA), same images and replacements as Table~\ref{tab:coding_corners}. ``Main kept minus later kept'' compares keeping only main with keeping only the later entries when the removed entry is replaced by its mean or by zeros. The kept entries differ in raw size, so this is not a comparison at equal bytes. Status: L, fixed in the protocol before any confirmation image was scored; F, follow-up comparison fixed after the first confirmation results and before the follow-up runs; P, post hoc. In this table the first five rows for Granite on all three image sets are F (the follow-up stated the mean comparison with the opposite sign), and every other entry is P. Intervals are paired bootstrap 95\% intervals over images (documents for DocVQA), 4,000 resamples.}
\label{tab:coding_differences}
\DSalign{table_coding_differences-1}
\begin{tabular}{@{}lcccc@{}}
\toprule
\rowcolor{DSTableHeader}
Comparison & Qwen8 & Qwen32 & Granite, 999 & DeepStack \\
\midrule
Later zeroed minus full & \DSv{-0.30}{[-1.55, 0.95]} & \DSv{-0.85}{[-2.25, 0.50]} & \DSv{-21.67}{[-24.27, -19.22]} & \DSv{-1.65}{[-3.05, -0.20]} \\
Later mean-replaced minus full & \DSv{-0.55}{[-1.80, 0.65]} & \DSv{-0.10}{[-1.40, 1.15]} & \DSv{-14.46}{[-16.82, -12.06]} & \DSv{-3.80}{[-5.35, -2.35]} \\
Main zeroed minus full & \DSv{-18.05}{[-20.55, -15.70]} & \DSv{-5.15}{[-6.95, -3.40]} & \DSv{-61.61}{[-64.02, -59.16]} & \DSv{-3.45}{[-4.80, -2.10]} \\
Main mean-replaced minus full & \DSv{-22.00}{[-24.50, -19.50]} & \DSv{-7.35}{[-9.20, -5.55]} & \DSv{-3.50}{[-5.16, -1.85]} & \DSv{-50.15}{[-52.75, -47.55]} \\
Main kept minus later kept, mean & \DSv{21.45}{[18.85, 24.05]} & \DSv{7.25}{[5.35, 9.10]} & \DSv{-10.96}{[-13.51, -8.46]} & \DSv{46.35}{[43.60, 49.05]} \\
Main kept minus later kept, zeros & \DSv{17.75}{[15.35, 20.20]} & \DSv{4.30}{[2.40, 6.15]} & \DSv{39.94}{[37.49, 42.49]} & \DSv{1.80}{[0.20, 3.40]} \\
\midrule
\rowcolor{DSTableHeader}
Comparison & Granite, 531 & Granite, 468 & & \\
Later zeroed minus full & \DSv{-21.75}{[-25.14, -18.27]} & \DSv{-21.58}{[-25.32, -17.84]} &  &  \\
Later mean-replaced minus full & \DSv{-14.97}{[-18.27, -11.58]} & \DSv{-13.89}{[-17.41, -10.47]} &  &  \\
Main zeroed minus full & \DSv{-62.62}{[-66.10, -59.23]} & \DSv{-60.47}{[-64.00, -56.94]} &  &  \\
Main mean-replaced minus full & \DSv{-3.67}{[-5.93, -1.41]} & \DSv{-3.31}{[-5.77, -0.85]} &  &  \\
Main kept minus later kept, mean & \DSv{-11.30}{[-14.60, -7.91]} & \DSv{-10.58}{[-14.10, -6.94]} &  &  \\
Main kept minus later kept, zeros & \DSv{40.87}{[37.48, 44.35]} & \DSv{38.89}{[35.15, 42.52]} &  &  \\
\midrule
\rowcolor{DSTableHeader}
Other comparison & Model, images & & & \\
Main zeroed minus both mean & Granite, 999 & \DSv{-30.63}{[-33.13, -28.03]} & & \\
Later zeroed minus full & Qwen32 DocVQA & \DSv{-2.58}{[-4.11, -1.14]} & & \\
\bottomrule
\end{tabular}
\end{table*}
\begin{table*}[t]
\centering\footnotesize
\setlength{\tabcolsep}{4pt}
\renewcommand{\arraystretch}{1.04}
\caption{\textbf{Single-entry predictions fixed in the protocol: five of six hold.} Each prediction states that keeping one entry (the other replaced by zeros) loses at most a fraction $f$ of the full visual gain, the accuracy of the full input minus that with both entries zeroed. Loss/gain is the observed loss as a fraction of that gain. The margin, in percentage points, is $f$ times the gain minus the loss; a prediction holds when the whole interval lies above zero. Granite's prediction fails on its first 468 images, on the 531 held-out images and on both together. The four predictions without data (Qwen8 on GQA spatial and on DocVQA, main kept, $f=0.25$; DeepStack on GQA spatial, each entry kept, $f=0.5$) belong to cells that were not run. Status codes and intervals as in Table~\ref{tab:coding_differences}.}
\label{tab:coding_endpoints}
\DSalign{table_coding_endpoints-1}
\begin{tabular}{@{}llrcccll@{}}
\toprule
\rowcolor{DSTableHeader}
Model & Entry kept & Images & $f$ & Loss/gain & Margin & Outcome & Status \\
\midrule
Qwen8 & main & \DSn{1{,}000} & \DSv{0.25}{} & \DSv{0.007}{} & \DSv{9.94}{[8.68, 11.24]} & holds & L \\
Qwen32 & main & \DSn{1{,}000} & \DSv{0.50}{} & \DSv{0.029}{} & \DSv{13.93}{[12.20, 15.55]} & holds & L \\
Qwen32 & later & \DSn{1{,}000} & \DSv{0.50}{} & \DSv{0.174}{} & \DSv{9.63}{[7.78, 11.53]} & holds & L \\
Granite & later & \DSn{468} & \DSv{0.25}{} & \DSv{0.861}{} & \DSv{-42.92}{[-45.91, -39.90]} & fails & L \\
Granite & later & \DSn{531} & \DSv{0.25}{} & \DSv{0.845}{} & \DSv{-44.09}{[-46.99, -41.17]} & fails & F \\
Granite & later & \DSn{999} & \DSv{0.25}{} & \DSv{0.852}{} & \DSv{-43.54}{[-45.58, -41.44]} & fails & F \\
DeepStack & main & \DSn{1{,}000} & \DSv{0.50}{} & \DSv{0.024}{} & \DSv{32.38}{[30.73, 34.00]} & holds & L \\
DeepStack & later & \DSn{1{,}000} & \DSv{0.50}{} & \DSv{0.051}{} & \DSv{30.58}{[28.90, 32.18]} & holds & L \\
\bottomrule
\end{tabular}
\end{table*}
\begin{figure*}[t]
\centering
\includegraphics[width=\textwidth]{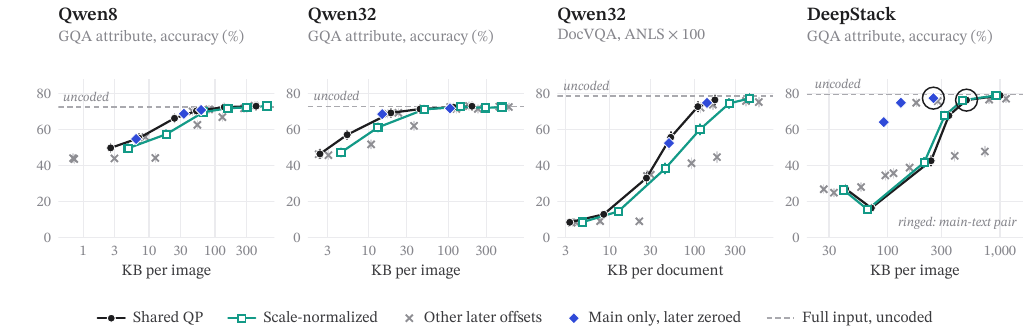}
\caption{\textbf{Accuracy against coded bytes on the confirmation images.} Each point is one configuration: GQA attribute accuracy on 1,000 images, or ANLS$\times$100 on 474 DocVQA documents, against NNCodec bytes per image or document on a log scale. Shared QP codes main and every later entry at one QP; scale-normalized codes later entry $i$ at QP $q+d_i$; other later offsets code the later entries coarser or finer than scale-normalized; main only codes main and sets the later entries to zero (0 bytes). Thin bars are paired bootstrap 95\% intervals over images or documents; the dashed line is the uncoded full input. Rings mark the DeepStack pair reported in Section~\ref{sec:coding} (main only at QP~$-4$, shared QP~$-2$). DeepStack's values are in Table~\ref{tab:coding_deepstack}, and the Qwen main-only and scale-normalized values in Table~\ref{tab:coding_qwen}. Every point, including Granite's configurations on 468 images, is listed in \texttt{coding\_\allowbreak{}configurations.csv}.}
\label{fig:coding_rate}
\end{figure*}

\subsection{Design and units}
\label{app:coding_design}
\textbf{What is coded.} Each entry hands the language model one feature tensor per image, taken after the vision encoder and the projector or merger, with one row per visual token. Qwen8 and Qwen32 have one main and three later tensors (at most 256 tokens of width 4,096 or 5,120), Granite one main and seven later tensors (344 to 1,032 tokens of width 2,560 on the calibration images, depending on its tiling), and DeepStack one main and two or four later tensors, depending on image geometry (576 tokens of width 4,096 each). Decoded features are cast to bfloat16 (BF16) and enter through the native entries (Appendix~\ref{app:methods}).

\textbf{Codec and configurations.} The codec is NNCodec 2.1.3 \citep{becking2023nncodec}, an implementation of the ISO/IEC Neural Network Coding (NNC) standard \citep{kirchhoffer2022nnr}, with plain uniform quantization (no dependent or rate-distortion-optimized quantization) and its entropy coder. Each tensor is coded as its own bitstream at exactly the requested QP. Larger QP is coarser: the step size is 1 at QP~0, and four QP steps double it. A configuration is written main\,|\,later. An integer is the QP of main or of every later entry, so $q\,|\,q$ is a shared QP, the setting in which Heidari et al.\ \citep{heidari2026compressing} code Qwen3-VL's four entry tensors. Z sets an entry to zeros. In the scale-normalized setting n, each later entry's step size follows its feature scale relative to main's, up to rounding. Later entry $i$ is coded at QP~$q+d_i$, where $q$ is main's QP and $d_i=\mathrm{round}(4\log_2(\sigma_i/\sigma_{\main}))$ compares the standard deviation of the entry's elements with main's on 48 calibration images. The offsets $d_i$ are $-4$, $-3$, $-3$ for Qwen8, $-5$, $-2$, $-7$ for Qwen32, $-12$ to $-6$ for Granite and $0$, $0$, $+1$, $+1$ for DeepStack. The settings n$+k$ and n$-k$ code every later entry $k$ QP units coarser or finer than n. The shift $k$ is relative to n and not to main: Granite's fixed configuration $0\,|\,\mathrm{n}{+}4$ still codes every later entry more finely than main (QP~$-8$ to $-2$ against 0).

\textbf{Bytes and replacements.} Bytes are the actual NNCodec bitstream bytes of the coded tensors, headers included; each question takes the bytes of its image or page, which are averaged within each image or document and then over units, and 1~KB is 1,000 bytes. A removed entry counts zero bytes, unchanged features count as uncoded BF16 (2 bytes per element), inactive DeepStack groups are never counted, and the mean vectors and offset tables are fixed side information that is not counted. ``Mean'' sets every position of an entry to one vector, the per-image token mean averaged over 48 calibration images (for DocVQA, the 56 page images of 48 calibration documents; for DeepStack's third and fourth later groups, the 24 of these images where the groups are active). Neither replacement depends on the test image.

\textbf{Data, scoring and statistics.} Test units come from the GQA validation-balanced attribute questions and the Single-Document DocVQA validation set. After excluding every image or document used in our other experiments, we ordered the remaining units by a fixed-seed hash and took them in that order with at most two questions per unit. This gives 1,000 GQA images with 1,838 questions (the confirmation images) and all 474 remaining DocVQA documents with 907 questions on 518 pages. Calibration uses the 240 units per task of the calibrated experiment (Appendix~\ref{app:v15}), which has 48 calibration and 192 evaluation units. The 48 calibration units give the mean vectors, the 48 GQA attribute calibration images give the offsets $d_i$ for all tasks, and all 240 units select configurations. Generation is greedy with at most 64 new tokens, as in Appendix~\ref{app:v14}, at batch size 1. GQA answers are scored by exact match after case-and-punctuation normalization (Appendix~\ref{app:v15}) and DocVQA answers by ANLS (a soft string-match score) in the organizers' framework, times 100; invalid outputs score zero, and scores are averaged within each image or document and then over units. Intervals are 95\% percentile intervals from a paired bootstrap over images (documents for DocVQA) with 4,000 resamples. The resampling weights are drawn once per image set and shared by every estimate on that set, so differences use identical resamples. A prediction holds when its whole interval lies on the predicted side of zero.

\begin{table*}[t]
\centering\footnotesize
\setlength{\tabcolsep}{4pt}
\renewcommand{\arraystretch}{1.04}
\caption{\textbf{DeepStack: every coded configuration on the 1,000 confirmation images.} GQA attribute accuracy in percent; KB per image (1 KB = 1,000 bytes of NNCodec bitstream, zeroed entries count 0). Configurations are written main\,|\,later: an integer is the QP of main or of every later entry (shared QP); n is the scale-normalized setting, in which later entry $i$ is coded at QP $q+d_i$ with $d_i$ its rounded scale offset (0, 0, +1, +1 for DeepStack); n+$k$ and n$-k$ code the later entries $k$ QP units coarser or finer than n. Larger QP is coarser, and four QP steps double the quantization step. Z sets all later features to zero. $^{\ast}$Selected on calibration images at the high budget. The uncoded full input scores $79.50$\%. Status codes and intervals as in Table~\ref{tab:coding_differences}.}
\label{tab:coding_deepstack}
\DSalign{table_coding_deepstack-1}
\begin{tabular}{@{}lcrrrc@{}}
\toprule
\rowcolor{DSTableHeader}
Configuration & Status & KB total & KB main & KB later & Accuracy \\
\midrule
\rowcolor{DSTableAccent}\multicolumn{6}{@{}l}{\textbf{Shared QP}}\\
$-4\,|\,{-4}$ & L & \DSv{1{,}002.32}{} & \DSv{255.08}{} & \DSv{747.24}{} & \DSv{78.95}{[77.00, 80.90]} \\
$-2\,|\,{-2}^{\ast}$ & L & \DSv{503.26}{} & \DSv{131.07}{} & \DSv{372.19}{} & \DSv{76.35}{[74.30, 78.35]} \\
$-1\,|\,{-1}$ & L & \DSv{350.29}{} & \DSv{92.11}{} & \DSv{258.18}{} & \DSv{67.75}{[65.50, 70.05]} \\
$0\,|\,0$ & L & \DSv{243.13}{} & \DSv{64.28}{} & \DSv{178.86}{} & \DSv{42.65}{[39.90, 45.25]} \\
$2\,|\,2$ & L & \DSv{71.43}{} & \DSv{17.80}{} & \DSv{53.62}{} & \DSv{16.40}{[14.50, 18.30]} \\
$4\,|\,4$ & L & \DSv{42.12}{} & \DSv{9.53}{} & \DSv{32.60}{} & \DSv{27.05}{[24.90, 29.35]} \\
\rowcolor{DSTableAccent}\multicolumn{6}{@{}l}{\textbf{Scale-normalized}}\\
$-4\,|\,\mathrm{n}$ & L & \DSv{924.20}{} & \DSv{255.08}{} & \DSv{669.12}{} & \DSv{78.95}{[77.00, 80.95]} \\
$-2\,|\,\mathrm{n}^{\ast}$ & L & \DSv{462.14}{} & \DSv{131.07}{} & \DSv{331.06}{} & \DSv{76.25}{[74.20, 78.25]} \\
$-1\,|\,\mathrm{n}$ & L & \DSv{321.71}{} & \DSv{92.11}{} & \DSv{229.60}{} & \DSv{67.70}{[65.45, 70.05]} \\
$0\,|\,\mathrm{n}$ & L & \DSv{211.39}{} & \DSv{64.28}{} & \DSv{147.11}{} & \DSv{41.75}{[39.05, 44.40]} \\
$2\,|\,\mathrm{n}$ & L & \DSv{66.16}{} & \DSv{17.80}{} & \DSv{48.35}{} & \DSv{15.55}{[13.70, 17.40]} \\
$4\,|\,\mathrm{n}$ & L & \DSv{40.30}{} & \DSv{9.53}{} & \DSv{30.77}{} & \DSv{26.45}{[24.30, 28.70]} \\
\rowcolor{DSTableAccent}\multicolumn{6}{@{}l}{\textbf{Later coarser than scale-normalized}}\\
$-2\,|\,\mathrm{n}{+}2^{\ast}$ & L & \DSv{278.19}{} & \DSv{131.07}{} & \DSv{147.11}{} & \DSv{75.80}{[73.70, 77.85]} \\
$-2\,|\,\mathrm{n}{+}4$ & L & \DSv{179.43}{} & \DSv{131.07}{} & \DSv{48.35}{} & \DSv{74.90}{[72.80, 77.05]} \\
$0\,|\,\mathrm{n}{+}2$ & L & \DSv{112.63}{} & \DSv{64.28}{} & \DSv{48.35}{} & \DSv{35.70}{[33.10, 38.30]} \\
$0\,|\,\mathrm{n}{+}4$ & L & \DSv{95.05}{} & \DSv{64.28}{} & \DSv{30.77}{} & \DSv{34.50}{[31.95, 37.10]} \\
$4\,|\,\mathrm{n}{+}2$ & L & \DSv{33.16}{} & \DSv{9.53}{} & \DSv{23.64}{} & \DSv{24.90}{[22.75, 27.05]} \\
$4\,|\,\mathrm{n}{+}4$ & L & \DSv{26.91}{} & \DSv{9.53}{} & \DSv{17.38}{} & \DSv{26.85}{[24.70, 29.20]} \\
\rowcolor{DSTableAccent}\multicolumn{6}{@{}l}{\textbf{Later finer than scale-normalized}}\\
$-2\,|\,\mathrm{n}{-}4$ & L & \DSv{1{,}132.93}{} & \DSv{131.07}{} & \DSv{1{,}001.86}{} & \DSv{77.15}{[75.10, 79.20]} \\
$-2\,|\,\mathrm{n}{-}2$ & L & \DSv{800.19}{} & \DSv{131.07}{} & \DSv{669.12}{} & \DSv{76.60}{[74.50, 78.65]} \\
$0\,|\,\mathrm{n}{-}4$ & L & \DSv{733.39}{} & \DSv{64.28}{} & \DSv{669.12}{} & \DSv{47.80}{[45.00, 50.65]} \\
$0\,|\,\mathrm{n}{-}2^{\ast}$ & L & \DSv{395.34}{} & \DSv{64.28}{} & \DSv{331.06}{} & \DSv{45.35}{[42.55, 48.10]} \\
$4\,|\,\mathrm{n}{-}4$ & L & \DSv{156.64}{} & \DSv{9.53}{} & \DSv{147.11}{} & \DSv{38.85}{[36.30, 41.40]} \\
$4\,|\,\mathrm{n}{-}2$ & L & \DSv{57.88}{} & \DSv{9.53}{} & \DSv{48.35}{} & \DSv{28.05}{[25.80, 30.35]} \\
\rowcolor{DSTableAccent}\multicolumn{6}{@{}l}{\textbf{Main only, later entries zeroed}}\\
$-4\,|\,\mathrm{Z}$ & F & \DSv{255.08}{} & \DSv{255.08}{} & \DSv{0.00}{} & \DSv{77.45}{[75.40, 79.45]} \\
$-2\,|\,\mathrm{Z}$ & F & \DSv{131.07}{} & \DSv{131.07}{} & \DSv{0.00}{} & \DSv{74.90}{[72.85, 77.05]} \\
$-1\,|\,\mathrm{Z}$ & F & \DSv{92.11}{} & \DSv{92.11}{} & \DSv{0.00}{} & \DSv{64.15}{[61.75, 66.60]} \\
\bottomrule
\end{tabular}
\end{table*}

\textbf{What was fixed in advance.} Before any confirmation image was scored, a protocol fixed 31 configurations for each of four models and three tasks (GQA attribute, GQA spatial and DocVQA): the seven single-entry configurations of Table~\ref{tab:coding_corners} and four families of six coded configurations (shared QP, scale-normalized, later coarser than n, later finer than n). Until then, the confirmation units had been used only for feature extraction and coding. The protocol also fixed ten single-entry predictions, the role-direction predictions and a selection rule. The rule defines a high and a low budget, namely the calibration bytes of the cheapest shared-QP configuration that reaches 95\% or 60\% of the way from both entries zeroed to the full input on calibration units, and in each family it takes the configuration with the highest calibration score within each budget.

Of the twelve cells (four models, three tasks), four were completed: Qwen8, Qwen32 and DeepStack on attribute questions, and Qwen32 on DocVQA. We ended the confirmation runs after seeing the first confirmation results, when Granite had 468 complete attribute images (its first 468 images, as in the tables below) and the other seven cells had not been run, so no model has a GQA spatial result. We then fixed six follow-up runs and their comparisons before running them. Four ran main alone with later zeroed, each compared with the shared QP at the same main QP: DeepStack, Qwen8 and Qwen32 on attribute questions, and Qwen32 on DocVQA. Their fixed comparisons also include DeepStack $-4\,|\,\mathrm{Z}$ against $-2\,|\,{-2}$, $-2\,|\,\mathrm{Z}$ against $-2\,|\,\mathrm{n}{+}2$ and Qwen8 $0\,|\,\mathrm{Z}$ against $2\,|\,2$. The other two ran on the 531 confirmation images that Granite had not seen: Granite's seven single-entry configurations, and the fixed Granite pair of coded configurations (Appendix~\ref{app:coding_granite}). Nothing was added or reselected from these results. Tables mark each estimate as fixed in the protocol (L), a fixed follow-up (F) or post hoc (P), that is, chosen or computed after the results were seen. The Qwen32 byte decomposition and the training comparisons with plain fine-tuning are also post hoc.

\textbf{Runs.} Confirmation and follow-up runs used BF16 on NVIDIA A100 80GB GPUs, with the software versions of Appendix~\ref{app:v14}. An unmodified pass through each model's adapter matched the native model, and our GPU implementation of the quantizer reproduced NNCodec's decoded values exactly for all four models. Exploratory runs used BF16 on an Apple-silicon GPU and the same 240 calibration units; we cite them as exploratory and do not pool them with confirmation results.

\textbf{What these results do not show.} No configuration was tested for near-lossless coding, and bytes count coded features, not end-to-end latency or memory, because zeroing replaces a tensor and leaves the branch in place. The comparisons do not establish a general role-based coding method, an allocation that needs no calibration or is optimal, or an advantage over a generic configuration search, and no rule for splitting bytes between entries follows from them. Training shows no role-specific benefit. The results also do not say which entry to transmit first under limited bandwidth: Granite's two single-entry options differ sevenfold in raw payload.

\begin{table*}[t]
\centering\footnotesize
\setlength{\tabcolsep}{4pt}
\renewcommand{\arraystretch}{1.04}
\caption{\textbf{DeepStack paired comparisons on the 1,000 confirmation images.} Byte change is the KB of the first configuration relative to the second, in percent; the difference is in accuracy points. Notation as in Table~\ref{tab:coding_deepstack}. The protocol predicted no role direction for this model and task, so the comparisons of the selected configuration carry no predicted sign. Status codes and intervals as in Table~\ref{tab:coding_differences}.}
\label{tab:coding_deepstack_pairs}
\DSalign{table_coding_deepstack_pairs-1}
\begin{tabular}{@{}lrrrcl@{}}
\toprule
\rowcolor{DSTableHeader}
Comparison & KB, first & KB, second & Byte change (\%) & Difference & Status \\
\midrule
\rowcolor{DSTableAccent}\multicolumn{6}{@{}l}{\textbf{Main only minus shared QP at the same main QP}}\\
$-4\,|\,\mathrm{Z}$ minus $-4\,|\,{-4}$ & \DSv{255.08}{} & \DSv{1{,}002.32}{} & \DSv{-74.55}{} & \DSv{-1.50}{[-2.90, -0.15]} & F \\
$-2\,|\,\mathrm{Z}$ minus $-2\,|\,{-2}$ & \DSv{131.07}{} & \DSv{503.26}{} & \DSv{-73.95}{} & \DSv{-1.45}{[-2.95, 0.05]} & F \\
$-1\,|\,\mathrm{Z}$ minus $-1\,|\,{-1}$ & \DSv{92.11}{} & \DSv{350.29}{} & \DSv{-73.70}{} & \DSv{-3.60}{[-5.35, -1.90]} & F \\
\rowcolor{DSTableAccent}\multicolumn{6}{@{}l}{\textbf{Across main QP}}\\
$-4\,|\,\mathrm{Z}$ minus $-2\,|\,{-2}$ & \DSv{255.08}{} & \DSv{503.26}{} & \DSv{-49.31}{} & \DSv{1.10}{[-0.40, 2.55]} & F \\
\rowcolor{DSTableAccent}\multicolumn{6}{@{}l}{\textbf{Configuration selected on calibration images}}\\
$-2\,|\,\mathrm{n}{+}2$ minus $-2\,|\,\mathrm{n}$ & \DSv{278.19}{} & \DSv{462.14}{} & \DSv{-39.80}{} & \DSv{-0.45}{[-1.40, 0.50]} & L \\
$-2\,|\,\mathrm{n}{+}2$ minus $-2\,|\,{-2}$ & \DSv{278.19}{} & \DSv{503.26}{} & \DSv{-44.72}{} & \DSv{-0.55}{[-1.55, 0.45]} & L \\
$-2\,|\,\mathrm{Z}$ minus $-2\,|\,\mathrm{n}{+}2$ & \DSv{131.07}{} & \DSv{278.19}{} & \DSv{-52.88}{} & \DSv{-0.90}{[-2.25, 0.45]} & F \\
\rowcolor{DSTableAccent}\multicolumn{6}{@{}l}{\textbf{Post hoc}}\\
$-4\,|\,\mathrm{Z}$ minus $-2\,|\,\mathrm{n}$ & \DSv{255.08}{} & \DSv{462.14}{} & \DSv{-44.80}{} & \DSv{1.20}{[-0.25, 2.65]} & P \\
\bottomrule
\end{tabular}
\end{table*}
\begin{table*}[t]
\centering\footnotesize
\setlength{\tabcolsep}{4pt}
\renewcommand{\arraystretch}{1.04}
\caption{\textbf{Granite: the fixed pair on held-out images and the scale-normalized configuration with the same later setting.} GQA attribute accuracy in percent, KB per image. The pair $0\,|\,\mathrm{n}{+}4$ versus $-4\,|\,{-4}$ was found on Granite's first 468 images and fixed before the 531 held-out images were scored; the 468 images are discovery data and are not pooled with the 531. Granite's scale offsets are $-12$ to $-6$, so in $0\,|\,\mathrm{n}{+}4$ every later entry is coded finer than main. $4\,|\,\mathrm{n}$ codes the later entries exactly as $0\,|\,\mathrm{n}{+}4$ does and main four QP units coarser. It was not run on the 531 images. Notation as in Table~\ref{tab:coding_deepstack}. Status codes and intervals as in Table~\ref{tab:coding_differences}.}
\label{tab:coding_granite}
\DSalign{table_coding_granite-1}
\begin{tabular}{@{}llrrrc@{}}
\toprule
\rowcolor{DSTableHeader}
Images & Configuration & KB main & KB later & KB total & Accuracy \\
\midrule
\rowcolor{DSTableAccent}\multicolumn{6}{@{}l}{\textbf{Configurations}}\\
531 held-out & $0\,|\,\mathrm{n}{+}4$ & \DSv{187.23}{} & \DSv{208.40}{} & \DSv{395.63}{} & \DSv{69.30}{[66.20, 72.32]} \\
531 held-out & $-4\,|\,{-4}$ & \DSv{305.90}{} & \DSv{86.99}{} & \DSv{392.89}{} & \DSv{66.67}{[63.84, 69.77]} \\
468 discovery & $0\,|\,\mathrm{n}{+}4$ & \DSv{189.32}{} & \DSv{209.77}{} & \DSv{399.09}{} & \DSv{66.56}{[63.25, 69.87]} \\
468 discovery & $-4\,|\,{-4}$ & \DSv{309.18}{} & \DSv{87.35}{} & \DSv{396.53}{} & \DSv{63.14}{[59.72, 66.56]} \\
468 discovery & $4\,|\,\mathrm{n}$ & \DSv{63.38}{} & \DSv{209.77}{} & \DSv{273.15}{} & \DSv{65.60}{[62.29, 68.91]} \\
\rowcolor{DSTableAccent}\multicolumn{6}{@{}l}{\textbf{Paired differences}}\\
\rowcolor{DSTableHeader}
Images & \multicolumn{2}{l}{Comparison} & Status & Byte change (\%) & Difference \\
531 held-out & \multicolumn{2}{l}{$0\,|\,\mathrm{n}{+}4$ minus $-4\,|\,{-4}$} & F & \DSv{0.70}{} & \DSv{2.64}{[0.47, 4.80]} \\
468 discovery & \multicolumn{2}{l}{$0\,|\,\mathrm{n}{+}4$ minus $-4\,|\,{-4}$} & P & \DSv{0.65}{} & \DSv{3.42}{[1.28, 5.45]} \\
468 discovery & \multicolumn{2}{l}{$4\,|\,\mathrm{n}$ minus $-4\,|\,{-4}$} & P & \DSv{-31.12}{} & \DSv{2.46}{[0.11, 4.81]} \\
468 discovery & \multicolumn{2}{l}{$0\,|\,\mathrm{n}{+}4$ minus $4\,|\,\mathrm{n}$} & P & \DSv{46.11}{} & \DSv{0.96}{[-0.96, 2.88]} \\
\bottomrule
\end{tabular}
\end{table*}

\subsection{Single-entry accuracy and predictions fixed in advance}
\label{app:coding_single}
Which entry suffices depends on the model and on the replacement (Tables~\ref{tab:coding_corners} and~\ref{tab:coding_differences}). With later zeroed, Qwen8, Qwen32 and DeepStack lose at most 1.65 points relative to the full input. Granite loses 14.46 to 21.67 points without its later entries, yet with later alone it scores 69.77\% against 73.27\% when main is set to its mean ($-3.50$ [$-5.16,-1.85$]) and only 11.66\% when main is zeroed. DeepStack depends on the replacement value in the opposite way, scoring 76.05\% with main zeroed and 29.35\% with main set to its mean. Under mean replacement, keeping main rather than later gains 21.45 points for Qwen8, whereas keeping later rather than main gains 10.96 points for Granite. The 999 Granite images pool its first 468 images and the 531 held-out images, and the two groups agree on this difference (Table~\ref{tab:coding_differences}).

Six of the ten single-entry predictions have data, and five hold (Table~\ref{tab:coding_endpoints}). All use zeros as the replacement and state that the kept entry alone loses at most a given fraction of the visual gain, the full input's accuracy minus that with both entries zeroed. Granite's prediction for later alone fails on all three image sets, and on the 999 images later alone loses 0.85 of the visual gain. The protocol kept zeros for this prediction even though it had already failed on the calibration units (0.89 of the gain) and the mean replacement had scored higher there.

A separate rule, written after the exploratory results for these two cells were known and not adopted by the protocol, picks the single-entry option (main or later kept, the other zeroed or set to its mean) with the highest calibration accuracy. On the 48 calibration images it selects main with later zeroed for Qwen8 (68.75\%) and later with main set to its mean for Granite (72.92\%). Both are also the best of the four options on the new images (Table~\ref{tab:coding_corners}), an agreement that is not a test.

The single-entry comparison asks which entry a model needs, not how many bytes it costs. Granite's seven later tensors together hold seven times main's feature elements, yet after coding at a shared QP main takes most of the bytes (305.90 of 392.89~KB at $-4\,|\,{-4}$ on the 531 images, Table~\ref{tab:coding_granite}). On DocVQA, Qwen32 scores 76.05 against 78.64 ANLS points with later zeroed ($-2.58$ [$-4.11,-1.14$]), whereas with main zeroed it scores 7.09, close to the 6.60 with both entries zeroed (Table~\ref{tab:coding_corners}); the protocol fixed no single-entry prediction for this cell.

\begin{table*}[t]
\centering\footnotesize
\setlength{\tabcolsep}{3.5pt}
\renewcommand{\arraystretch}{1.04}
\caption{\textbf{Qwen: main only and scale-normalized coding against shared QP.} Score in percent (ANLS$\times$100 for DocVQA), KB per image or document, differences in points. Main-only rows keep main at the stated QP and zero the later entries. They are compared with the shared configuration at the same main QP, and the byte change is relative to that configuration. Every row is also compared with the shared-QP curve at the same KB, read by linear interpolation in log KB between the two neighboring shared points and recomputed in each bootstrap resample. Of the seven main-only differences, five intervals contain zero, one lies wholly above zero (Qwen32 attribute, $12\,|\,\mathrm{Z}$) and one wholly below (Qwen32 DocVQA, $10\,|\,\mathrm{Z}$). Of the 13 scale-normalized differences, no interval lies wholly above zero and eight lie wholly below. Where the curve bends, linear interpolation can fall below it and overstate a positive difference: $12\,|\,\mathrm{Z}$ lies between $14\,|\,14$ ($5.33$ KB, $57.10$) and $12\,|\,12$ ($18.90$ KB, $69.25$). Five scale-normalized configurations lie outside the byte range of the shared points and are omitted. The shared-QP points are plotted in Figure~\ref{fig:coding_rate} and listed in \texttt{coding\_\allowbreak{}configurations.csv}. Two further Qwen8 comparisons: $0\,|\,\mathrm{Z}$ minus $2\,|\,2$ is $0.55$ $[-0.70, 1.85]$ points with $17.98$\% more bytes (F), and the configuration selected on calibration images at the high budget, $2\,|\,\mathrm{n}{+}4$ ($45.64$ KB, $70.10$), minus $2\,|\,2$ is $-0.25$ $[-1.20, 0.65]$ points with $12.86$\% fewer bytes (L). Uncoded full input: Qwen8 $72.50$, Qwen32 $72.75$, Qwen32 DocVQA $78.64$. Notation as in Table~\ref{tab:coding_deepstack}. Status codes and intervals as in Table~\ref{tab:coding_differences}.}
\label{tab:coding_qwen}
\DSalign{table_coding_qwen-1}
\begin{tabular}{@{}llrccrrc@{}}
\toprule
\rowcolor{DSTableHeader}
\multicolumn{4}{@{}l}{} & \multicolumn{2}{c}{Shared QP, same main QP (F)} & \multicolumn{2}{c@{}}{Shared-QP curve, same KB (P)} \\ \cmidrule(lr){5-6}\cmidrule(l){7-8}
\rowcolor{DSTableHeader}
Configuration & Family & KB & Score & Difference & Byte change (\%) & Curve & Difference \\
\midrule
\rowcolor{DSTableAccent}\multicolumn{8}{@{}l}{\textbf{Qwen8, GQA attribute (1,000 images)}}\\
$0\,|\,\mathrm{Z}$ & main only & \DSv{61.79}{} & \DSv{70.90}{[68.75, 72.95]} & \DSv{-1.55}{[-2.80, -0.25]} & \DSv{-55.34}{} & \DSv{70.71}{} & \DSv{0.19}{[-0.94, 1.41]} \\
$2\,|\,\mathrm{Z}$ & main only & \DSv{33.78}{} & \DSv{68.80}{[66.65, 70.90]} & \DSv{-1.55}{[-2.85, -0.25]} & \DSv{-35.51}{} & \DSv{68.01}{} & \DSv{0.79}{[-0.38, 1.94]} \\
$6\,|\,\mathrm{Z}$ & main only & \DSv{6.38}{} & \DSv{54.75}{[52.45, 57.10]} & \DSv{-0.50}{[-1.35, 0.35]} & \DSv{-9.07}{} & \DSv{54.73}{} & \DSv{0.02}{[-0.80, 0.87]} \\
$0\,|\,\mathrm{n}$ & scale-normalized & \DSv{304.32}{} & \DSv{72.30}{[70.15, 74.35]} &  &  & \DSv{72.84}{} & \DSv{-0.54}{[-1.46, 0.37]} \\
$2\,|\,\mathrm{n}$ & scale-normalized & \DSv{156.79}{} & \DSv{71.85}{[69.70, 73.90]} &  &  & \DSv{72.51}{} & \DSv{-0.66}{[-1.78, 0.47]} \\
$4\,|\,\mathrm{n}$ & scale-normalized & \DSv{66.76}{} & \DSv{69.65}{[67.50, 71.80]} &  &  & \DSv{70.87}{} & \DSv{-1.22}{[-2.53, 0.08]} \\
$6\,|\,\mathrm{n}$ & scale-normalized & \DSv{18.25}{} & \DSv{57.40}{[55.15, 59.75]} &  &  & \DSv{63.71}{} & \DSv{-6.31}{[-7.91, -4.68]} \\
$8\,|\,\mathrm{n}$ & scale-normalized & \DSv{4.81}{} & \DSv{49.60}{[47.30, 52.00]} &  &  & \DSv{53.17}{} & \DSv{-3.57}{[-4.88, -2.29]} \\
\rowcolor{DSTableAccent}\multicolumn{8}{@{}l}{\textbf{Qwen32, GQA attribute (1,000 images)}}\\
$6\,|\,\mathrm{Z}$ & main only & \DSv{105.41}{} & \DSv{71.75}{[69.55, 73.85]} & \DSv{-1.20}{[-2.55, 0.15]} & \DSv{-47.09}{} & \DSv{72.30}{} & \DSv{-0.55}{[-1.76, 0.71]} \\
$12\,|\,\mathrm{Z}$ & main only & \DSv{14.74}{} & \DSv{68.60}{[66.40, 70.80]} & \DSv{-0.65}{[-1.70, 0.45]} & \DSv{-22.01}{} & \DSv{66.86}{} & \DSv{1.74}{[0.69, 2.79]} \\
$10\,|\,\mathrm{n}$ & scale-normalized & \DSv{143.61}{} & \DSv{72.70}{[70.55, 74.85]} &  &  & \DSv{72.62}{} & \DSv{0.08}{[-1.07, 1.24]} \\
$12\,|\,\mathrm{n}$ & scale-normalized & \DSv{49.90}{} & \DSv{71.20}{[69.00, 73.35]} &  &  & \DSv{71.53}{} & \DSv{-0.33}{[-1.60, 0.97]} \\
$14\,|\,\mathrm{n}$ & scale-normalized & \DSv{12.97}{} & \DSv{61.15}{[58.80, 63.55]} &  &  & \DSv{65.64}{} & \DSv{-4.49}{[-6.08, -2.88]} \\
$16\,|\,\mathrm{n}$ & scale-normalized & \DSv{4.45}{} & \DSv{47.25}{[44.90, 49.60]} &  &  & \DSv{54.67}{} & \DSv{-7.42}{[-9.20, -5.68]} \\
\rowcolor{DSTableAccent}\multicolumn{8}{@{}l}{\textbf{Qwen32, DocVQA (474 documents, ANLS$\times$100)}}\\
$6\,|\,\mathrm{Z}$ & main only & \DSv{140.00}{} & \DSv{74.86}{[72.23, 77.50]} & \DSv{-1.61}{[-3.08, -0.17]} & \DSv{-19.19}{} & \DSv{74.75}{} & \DSv{0.11}{[-1.26, 1.48]} \\
$10\,|\,\mathrm{Z}$ & main only & \DSv{49.79}{} & \DSv{52.46}{[49.36, 55.54]} & \DSv{-3.20}{[-4.67, -1.82]} & \DSv{-4.74}{} & \DSv{53.99}{} & \DSv{-1.54}{[-2.95, -0.20]} \\
$10\,|\,\mathrm{n}$ & scale-normalized & \DSv{115.75}{} & \DSv{60.10}{[57.15, 63.01]} &  &  & \DSv{73.21}{} & \DSv{-13.12}{[-15.63, -10.64]} \\
$12\,|\,\mathrm{n}$ & scale-normalized & \DSv{44.97}{} & \DSv{38.33}{[35.45, 41.26]} &  &  & \DSv{50.50}{} & \DSv{-12.18}{[-14.32, -9.96]} \\
$14\,|\,\mathrm{n}$ & scale-normalized & \DSv{12.55}{} & \DSv{14.46}{[12.50, 16.47]} &  &  & \DSv{19.70}{} & \DSv{-5.24}{[-6.54, -3.93]} \\
$16\,|\,\mathrm{n}$ & scale-normalized & \DSv{4.74}{} & \DSv{8.52}{[6.98, 10.08]} &  &  & \DSv{10.10}{} & \DSv{-1.57}{[-2.35, -0.79]} \\
\bottomrule
\end{tabular}
\end{table*}

\subsection{DeepStack: bytes and accuracy}
\label{app:coding_deepstack}
Coding only DeepStack's main entry takes about half the bytes of coding all entries at a shared QP, at similar accuracy. Accuracy under a shared QP falls monotonically with bytes down to 71.43~KB (QP~2, 16.40\%) and is higher again at 42.12~KB (QP~4, 27.05\%; Table~\ref{tab:coding_deepstack}), so the curve is not monotone below about 75~KB. We therefore compare DeepStack configurations directly in pairs, and the main paper reports two such readouts (Table~\ref{tab:coding_deepstack_pairs}). First, main alone at QP~$-4$ with later zeroed takes 255.08~KB for 77.45\%, against 503.26~KB for 76.35\% at shared QP~$-2$: 49.31\% fewer bytes and a difference of $1.10$ [$-0.40,2.55$] points. An exploratory run on the 240 calibration units had shown about 494~KB for shared QP~$-2$ against 254~KB for main alone at QP~$-4$, both within three points of the full input. We fixed this pair after the first confirmation results, which already included uncoded main alone ($77.85\%$), main's bytes at QP~$-4$ and the score of the shared configuration, and before $-4\,|\,\mathrm{Z}$ itself was run.

Second, the protocol rule selected $-2\,|\,\mathrm{n}{+}2$ at the high budget, which takes 278.19~KB for 75.80\%: 39.80\% fewer bytes than the scale-normalized $-2\,|\,\mathrm{n}$ at the same main QP ($-0.45$ [$-1.40,0.50$]) and 44.72\% fewer than $-2\,|\,{-2}$ (Table~\ref{tab:coding_deepstack_pairs}). The protocol had classed this cell as one without a predicted role direction, so these comparisons carry no predicted sign. The protocol also set an accuracy target of 95\% of the gap between both entries zeroed and the full input, which is 76.10\% on these images. The shared and scale-normalized configurations at main QP~$-2$ reach it (76.35\% and 76.25\%), and the selected $-2\,|\,\mathrm{n}{+}2$ falls just below it (75.80\%).

At an equal main QP, the later entries still contribute: zeroing them lowers accuracy by 1.45 to 3.60 points at QP~$-4$, $-2$ and $-1$ and removes about 74\% of the bytes each time, with two of the three intervals excluding zero (Table~\ref{tab:coding_deepstack_pairs}). Post hoc, main alone at QP~$-4$ scores $1.20$ [$-0.25,2.65$] points above $-2\,|\,\mathrm{n}$ with 44.80\% fewer bytes.

The cost of removing an entry depends on the task. On the new attribute images, keeping main alone changes accuracy by $-1.65$ points and keeping later alone by $-3.45$, both with intervals excluding zero (Table~\ref{tab:coding_differences}); both differences are post hoc, because the protocol fixed these configurations but not their differences from the full input. Exploratory runs on an Apple-silicon GPU, with zeros and on the 192 evaluation units of Appendix~\ref{app:v15}, gave $-3.91$ [$-7.03,-1.04$] for main alone and $0.26$ [$-3.91,4.43$] for later alone on GQA spatial questions. On DocVQA, where the full input scores only 34.56 ANLS points, they gave $-9.48$ [$-13.42,-5.56$] for main alone and $-6.03$ [$-9.99,-2.28$] for later alone. These four differences are exploratory, not confirmation results, and their intervals are paired bootstrap intervals over the 192 units, computed for those runs.

\begin{table*}[t]
\centering\footnotesize
\setlength{\tabcolsep}{3.2pt}
\renewcommand{\arraystretch}{1.04}
\caption{\textbf{Role-direction predictions fixed in the protocol: they hold for Qwen and fail for Granite.} For a model whose main entry carried most of the accuracy (Qwen8 attribute, Qwen32 DocVQA) the role allocation codes the later entries coarser than scale-normalized (n+$k$); for Granite, whose later entries carried most of it, it codes them finer (n$-k$). The mirror reverses the sign of $k$. At each budget each allocation takes its best configuration on calibration images within the byte cap, so the two compared configurations need not have equal bytes. The high and low budgets are the cheapest shared-QP configurations reaching 95\% and 60\% of the gap between both entries zeroed and the full input on calibration images. The prediction was role minus mirror above zero. Granite has only its first 468 images, so the protocol's analysis was not run for it; its differences are computed with the same bootstrap as the other rows. Score in percent (ANLS$\times$100 for DocVQA), KB per image or document, difference in points. All rows are L. Notation as in Table~\ref{tab:coding_deepstack}. Status codes and intervals as in Table~\ref{tab:coding_differences}.}
\label{tab:coding_role}
\DSalign{table_coding_role-1}
\begin{tabular}{@{}llcrrcrrcl@{}}
\toprule
\rowcolor{DSTableHeader}
Model, task & Budget & Role & KB & Score & Mirror & KB & Score & Difference & Outcome \\
\midrule
Qwen8, attribute & high & $2\,|\,\mathrm{n}{+}4$ & \DSv{45.64}{} & \DSv{70.10}{} & $10\,|\,\mathrm{n}{-}4$ & \DSv{12.44}{} & \DSv{44.20}{} & \DSv{25.90}{[23.25, 28.50]} & holds \\
Qwen8, attribute & low & $6\,|\,\mathrm{n}{+}4$ & \DSv{6.77}{} & \DSv{54.75}{} & $10\,|\,\mathrm{n}{-}2$ & \DSv{2.98}{} & \DSv{43.95}{} & \DSv{10.80}{[8.95, 12.85]} & holds \\
Qwen32, DocVQA & high & $8\,|\,\mathrm{n}{+}4$ & \DSv{117.49}{} & \DSv{72.15}{} & $12\,|\,\mathrm{n}{-}2$ & \DSv{91.80}{} & \DSv{41.24}{} & \DSv{30.91}{[27.86, 34.02]} & holds \\
Qwen32, DocVQA & low & $12\,|\,\mathrm{n}{+}2$ & \DSv{30.39}{} & \DSv{34.92}{} & $16\,|\,\mathrm{n}{-}4$ & \DSv{22.26}{} & \DSv{9.00}{} & \DSv{25.91}{[23.12, 28.73]} & holds \\
Granite, attribute (468) & high & $8\,|\,\mathrm{n}{-}4$ & \DSv{212.91}{} & \DSv{40.92}{} & $0\,|\,\mathrm{n}{+}2$ & \DSv{614.73}{} & \DSv{70.19}{} & \DSv{-29.27}{[-33.33, -25.11]} & fails \\
Granite, attribute (468) & low & $12\,|\,\mathrm{n}{-}4$ & \DSv{36.51}{} & \DSv{9.62}{} & $8\,|\,\mathrm{n}{+}2$ & \DSv{19.72}{} & \DSv{21.05}{} & \DSv{-11.43}{[-14.53, -8.33]} & fails \\
\bottomrule
\end{tabular}
\end{table*}

\subsection{Granite: a fixed pair on held-out images}
\label{app:coding_granite}
A fixed Granite configuration scores above a shared QP on held-out images at nearly equal bytes. On the 531 held-out images, $0\,|\,\mathrm{n}{+}4$ scores 69.30\% and the shared $-4\,|\,{-4}$ 66.67\%, a difference of $2.64$ [$0.47,4.80$] points at 0.70\% more bytes (395.63 against 392.89~KB; Table~\ref{tab:coding_granite}). Relative to the shared QP, the configuration moves bytes from main to later.

The pair was found post hoc: $0\,|\,\mathrm{n}{+}4$ is one of the 31 protocol configurations, but the selection rule did not pick it. We first read its advantage over $-4\,|\,{-4}$ on Granite's first 468 images ($3.42$ [$1.28,5.45$]), among several comparisons with shared-QP configurations, and then fixed this single pair and scored it on the 531 images that Granite had not seen.

We credit the gain to this configuration and not to the role diagnosis that Granite is later-limited. On the 468 discovery images, the scale-normalized $4\,|\,\mathrm{n}$, which codes the later entries exactly as $0\,|\,\mathrm{n}{+}4$ does and main four QP units coarser, already scores $2.46$ [$0.11,4.81$] points above $-4\,|\,{-4}$ with 31.12\% fewer bytes, and $0\,|\,\mathrm{n}{+}4$ minus $4\,|\,\mathrm{n}$ is $0.96$ [$-0.96,2.88$]. Both comparisons are post hoc. An exploratory run on the 192 evaluation images of Appendix~\ref{app:v15} that coded only the later entries, all at QP~$-8$, with main replaced by its mean, did not differ from $-4\,|\,{-4}$ at matched bytes (395.51 against 395.24~KB, $0.00$ [$-4.69,4.43$]).

\subsection{Qwen: no consistent advantage over a shared QP}
\label{app:coding_qwen}
Qwen8/32 show no consistent advantage over a shared QP: at the same main QP, zeroing their later entries lowers the point estimate at every tested QP, by 0.50 to 3.20 points, and saves 4.74\% to 55.34\% of the bytes (Table~\ref{tab:coding_qwen}). Against the shared-QP curve at the same bytes (post hoc; interpolated between neighboring shared points), one of the seven main-only differences has an interval wholly above zero (Qwen32 attribute $12\,|\,\mathrm{Z}$, at a bend where linear interpolation can fall below the curve) and one wholly below (Qwen32 DocVQA $10\,|\,\mathrm{Z}$). Of the 13 scale-normalized differences, no interval lies wholly above zero and eight lie wholly below (Table~\ref{tab:coding_qwen}).

The configurations that the rule selected on calibration units give the same picture. For Qwen8 at the high budget it selected $2\,|\,\mathrm{n}{+}4$, with 12.86\% fewer bytes than $2\,|\,2$ and a difference of $-0.25$ [$-1.20,0.65$]. For Qwen32 attribute questions it selected $8\,|\,\mathrm{n}{+}4$ (105.75~KB, 72.50\%), which used 46.91\% fewer bytes than the shared configuration selected at the same budget, $6\,|\,6$ (199.21~KB, 72.95\%). Most of that byte difference (88.64\%) lies between the shared points $6\,|\,6$ and $8\,|\,8$: at the fixed main QP~8, $8\,|\,\mathrm{n}{+}4$ saved 9.13\% of the bytes of $8\,|\,8$ with a difference of $0.10$ [$-0.80,1.00$] (post hoc).

\subsection{Role-direction comparisons and training pilot}
\label{app:coding_role}
The role-direction predictions hold for the Qwen8/32 cells that were run and fail for Granite, and fine-tuning on coded features shows no role-specific benefit.

\textbf{Role allocation.} The protocol classified each cell from the single-entry losses of the calibrated experiment (Appendix~\ref{app:v15}) as main-limited (main carries most of the accuracy: Qwen8 on all three tasks, Qwen32 on DocVQA), later-limited (later carries it: Granite attribute, DeepStack DocVQA), without a predicted direction (Qwen32 and DeepStack attribute, DeepStack spatial), or descriptive only. The role allocation codes later coarser than n in a main-limited cell and finer in a later-limited cell, and its mirror reverses the sign of that shift. The prediction was that the role allocation beats its mirror (Table~\ref{tab:coding_role}). It holds at both budgets for Qwen8 attribute and Qwen32 DocVQA (differences of 10.80 to 30.91 points, all four intervals above zero) and fails for Granite on the 468 complete images (differences of $-29.27$ and $-11.43$ points, both intervals below zero). The other cells with a role prediction were not run, and the two arms of a pair are not matched in bytes: at Granite's high budget the mirror used 614.73~KB and the role allocation 212.91~KB. Beating the mirror also does not imply beating a shared QP: Qwen8's high-budget role configuration is the $2\,|\,\mathrm{n}{+}4$ whose difference from $2\,|\,2$ has an interval containing zero (Appendix~\ref{app:coding_qwen}).

\textbf{Training pilot.} A pilot asked whether fine-tuning on coded features helps and whether role-specific offsets help more than generic ones. For Qwen8 and Granite, we trained rank-16 LoRA adapters (low-rank weight updates) on the language model, with the vision side frozen, for one pass over 8,000 GQA training attribute questions on 4,352 images (500 optimizer steps of 16 samples), disjoint from all evaluation units. Plain fine-tuning sees the full input, whereas three arms see coded features with the later entries offset by $|k|\in\{2,4\}$ QP units from n, with a random sign (generic), the role sign or the opposite sign (anti-role). For each sample the three coded arms share the image, $|k|$ and a byte cap, the bytes of that image at $0\,|\,\mathrm{n}$, $2\,|\,\mathrm{n}$, $4\,|\,\mathrm{n}$ or $6\,|\,\mathrm{n}$, and each arm takes the finest main QP that fits the cap, so actual bytes differ slightly (role-to-anti-role byte ratio 0.97 for Qwen8, 1.02 for Granite). All arms were evaluated on the 192 evaluation images of Appendix~\ref{app:v15} (349 questions) at fixed configurations that give every arm the same bytes. The pilot was not evaluated on the confirmation images, so it is exploratory.

At $8\,|\,\mathrm{n}$ (4.89~KB), a coarser level than any training budget, Qwen8's generic arm scored 67.97\% against 51.30\% after plain fine-tuning ($16.67$ [$10.94,22.14$], post hoc). The role-specific arm did not differ from the generic arm ($-0.78$ [$-4.95,3.13$]) and scored below the anti-role arm ($-3.91$ [$-7.55,-0.26$]). For Granite, all four fine-tuned arms scored 75.78\% at $4\,|\,\mathrm{n}$. All pilot scores and comparisons are in \texttt{coding\_\allowbreak{}configurations.csv} and \texttt{coding\_\allowbreak{}comparisons.csv}.

\subsection{Complete estimates}
\label{app:coding_data}
\texttt{coding\_\allowbreak{}single\_\allowbreak{}entry.csv} holds every single-entry score, paired difference and prediction margin on each image set. \texttt{coding\_\allowbreak{}configurations.csv} holds every configuration on every image set with its bytes per entry and score, including Granite's 31 configurations on the 468 images, the shared-QP points of Figure~\ref{fig:coding_rate} and both training checkpoints (150 and 500 steps). \texttt{coding\_\allowbreak{}comparisons.csv} holds every paired comparison with both byte sizes. Values are unrounded and carry intervals and status codes L, F and P (Appendix~\ref{app:data}). The exploratory Apple-silicon results quoted here are not in these files.

\section{Additional experiments}
\label{app:additional}
This appendix reports results that the main text cites only briefly or not at all. Sections~\ref{app:blink}--\ref{app:dense_dose} describe exploratory experiments, which are not pooled with the main-text panels, and further readouts of the spatial and whole-source panels; Sections~\ref{app:scores_answers}--\ref{app:recovery_more} mainly give further readouts and cells of the main experiments.

\subsection{Removing later entries on a benchmark}
\label{app:blink}
Removing Qwen3-VL-4B's later entries barely changes BLINK accuracy but changes many individual predictions. We replaced each of its three later entries by zeros at its own injection site, with main unchanged, on all 1,901 labeled BLINK validation items \citep{fu2024blink}, and scored every complete option label under the eight on/off combinations (BF16 on an Apple-silicon GPU). Removing all three lowered accuracy only from 49.18\% to 47.29\% and the mean correct-option margin (a log-probability gap between options) by 0.14 nats (SE 0.03), yet it changed 251 predictions: 115 from correct to wrong, 79 from wrong to correct and 57 between wrong options. Removing one entry changed 140 to 147 predictions, with mean margin changes between $-0.02$ and $-0.004$, and 442 items changed under at least one combination. Item responses were entry-specific: on the 1,789 items other than the 112 analyzed first, repeated removals of the same entry correlated at 0.41 across items against 0.10 for different entries, and keeping any one entry recovered 50\% to 73\% of the margin lost by removing all three. Changed predictions concentrated in near-tie items (options nearly tied), which made up 17\% of these 1,789 items but gave 68\% to 75\% of single-entry changes.

Item effects also depended on the answer format. On 204 items with an explicit letter format, removal lowered the number of correct items from 119 to 110 (Qwen3-VL-4B) and from 117 to 105 (Qwen3-VL-8B) with options in their original order, but only from 96 to 94 and from 91 to 88 with cyclically permuted options, and the item-level changes correlated at 0.05 and $<0.001$ between the two orders. Removal therefore gives small net accuracy changes, many offsetting item changes and format-dependent item effects. By contrast, the conflict and recovery experiments vary entry inputs and read effects through target and control queries, and Appendix~\ref{app:coding} removes entries in Qwen8, Qwen32, Granite and DeepStack on new GQA images and DocVQA documents, with a calibration mean as well as zeros as the replacement. A post hoc analysis of 135 Visual Similarity items also found that single later entries act conditionally, that is, their effect depends on main's input: in Qwen3-VL-8B, exchanging the main input raised the third entry's effect by 0.27 nats (SE 0.08).

\subsection{Spatial correspondence across designs and models}
\label{app:spatial_more}
Three further designs test whether entry effects follow the object currently at a position, and all are consistent with it. First, a factorial design crossed the receiving scene's layout ($M0/M1$: original or exchanged), the donor's layout and the placement of later-entry tokens. One interaction in it separates effects that follow the object currently at a position (current scene correspondence) from effects tied to the original coordinates, and it was above zero in 12 of 12 attribute, background and readout strata in Qwen3-VL-4B and 8B. Real Visual Genome crops recomposed into two-object canvases gave the same pattern in Qwen3-VL-4B and in Qwen3-VL-8B's unit means. We repeated the factorial with Granite (all later entries and its two entry groups), Qwen32, Qwen30 (Qwen3-VL-30B-A3B) and both Cosmos3 checkpoints. A least-squares decomposition onto three templates (current scene layout, original coordinates, donor identity) gave a scene-layout component above zero in all 84 pooled conditions, whereas the original-coordinate component included zero in 205 of 210 label-order-specific rows, and donor identity also contributed in some conditions.

Second, three-object scenes placed objects in three columns or in three cells of a $2\times2$ grid, with an exact cyclic layout exchange, and all 24 predefined statistics were positive with intervals above zero for Granite, Qwen8 and Qwen32. With direct answers at $\alpha=.08$, the layout interaction in three columns was $5.64$ [$5.11,6.18$], $3.50$ [$3.07,3.93$] and $12.5$ [$11.2,13.9$], and real crops recomposed into the same layouts gave $3.52$ [$3.10,3.94$], $1.47$ [$1.14,1.80$] and $8.80$ [$7.52,10.1$]. These crops change color, instance and context together. Third, a held-out model of the local panels predicted each signed effect from whether the destination holds the edited object. Relative to an identity baseline, it reduced the cross-validated error of the relative selectivity $R$ (target minus control effect) from 2.08 to 0.72 (Granite), 11.48 to 4.75 (Qwen8) and 20.05 to 5.44 (Qwen32), close to the error of training-condition means. This comparison was specified after the effects had been seen.

The prediction analysis also compares current object correspondence with a fixed-destination explanation. Both explanations have the same number of parameters, with separate gains for each entry, and fit one scene layout identically, but they predict differently once the objects exchange positions. In each of four folds, we fit one layout on 48 scenes and predict the other layout on 16 different scenes, keeping all queries and conditions of a scene in the same fold. Holding out $M0$, the mean absolute error of direct-answer score selectivity $R$ falls from $4.15$ to $0.77$ in Granite, $22.63$ to $4.85$ in Qwen8 and $40.05$ to $5.29$ in Qwen32. Holding out $M1$ gives the same ordering, as do the later-only comparisons at both background strengths. These errors compare explanations only within each model's score scale. Because layout exchange changes main and the unselected later background together while keeping the local donor and its destination fixed, the comparison supports current scene correspondence, not an independent effect of main on binding (attribute-to-object assignment). This exploratory analysis tests the controlled left--right layout exchange, and its competing models were specified after inspecting the original effects.

For the local transfers of the main paper's Figure~\ref{fig:spatial}, the score interactions (spatial interactions $I$; Appendix~\ref{app:spec}) of Granite, Qwen8 and Qwen32 were $12.12$ [$11.77,12.48$], $6.35$ [$5.76,6.94$] and $31.99$ [$29.95,34.04$], and the generation interactions with direct answers at $\alpha=.08$ were 1.56 (SE 0.11, 32 of 32 scenes) for Granite, 1.80 (SE 0.13, 32 of 32) for Qwen32 and 0.39 (SE 0.09, 26 of 32) for Qwen8. Among the added checkpoints, Qwen30's main-entry color selectivity changed from $7.08$ [$5.88,8.28$] at the original destination to $-4.74$ [$-5.66,-3.83$] at the opposite one, with identity labels at full background strength. DeepStack's later-entry color generation selectivity changed from $0.79$ [$0.72,0.85$] to $-0.37$ [$-0.46,-0.27$], and it reversed again under the exchanged layout ($-0.40$ [$-0.49,-0.30$] and $0.75$ [$0.68,0.82$]).

\subsection{Whole-image source competition}
\label{app:whole_image}
In whole-image source competition, the role contrast favors later for Granite and main for Qwen8 and Qwen32 on identity questions about the edited object. The source-competition panel re-rendered the 64 two-object scenes with a gray cross between the objects and gave the main entry and all later entries independent sources, layouts and strengths. It asked identity questions (``What color is the circle?'') and relation questions about the side of the cross. For identity questions about the edited object, the role contrast $D=E_{\main}-E_{\later}$ was negative in all 64 conditions for Granite (60 intervals below zero) and positive in all 64 for Qwen8 and Qwen32 (all intervals above zero); in the reference condition, its score values were $-0.88$ [$-0.98,-0.79$], $38.40$ [$36.54,40.26$] and $45.00$ [$44.62,45.38$]. Qwen's later effect remained measurable at full strength, with intervals [$0.14,0.38$] (Qwen8) and [$0.37,0.60$] (Qwen32). The source four-cell interaction $J=y_{bb}-y_{ba}-y_{ab}+y_{aa}$ (the nonadditivity of the four main and later source assignments) of the score was $1.19$ [$0.99,1.39$], $0.15$ [$-0.0002,0.30$] and $1.22$ [$0.94,1.49$]. In generation it was $0.28$ [$0.13,0.43$] for Granite and 0 for both Qwen models, whose answers changed with the main source and never with the later source, so a zero $J$ does not mean that neither entry is preferred. Whole-source conflicts on the controlled scenes complement the Qwen30 and DeepStack whole-source results of Appendix~\ref{app:figures}, whose quantities are defined in Appendix~\ref{app:source_tables}. With direct answers, the generation entry contrast $\Delta R$ (main minus later relative selectivity) was $-0.41$ [$-0.48,-0.33$] for Granite, $1.00$ for Qwen8 and $0.98$ [$0.95,1.00$] for Qwen32 with color, and $-0.09$ [$-0.17,-0.02$], $0.71$ [$0.61,0.81$] and $0.45$ [$0.36,0.55$] with stripes.

\subsection{Dose curves on a dense subset}
\label{app:dense_dose}
Entry effects are not monotone in dose. On an outcome-independent subset of 16 whole-image scenes, raw multipliers $k\in\{0,.25,.5,1,2,4\}$ (the same multiplier of the replacement for every entry, unlike magnitude-matched doses) traced dose curves in 12 backgrounds. Qwen8's main score effect was 39.17 at $k=1$ and 37.73 at $k=4$, Qwen32's later effect rose from 0.32 at $k=1$ to 40.77 at $k=4$, and Granite's main and later effects changed order between $k=2$ (6.69 against 8.28) and $k=4$ (7.77 against 7.20). Across backgrounds, the Qwen main score fell from $k=1$ to $k=4$ in 8 of 12, whereas the main generation mean did not change in 9 of 12 (Qwen8) and 11 of 12 (Qwen32), although individual answers still changed (5 of 377 and 8 of 380). The delivered magnitudes stayed within 0.99 and 1.01 of their canonical values (Appendix~\ref{app:dose}). Matched magnitudes therefore compare entries at one dose rather than along a linear dose scale.

\subsection{Later entries move scores without changing answers}
\label{app:scores_answers}
On the photographs, later interventions in Qwen8/32 change no target answer in the reference condition, although their later score selectivity is positive: $0.69$ [$0.52,0.85$] and $1.30$ [$1.06,1.55$]. A 32-photograph later-only panel measured shifts of the candidate margin (the score gap between the two candidate answers). For Qwen32's direct answers about the edited object in the reference background, the median absolute shift was 0.76, against a median starting margin of 19.50. Over all pairs, the median shift was 1.04\% (Qwen8) and 2.35\% (Qwen32) of the starting margin, against 77.85\% for Granite, whose answers changed in 126 of 456 pairs. Rescaled to main's native magnitude (Appendix~\ref{app:dose}), Qwen's later generation effects remain small (0.01 [$-0.01,0.02$] and 0.06 [$0.02,0.09$] under RSS, root-sum-square matching), whereas main rescaled to later's magnitude keeps selectivity of 0.15 and 0.25, so magnitude accounts for part of the raw contrast but not for its direction. Qwen's later features thus move its evidence without crossing the answer threshold, and a change of main quality can reveal their influence (main paper, Section~\ref{sec:conditional}).

\subsection{Conditional influence in other cells}
\label{app:conditional_more}
The accuracy-invisible change of the main paper's Figure~\ref{fig:conditional_quality}a (degrading main strengthens unchanged later features' effect while accuracy stays fixed) recurs in other cells of the calibrated experiment. Here $I_{\mathrm{quality}}$ is the change in later selectivity when main is degraded rather than clean. For Qwen32 stripes at the mildest blur, $I_{\mathrm{quality}}$ is $0.61$ [$0.50,0.72$] in generation and 18.64 in score under identity labels, and $0.47$ [$0.34,0.60$] and 15.97 under cyclic labels. For relative position at the same blur, all four Qwen32 accuracies are again 1 while later generation selectivity rises from 0.13 to 0.42 ($I_{\mathrm{quality}}=0.30$ [$0.16,0.43$]), and its score $I_{\mathrm{quality}}$ grows with blur (4.66, 9.46 and 9.80). For color at the mildest desaturation, Qwen32's change appears only in scores ($3.61$ [$2.82,4.40$]). The dependence is model- and task-specific: Qwen8's stripe score $I_{\mathrm{quality}}$ falls to $-1.68$ [$-2.40,-0.97$] at the strongest blur, and Granite's relative-position score $I_{\mathrm{quality}}$ is negative at all three levels ($-0.25$, $-0.62$ and $-0.75$).

In the cell of Figure~\ref{fig:conditional_quality}a, the target's starting margin barely changes ($-19.57$ to $-18.52$) while its target effect rises from 8.82 to 27.63, so the change is not a starting-margin artifact. For Qwen32 color, by contrast, the starting margin shrinks from $-23.73$ to $-10.03$ as the target effect grows from 0.47 to 8.91, so margin dependence must be judged per cell. On shared units, conflict and recovery can move together: at the strongest desaturation, Qwen32's later color selectivity rises from 0.00 to 0.42 ($I_{\mathrm{quality}}=0.42$ [$0.28,0.57$]) while clean later features restore $0.25$ [$0.15,0.35$] of the color loss; for Qwen32 text, both stay near zero.

\subsection{Recovery in other cells}
\label{app:recovery_more}
Further cells extend the recovery patterns of the main paper's Section~\ref{sec:recovery}. Recovery ($G_D$) is the accuracy that clean later entries regain when main is degraded, and $I_{\mathrm{rec}}$ is the nonadditivity of the two entries' losses. On relative position at the strongest blur, Qwen32 recovered $0.19$ [$0.10,0.28$] of a 0.20 loss ($I_{\mathrm{rec}}=0.11$ [$0.03,0.19$]) and Qwen8 $0.17$ [$0.05,0.29$] of a 0.25 loss, while Qwen8 recovered color only partly and non-monotonically ($G_D$ of 0.05, 0.00 and 0.13 at losses of 0.05, 0.13 and 0.28). On GQA, Qwen accuracy is not purely main-limited (main carries most of the accuracy) as it is on DocVQA: Qwen32's attribute recovery rose with loss (0.06, 0.15 and 0.22 at losses of 0.12, 0.22 and 0.30; $I_{\mathrm{rec}}=0.19$ [$0.12,0.26$] at the strongest level), and Qwen8's spatial interactions excluded zero at every level.

The model-specific levels, chosen so that each model keeps about half of its ability above chance (Appendix~\ref{app:v15}), give another pair of similar losses with different recovery: on GQA spatial questions, Qwen8 lost $0.23$ [$0.17,0.30$] at a 7-pixel long side and recovered $0.01$ [$-0.02,0.04$], whereas Qwen32 lost $0.26$ [$0.20,0.33$] at 5 pixels and recovered $0.19$ [$0.13,0.24$]. The levels differ between the two models, so these rows form a separate estimand and are not pooled with the shared levels. The equal-loss comparison of Qwen32 and Qwen8 on stripes (main paper, Section~\ref{sec:recovery}) holds for direct answers only: under identity labels the recovery difference is $0.50$ [$0.36,0.64$], but Qwen8's loss is 0.39 (paired loss difference $0.11$ [$-0.04,0.26$]). In the fixed-contrast experiment, Qwen's direct and identity color cells saturate, whereas cyclic labels expose recovery of $0.11$ [$0.03,0.19$] (Qwen8) and $0.30$ [$0.21,0.39$] (Qwen32); DeepStack's cyclic $I_{\mathrm{rec}}$ of $0.25$ [$0.11,0.39$] arises because clean later features lower its accuracy when main is clean (0.83, against 0.97 with degraded later).

\section{Supplementary related work}
\label{app:related}
\textbf{Visual interfaces and evaluation controls.}
LLaVA projects visual-grid features into language embeddings \citep{liu2023llava}, while BLIP-2 and Honeybee study learned-query interfaces and locality-preserving compression \citep{li2023blip2,cha2024honeybee}. Flamingo offers repeated cross-attention access, and Cambrian-1 and DeepStack supply spatial visual features at multiple language-model depths \citep{alayrac2022flamingo,tong2024cambrian,meng2024deepstack}. Qwen3-VL instead projects intermediate encoder features into early decoder layers \citep{qwen2025qwen3vl}, and Cosmos3 retains this interface after continued training \citep{nvidia2026cosmos3}. Granite Vision describes its design as ``a variant of Deepstack'' in which visual features are added to the language-model hidden states at multiple layers \citep{ibm2026granitevision}. We preserve these native differences when comparing entries. Whereas these designs specify how visual features reach the language model, our entry-wise interventions give the main entry and the later entries different versions of one image (under conflict, degraded main input, and recovery) or remove or code each entry.

CLEVR, Visual Genome, and GQA support controlled or annotated object questions \citep{johnson2017clevr,krishna2017visualgenome,hudson2019gqa}. Compositional tests expose failures to preserve object--attribute assignments, while spatial benchmarks vary relations or object positions \citep{thrush2022winoground,yuksekgonul2023aro,liu2023vsr,kamath2023whatsup}. Unlike these benchmarks, which vary the scene, relation, or question, we vary which entry carries which version of one image, and our paired queries use the same visual intervention to separate the edited object's response from an unchanged object's response.

\textbf{Intervention interpretation.}
Activation patching and interchange interventions test causal output effects or correspondence with a proposed computation \citep{vig2020causal,meng2022locating,geiger2021causal}, and their interpretation depends on corruption, grouping, metric, direction, and redundancy \citep{zhang2024patching,heimersheim2024patching,makelov2024illusion}. Candidate evidence and generated answers can also diverge: increased target evidence can stay below an answer-selection threshold \citep{wangxu2026knowing}. In a text-only setting, language models likewise linearly encode that a math or code question is structurally unanswerable, but often still answer it \citep{du2026recognition}. Our entry-wise interventions apply these cautions: we patch the main entry or all later entries alone, specify those choices, retain magnitude controls, and report candidate scores separately from generated answers.

\textbf{Additional visual-reliance controls.}
MMVP tests difficult visual distinctions associated with similar CLIP representations \citep{tong2024eyes}, and VALSE and SugarCrepe use language-controlled contrasts to limit textual shortcuts \citep{parcalabescu2022valse,hsieh2023sugarcrepe}. FiVL compares answer-region dependence with same-area random masks \citep{aflalo2024fivl}. These controls motivate retaining native capability and target/control components, without identifying the same internal intervention. Visual concepts can emerge in frozen text models \citep{schwettmann2023multimodal}, and image-token editing and transfers further distinguish available information from its use \citep{jiang2025editing,cohen2025performance}. POPE balances yes/no object-presence questions \citep{li2023pope}. Our attribute-generation and continuation-score endpoints instead have different answer events, conditioned on native entry, spatial correspondence, and receiving state. Unlike controls that change the question, mask the answer region, edit image tokens, or probe frozen text models, our entry-wise interventions give the main entry and the later entries different versions of one image.

\section{Data use and availability}
\label{app:data}
We reuse existing datasets and checkpoints, with no new participants or human judgments, and DocVQA was obtained through its official access process. Recolored photographs are counterfactual stimuli (edited images showing colors the original scenes lacked). The photograph in Figure~\ref{fig:overview} is \href{https://www.flickr.com/photos/mualphachi/3736968562/}{\emph{River Brent at Brentford}} by Maxwell Hamilton (Flickr), licensed \href{https://creativecommons.org/licenses/by/2.0/}{CC BY 2.0}; we cropped it and recolored the bus. We assert neither institutional approval nor unrestricted redistribution rights.

The \texttt{supplementary\_data} folder holds the complete estimates of Appendices~\ref{app:v14}, \ref{app:v15} and~\ref{app:coding} as twelve CSV files, with a README that defines every column. For the two quality experiments (fixed-contrast and calibrated), the files give the unrounded estimates from the same analysis that produced the paper's tables, with descriptive 95\% intervals; a zero standard error is flagged explicitly in the fixed-contrast files and shown as an empty interval in the calibrated files. Nothing is re-estimated. Rows identify model, task, readout, response kind, degradation level, metric and invalid-answer policy, so any estimate quoted in the paper can be traced to one row. The three coding files are recomputed from the per-question observations with the paired bootstrap of Appendix~\ref{app:coding_design}, and their rows identify model, task, image set and configuration.

\end{document}